\documentclass{article} 
\usepackage{iclr2026_conference,times}

\usepackage{hyperref}
\usepackage{float}

\usepackage{standalone}
\usepackage{amsthm,amsmath,amssymb}
\usepackage{mathtools}
\usepackage{tikz}
\usepackage[ruled]{algorithm2e}
\usepackage{caption}
\usepackage{subcaption}
\usepackage[capitalise,nameinlink]{cleveref}
\usepackage{bm}
\usepackage{enumitem}  
\usepackage{booktabs}  
\usepackage{colortbl}  
\usepackage{multirow}
\usepackage{array}
\usepackage[most]{tcolorbox}

\graphicspath{ {figs/} }

\SetKwInOut{Input}{Input}
\SetKwInOut{Output}{Output}
\SetKwComment{Comment}{\(\triangleright\)\ }{}
\SetCommentSty{itshape}

\setlist[itemize]{leftmargin=*}
\setlist[enumerate]{leftmargin=*}

\theoremstyle{definition}

\theoremstyle{remark}

\newcommand{\1}{\bm{1}}

\newcommand{\basecfx}{MILP-MinDist}
\newcommand{\confexnaive}{CONFEX-Naive}
\newcommand{\confex}{CONFEX}
\newcommand{\confexlcp}{CONFEX-LCP}
\newcommand{\confextree}{CONFEX-Tree}
\newcommand{\confextreelcp}{CONFEX-Tree}

\newcommand{\confexapprox}{CONFEX-Tree}

\usepackage{amsmath,amsfonts,bm}

\def\eqref#1{equation~\ref{#1}}

\def\1{\bm{1}}

\DeclareMathAlphabet{\mathsfit}{\encodingdefault}{\sfdefault}{m}{sl}
\SetMathAlphabet{\mathsfit}{bold}{\encodingdefault}{\sfdefault}{bx}{n}

\DeclareMathOperator*{\argmin}{arg\,min}

\usepackage{url}

\title{CONFEX: Uncertainty-Aware Counterfactual Explanations with Conformal Guarantees}

\author{Aman Bilkhoo, Milad Kazemi \& Nicola Paoletti \\ 
Department of Informatics, King's College London\\
\texttt{\{aman.bilkhoo,milad.kazemi,nicola.paoletti\}@kcl.ac.uk} \\
\And
Mehran Hosseini \\
Department of Computer Science, University of Manchester\\
Department of Informatics, King's College London\\
\texttt{mehran.hosseini@manchester.ac.uk} \\
}

\iclrfinalcopy 
\begin{document}

\maketitle

\begin{abstract}

Counterfactual explanations (CFXs) provide human-understandable justifications for model predictions, enabling actionable recourse and enhancing interpretability. To be reliable, CFXs must avoid regions of high predictive uncertainty, where explanations may be misleading or inapplicable. However, existing methods often neglect uncertainty or lack principled mechanisms for incorporating it with formal guarantees. We propose CONFEX, a novel method for generating uncertainty-aware counterfactual explanations using Conformal Prediction (CP) and Mixed-Integer Linear Programming (MILP). CONFEX explanations are designed to provide local coverage guarantees, addressing the issue that CFX generation violates exchangeability. To do so, we develop a novel localised CP procedure that enjoys an efficient MILP encoding by leveraging an offline tree-based partitioning of the input space. This way, CONFEX generates CFXs with rigorous guarantees on both predictive uncertainty and optimality. We evaluate CONFEX against state-of-the-art methods across diverse benchmarks and metrics, demonstrating that our uncertainty-aware approach yields robust and plausible explanations.

\end{abstract}

\section{Introduction}
\label{sec: Introduction}
Machine learning models are deployed in high-stakes decision-making scenarios like loan approvals, medical diagnoses, and employment screening. In these contexts, algorithmic recourse---providing actionable feedback to individuals influenced by these decisions---is not just a technical concern but also an ethical and legal imperative. Although the legal status of ``right to explanations'' under the EU's General Data Protection Regulation (GDPR) remains contested \citep{wachter2017counterfactual,Selbst2018TheIA}, there is growing consensus that individuals should be offered meaningful information about algorithmic decisions that impact them \citep{Eale2017AlgSlave,Binns+18-Percentage}.

Counterfactual explanations (\emph{CFX}) were formally introduced by \citet{wachter2017counterfactual} as a method for algorithmic recourse. 

CFXs answer questions like: ``What minimal changes to my input features would have altered the model's decision desirably?'', and Wachter's formalisation focuses on finding counterfactual explanations that are minimally close to the original point (\emph{factual instance}) or have sparse feature changes. 
These criteria of closeness and sparseness have been extended in later methods to other desiderata such as diversity, causality, actionability, and plausibility, to generate explanations that work better as a recourse path and are distinguished from adversarial examples. 

However, most existing CFX methods fail to account for the inherent uncertainty in both data and model predictions. 
This is problematic because explanations that ignore uncertainty may lead to false confidence in suggested changes, potentially resulting in ineffective recourse actions when deployed in practice. Uncertainty quantification in CFX is thus crucial for generating reliable and actionable insights.

We introduce \confex, an uncertainty-aware CFX generator that builds on \emph{Conformal Prediction} (\emph{CP}) \citep{vovk2005algorithmic,angelopoulos2023conformal}. CP is a popular uncertainty quantification framework that offers distribution-free and finite-sample coverage guarantees. It works by using calibration data to construct prediction regions that contain the true (unknown) outcome with a user-specified probability. CP does not require assumptions on the data distribution and the underlying model, except that the calibration data and the test point must be exchangeable. The core idea of our \confex\ method is to constrain the search space for CFXs only to those points leading to a singleton prediction region $\{y^+\}$, i.e., points that yield the desired outcome $y^+$ with a high degree of certainty, since  non-singleton CP regions represent uncertain predictions. 

\begin{figure*}[]
  \centering
  \begin{subfigure}[b]{0.245\textwidth}
    \includegraphics[width=\linewidth]{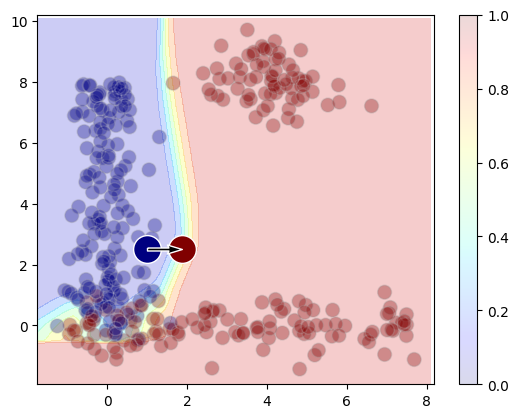}
    \caption{Minimum distance}
    \label{subfig: Introduction-a}
  \end{subfigure}
  \begin{subfigure}[b]{0.245\textwidth}
    \includegraphics[width=\linewidth]{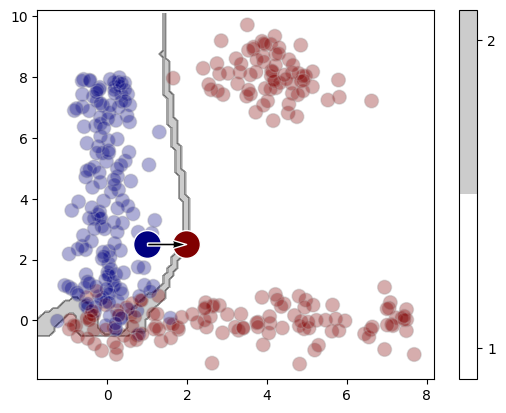}
    \caption{CONFEX-Naive}
    \label{subfig: Introduction-b}
  \end{subfigure}
  \begin{subfigure}[b]{0.245\textwidth}
    \includegraphics[width=\linewidth]{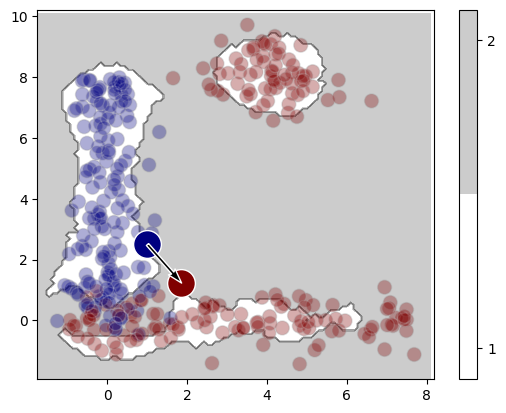}
    \caption{CONFEX-LCP}
    \label{subfig: Introduction-c}
  \end{subfigure}
  \begin{subfigure}[b]{0.245\textwidth}
    \includegraphics[width=\linewidth]{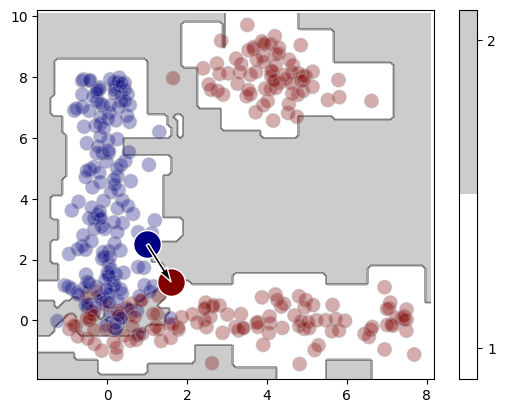}
    \caption{CONFEX-Tree}
    \label{subfig: Introduction-d}
  \end{subfigure}
  \caption{Counterfactuals produced for the same factual instance (marked in blue) for a MLP classifier using approaches \basecfx, \confexnaive, \confexlcp, \confextree. CONFEX approaches use bandwidth as 35\% of the median pairwise distance between calibration points, and alpha as 2\%.}
  \label{fig: Introduction}
\end{figure*}

To illustrate our methods, \cref{subfig: Introduction-a} displays CFXs produced over a synthetic 2D dataset inspired from  \cite{poyiadzi2020face}. We can observe that  counterfactuals produced by the minimal distance approach and by a naive application of CP to the CFX generation problem, called \confexnaive\ (\cref{sec: Naive}), fail to be plausible with respect to the data distribution. 

These issues with naively applying CP to CFX generation stem from the fact that the generated (test-time) CFX may not be exchangeable with the calibration points, thereby affecting the validity of CP's guarantees.  
We solve this by imposing stricter coverage requirements for CP: we build prediction regions that approximately\footnote{Exact conditional guarantees for CP are known to be impossible unless the inputs are discrete \citep{vovk2012conditional,barber2020limitsdistributionfreeconditionalpredictive}.} attain \textit{local} (aka \textit{test-conditional}) guarantees, i.e., the target coverage probability is achieved for \textit{any} test point. In contrast, normally, CP guarantees are marginal, i.e., the coverage probability is averaged over the joint calibration and test distribution.  

Our \confex\ method relies on a \emph{Mixed-Integer Linear Programming} (\emph{MILP}) encoding of the optimisation problem, which not only guarantees optimality of solutions but also ensures satisfaction of the CP constraints. We present two methods for incorporating local coverage constraints. The first is \textit{localised CP} \citep{guan2023localized}, which frames conditional coverage as a covariate shift problem \citep{tibshirani2019conformal}. However, it requires encoding and solving calibration quantiles in MILP, which is computationally expensive and scales poorly with the dataset size. The second, more efficient, method is a KD-tree-based encoding of local calibration quantiles. For this method, we use regression trees, which can be efficiently encoded in MILP.

\noindent In summary, our main contributions are:
\begin{enumerate}
    \item a mathematical formulation for distribution-free uncertainty-aware counterfactual explanations, the first to apply conformal prediction in a principled manner (i.e., by addressing the exchangeability problem via test-conditional coverage); 
    \item a novel localised CP procedure which, with an efficient MILP encoding, for generation of CFXs, which can be used more generally to incorporate (test-conditional) CP uncertainty constraints in any search problem; 
    \item an extensive experimental evaluation demonstrating that our \confex\ method outperforms competing generators by providing more plausible and stable explanations, as well as enjoying formal guarantees on uncertainty.
\end{enumerate}


%
\section{Background}
\label{sec: Background}

\paragraph{Counterfactual Explanations} Let $ \hat f: \mathcal{X} \rightarrow \mathcal{Y}$ denote a trained classifier for which we seek to generate counterfactual explanations. Given an instance $x_0 \in \mathcal{X}$ such that $\hat f(x_0) \ne y^+$, the goal is to identify a counterfactual instance $x'$ such that $\hat f(x') = y^+$. \citet{wachter2017counterfactual} frame this as an optimisation problem and solve it via gradient descent.
\begin{equation}\label{eq:wachter}
x_\text{cf} \in \underset{x^{\prime}}{\argmin} \max_\lambda \left( \lambda\ \operatorname{yloss}\left(\hat f\left(x^{\prime}\right),y^{+}\right)+\operatorname{dist}\left(x_0, x^{\prime}\right) \right)\!.
\end{equation}

The loss function aims to find an explanation that changes the predicted class to the target class (first term), while also ensuring that the explanation is close to the input instance (second term). Closeness is often defined as an $L_p$ norm, 
which can be weighted based on the observed data (e.g. the inverse median absolute deviation), or to reflect domain knowledge \citep{dandl2020multi}. However, by optimising solely for closeness, this formulation often leads to counterfactual explanations that resemble adversarial examples and may not be actionable or robust.

Desirable properties of CFXs include \textit{validity} (prediction flips to \( y^+ \)), \textit{proximity} (closeness to the factual instance), \textit{sparsity} (few feature changes), \textit{plausibility} (realistic and likely under the data distribution), \textit{actionability} (only mutable features are altered), \textit{causality} (identified counterfactual satisfies causal relationships) and \textit{robustness} (stability under input perturbations); see \citep{verma2020counterfactual, karimi2020survey}. 

Uncertainty-aware CFX methods show promise for enhancing the robustness and plausibility of CFXs. In this line of work, \citet{schut2021generating} propose minimising predictive entropy across an ensemble of models to consider the effect of uncertain regions.
Bayesian approaches, such as CLUE \citep{antoran2020getting}, leverage predictive uncertainty from Bayesian neural networks to generate epistemically informative counterfactuals.

\paragraph{Conformal Prediction and CFXs} CP is a distribution-free inference framework that complements any predictive model with rigorous uncertainty quantification. CP outputs prediction sets guaranteed to contain the true (unknown) outcome with a user-specified probability $1-\alpha$ without relying on asymptotic or parametric assumptions \citep{vovk2005algorithmic,angelopoulos2023conformal}. 
To construct these sets, CP performs the following steps:
\begin{enumerate}
    \item \textbf{Calibration}: use a held-out calibration dataset $\mathcal{D}_\text{cal} = \{(x_i,y_i)\}_{i=1}^n$ to find the critical value $q_{1-\alpha}$ (i.e., the $1-\alpha$ quantile) of a chosen test statistic called the \emph{(non-conformity) score} $s(x, y)$, which is normally chosen to quantify the deviation between the model prediction $\hat f(x)$ and the ground truth $y$. This step is performed only once, offline. Formally, 
\begin{equation}
{q}_{1-\alpha}= Q_{1-\alpha}\left(\sum_{i=1}^n \frac{1}{n+1} \delta_{s\left(x_i, y_i\right)}+ \frac{1}{n+1}\delta_{+\infty}\right),
\label{eq: CP Quantile}
\end{equation}
where $Q_{1-\alpha}$ is the $1-\alpha$ quantile function and $\delta_v$ is the Dirac distribution centered at $v$.

    \item \textbf{Inference}: for a test input $x^*$, construct a prediction region $C(x^*)$ by including all labels $y$ whose score is below the critical value (i.e., such that $s(x^*,y)\leq q_{1-\alpha}$). 
\end{enumerate}

\noindent The CP procedure provides the following marginal guarantee for an unseen test point $(x^*, y^*)$: 
\begin{equation}\label{eq:cp_guarantees}
    \underset{\mathcal{D}_\text{cal},(x^*, y^*)}{\mathbb{P}}(y^* \in C_{1-\alpha}(x^*)) \ge 1-\alpha.
\end{equation}
The above holds in finite sample regimes (as opposed to asymptotic) under the mild condition of exchangeability (a weaker assumption than IID), i.e., the joint distribution of calibration and test points is invariant under permutations. By marginal guarantees, we mean that the coverage probability of \eqref{eq:cp_guarantees} is achieved on average over the joint calibration and test distribution.

To our knowledge, there exist only two methods which apply conformal prediction to CFX generation: ECCCo \citep{altmeyer2024faithful} and CPICF \citep{adams2025individualised}.

CPICF \citep{adams2025individualised} assumes an alternative ``individualised'' setting,  where an institution holds a private black-box classifier and aims to provide CFXs to individuals without disclosing the classifier. The knowledge of each individual is modelled by their own classifier, and the organisation produces a CFX to reduce uncertainty in the global classifier via CP. This is a fundamentally different setting to ours, furthermore CPICF's formulation does not retain any formal CP guarantees.

In the standard setting, ECCCo extends Wachter's formulation (\eqref{eq:wachter}) with two additional terms: one that optimises the energy of the identified counterfactual to enhance plausibility, and one that minimises uncertainty through the smooth conformal set size loss of \citet{stutz2022learningoptimalconformalclassifiers}. However, ECCCo has the following drawbacks: 
1) it incorporates conformal prediction, but in a way that does not address 
exchangeability issues, which we detail in \cref{Subsec: issues}; 2) the procedure does not guarantee CP regions will have the required size (e.g., singletons); 3) it relies on energy-based training to obtain plausible CFXs.  As we will show, our approach instead induces plausible CFXs solely by using CP constraints, formulating these constraints to enforce local validity (thereby solving the exchangeability issues), and thanks to the MILP formulation, it ensures satisfaction of the set size constraints whilst being optimally close.

\paragraph{Mixed Integer Linear Programming (MILP) and CFXs} 
MILP provides a framework for formulating and deriving CFXs as a constraint-solving problem. The problem is of finding 
a point $x'$ which minimises the distance to the original instance $x_0$ whilst being classified as $y^+$.

\begin{equation}\label{eq:cfx-base}
    x_\text{cf} \in \underset{x^{\prime}}{\argmin} \operatorname{dist}(x_0, x^{\prime}) \quad
    \text { s.t. } \hat f(x^{\prime})=y^+
\end{equation}

We refer to this method as \basecfx, and it serves as a baseline for our \confex\ method.

For the model $f$ to be encoded in MILP, $f$ must be linearly representable; this is the case for e.g. 
linear classifiers and multilayer perceptrons with ReLU activations, as well as non-differentiable models such as decision trees. Neural network layers like sigmoid or softmax are not linearly representable, but can be omitted from the MILP encoding if used at the last layer since we can identify if $f(x_\text{cf})=y^+$ based on the logits alone.

When presented to an MILP solver, this approach is guaranteed to yield a valid and optimal CFX, if such an explanation exists. Gradient-based methods, on the other hand, are incomplete, meaning that they may fail to find valid CFXs or may return suboptimal solutions.

We note that properties like causality and actionability can be incorporated in \eqref{eq:cfx-base} through MILP constraints on the input variables; similarly, a set of diverse explanations (as opposed to an individual one) can be generated by repeatedly solving the problem and adding constraints or objective function terms to block or penalize explanations similar to those already identified \citep{kanamori2020dace}. By adding such constraints, our method can accommodate these desiderata as well.

\section{CFXs with CP Constraints: A Naive Attempt}
\label{sec: Naive}

We first present a naive approach to apply conformal prediction to minimise the uncertainty in the generated CFX, which we call \confexnaive. This approach extends \basecfx\ (see \eqref{eq:cfx-base}) by restricting the search space to points yielding the singleton CP region $\{y^+\}$, i.e., points attaining the target class and with a high degree of certainty:
\begin{equation}\label{eq:confex_naive}
    x_\text{cf} \in \arg \min _{x^{\prime}}  \operatorname{dist}(x_0, x^{\prime}) \quad
    \text { s.t. } C_{1-\alpha}(x^{\prime})=\{y^+\}
\end{equation}

Note that the above constraint is equivalent to the constraints $s(x^{\prime},y^+)\leq q_{1-\alpha}$ and $\bigwedge_{y\neq y^+} s(x^{\prime},y)> q_{1-\alpha}$. 

The quantile $q_{1-\alpha}$ is pre-computed on the held-out calibration set.

For multi-layer perceptrons, we use the following log-likelihood ratio as the score function
\begin{equation}\label{eq:our_score}
    s(x, y) = \log \left( \dfrac{\operatorname{max}_{y' \ne y} p(x)_{y'}}{p(x)_{y}}\right),
\end{equation}
where $p(x)_{y}$ is the softmax probability of $y$ predicted by the model $f$ for input $x$. When the correct class is predicted, the ratio is below 1 and we obtain a negative score. When the model is wrong, the ratio is positive and the score grows bigger as the model confidence on $y$ decreases relative to that on the predicted class. 
Importantly, \eqref{eq:our_score} can be equivalently expressed in a linear form as $s(x, y) = - l(x)_y + \operatorname{max}_{y' \ne y} l(x)_{y'}$, where $l(x)$ is the predicted vector of logits, making it efficiently representable in MILP.

\paragraph{Relation with \basecfx} We note that our score function is well-formed, i.e., $s(x,y)$ is lowest when $y$ is the label predicted by the model $f$ (and, in particular, $s(x,y)$ increases as the softmax probability of $y$ decreases). Thus, when a CP prediction region returns the singleton $\{y^+\}$, then $y^+$ is the class with the lowest score, i.e., the class predicted by $f$. That is, for any $\alpha \in (0,1)$, $C_{1-\alpha}(x)=\{y^+\} \rightarrow f(x)=y^+$. This implies that the feasible set of CONFEX is a subset of that of \basecfx, and so, CONFEX explanations can never attain smaller (better) distances than CFX-base. 
Importantly, since the above property holds for any $\alpha$, it also holds for any choice of quantile $q_{1-\alpha}$. This property also applies to the localised CP methods described later, which define a different quantile value.

\subsection{Need for Conditional Guarantees} \label{Subsec: issues}

A visual example of using \confexnaive\ to generate a counterfactual explanation is shown in Figure \ref{fig: Introduction} (plot b). We observe that that the effect of including the singleton set size constraint is to push the counterfactual explanation closer to the further past the decision boundary compared to \basecfx\ (plot a), which is desirable since the identified CFX would resemble less an adversarial example. However, the counterfactual explanation the identified CFX is somewhat counterintuitive: it lies in an area without local datapoints, i.e., away from the data support (see plot d). Since the CP constraints enforce low-uncertainty predictions, we would expect to find the CFX in a region where datapoints unambiguously belong to the target class, and not in regions near the decision boundary, where multiple classes overlap, or with no or little data support.

The main issue is that \confexnaive\ can return CFXs that are not exchangeable with the calibration points, violating CP's marginal guarantees. Hence, our prediction regions should be valid  \textit{for any} choice of test inputs (not just exchangeable ones), requiring the coverage requirements to be strengthened to enforce \textit{conditional validity}, i.e., for \emph{any} choice of $x= x'$, the following must hold:
\begin{equation}\label{eq:conditional_validity}
    \underset{\mathcal{D}_\text{cal},(x, y)}{\mathbb{P}}\left(y \in C_{1-\alpha}(x) \mid x = x'\right) \ge 1-\alpha.
\end{equation}

However, unless the inputs are discrete, the above exact conditional guarantees are known to be impossible if we require distribution-free and finite-sample guarantees  \citep{vovk2012conditional,barber2020limitsdistributionfreeconditionalpredictive}. 

To solve this issue, among the several methods recently proposed for CP with approximate conditional validity \citep{jung2022batch,hore2023conformal,ding2023class,gibbs2025conformal,cabezas2025regression}, we focus on the \emph{localised CP (LCP)} method of \citet{guan2023localized}, described next.

\section{The CONFEX Approach}
\label{sec: Theory}

Our method CONFEX uses Localised Conformal Prediction (LCP) to generate CFXs with more principled, local coverage guarantees. We introduce two variants: CONFEX-LCP, which encodes LCP constraints via MILP, and CONFEX-Tree, which also provides local guarantees via MILP but is more computationally efficient thanks to an offline tree-based representation of the local quantiles.

\subsection{Localised Conformal Prediction (LCP) and \confexlcp} 

Localised Conformal Prediction (LCP) \citep{guan2023localized} relaxes strict conditional coverage (see \eqref{eq:conditional_validity}) by requiring coverage to hold only within a local neighbourhood around a test input $x^*$. To achieve this, LCP reweights the calibration points as if they were drawn under the localised distribution of $x^*$, thereby restoring exchangeability. The reweighted probabilities are computed by a \textit{localiser kernel} $H:\mathcal{X}\times\mathcal{X}\to[0,1]$, which measures how ``close'' $x'$ is to $x$, with $H(x,x)=1$. In our method, we use the $L_1$-box kernel
\begin{equation}\label{eq:l1-kernel}
H(x, x') = \mathbf{1}(\| x-x'\|_1 \le h),
\end{equation}
where $h$ is the kernel bandwidth controlling the degree of localisation. For numerical and ordinal features, the $L_1$ distance is computed after normalisation; for categorical features, we require exact matches over all or some categorical features, else $H(x,x')=0$. Other kernels (e.g., based on infinity norm or Gaussian smoothing) are also possible. 

For a test input $x^*$, the local quantile is
\begin{equation}
q^{\text{LCP}}_{1-\alpha}(x^*)=Q_{1-\alpha}\left(\sum_{i=1}^n w_i \delta_{s(x_i, y_i)}+w^* \delta_{+\infty}\right),
\label{eq: Local Quantile}
\end{equation}

where $w_i=\frac{H\left(x^*, x_i\right)}{W}$ for $i =1,\ldots,n$ and $w^*=\frac{H\left(x^*, x^*\right)}{W}=\frac{1}{W}$, with 
$W=1+\sum_{i=1}^{n} H\left(x^*, x_i\right)$ being a normalizing factor.

This reweighting step and the resulting prediction region $C^{LCP}_{1-\alpha}(x^*)=\{y : s(x^*,y)\leq q^{\text{LCP}}_{1-\alpha}(x^*)\}$ ensure, for any test point $x^*$, the following approximate conditional guarantee:
\begin{equation} \label{eq: lcp_guarantee}
    \underset{\mathcal{D}_\text{cal}\sim P^n_{X,Y},(x, y)\sim P_{X,Y}^*}{\mathbb{P}}\left(y \in C_{LCP,1-\alpha}(x) \right) \ge 1-\alpha,
\end{equation}
where $P^n_{X,Y}$ is the (product) distribution of the $n$ calibration points, and $P_{X,Y}^* = P_{Y\mid X} \times P^*_X$ is the localised test distribution, with $P^*_X=P_{X} \circ H(x^*,X)$ being the distribution of $X$ obtained by applying to $P_X$ the kernel $H$ centered at $x^*$.

\paragraph{\confexlcp}  
We extend \confexnaive\ by replacing CP regions with LCP regions, yielding more principled and adaptive counterfactual generation. Formally,
\begin{equation}\label{eq:confex_lcp}
x_\text{cf} \in \arg \min _{x^{\prime}} \operatorname{dist}(x_0, x^{\prime}) 
\quad \text{s.t. } C^{LCP}_{1-\alpha}(x^{\prime})=\{y^+\},
\end{equation}
which enforces $s(x',y^+)\leq q^{LCP}_{1-\alpha}(x')$ and $s(x',y)>q^{LCP}_{1-\alpha}(x')$ for all $y\neq y^+$. Unlike \confexnaive, which uses a single global quantile $\hat q$, here the quantile depends on the candidate $x'$, requiring explicit encoding in the MILP formulation (see Algorithm~\ref{alg: confex-lcp-quantile} in the Appendix). This introduces additional variables and big-M constraints linear in the calibration set size. \cref{fig: Introduction} (plot c) shows a CFX computed using \confexlcp.  

\paragraph{Properties.}
Thanks to the LCP method, \confexlcp\ computes quantiles using only points local to the test input $x$, where locality is defined by the L1 kernel. This yields more adaptive and reliable uncertainty estimates than vanilla CP (and \confexnaive), with larger prediction sets in sparse or ambiguous regions, whilst ensuring that counterfactual is grounded with the data, i.e., similar (local) individuals which are correctly predicted to be in the target class. We note that features in the kernel can be assigned different weights based on domain knowledge. The choice of the kernel bandwidth $h$ is application-specific and it allows us to balance between local and marginal coverage.

\subsection{\confextree: Fast Variant of CONFEX-LCP}\label{sec: CONFEX-RT}
Due to the increased cost of resolving quantiles using MILP, LCP is infeasible for practical use with large calibration sets.  

In this section, we introduce \confextreelcp, an efficient alternative formulation of Localised CP which retains formal guarantees. \confexapprox\ leverages that decision trees are efficiently representable in MILP and uses precomputed local quantiles. While LCP operates at test-time by retaining only the calibration points within distance $h$ of the point, \confextreelcp\ works offline to determine locality constraints: it splits the feature space recursively to obtain local neighbourhoods of calibration points having kernel width of at most $h$.  

The construction procedure is inspired by kd-trees \citep{skrodzki2019kd} and detailed in Algorithm 1. 
Each leaf specifies a precomputed local quantile using only calibration points within that leaf. From these points, we also compute the midpoint of the smallest enclosing hyper-rectangle. The tree construction ensures that no two points in a leaf can have a bigger $L_{\infty}$ distance than the kernel bandwidth $h$. 
Then, each new test point $x'$ is assigned to a leaf of the tree and is associated with the corresponding quantile if $x'$ is within $L_{\infty}$ distance of $h/2$ from the midpoint, which means that it is within distance of $h$ from any calibration point of that leaf. To handle categorical features, we stratify the dataset by each combination of (all or select) categorical values and generate a tree for each stratum (which is equivalent to first splitting on all categorical features).

The resulting tree is encoded in MILP and used to provide the quantile value for the test point, replacing the LCP regions from \confexlcp\ . Formally, explanations are derived by solving
\begin{equation}\label{eq:confex_lcp}
x_\text{cf} \in \arg \min _{x^{\prime}} \operatorname{dist}(x_0, x^{\prime}) 
\quad \text{s.t. } C^{\text{Tree}}_{1-\alpha}(x^{\prime})=\{y^+\},
\end{equation}

where $C^{\text{Tree}}_{1-\alpha}$ is constructed using the local tree-based quantiles returned by Algorithm 1.

\begin{algorithm}[ht!]
\Input{Calibration set $\mathcal{D}_\text{cal}$, score function $s$, coverage level $1-\alpha$, bandwidth $h$}

\Output{Tree-based quantile encoding}
\textbf{Categorical Stratification}:
\begin{enumerate}
    \item Stratify the calibration dataset by each distinct combination of (all or some) categorical feature values.
    \item Generate a tree for each group using the Tree Construction procedure over the normalised numerical and ordinal values only.
\end{enumerate}

\textbf{Tree Construction:} 
\begin{enumerate}
    \item If the maximum range along any feature dimension of all calibration points in the node is less than $h$, stop and create a leaf node. At each leaf, compute and store:
    \begin{itemize}
        \item the $1 - \alpha$ quantile of the scores $s(x, y)$ of the calibration points assigned to the leaf;
        \item the midpoint of the calibration features in the leaf.
    \end{itemize}
    \item Otherwise, split the current node along the feature with the maximum spread, using the midpoint of that feature's values as the split point. Recurse on the left and right subsets to build subtrees.
\end{enumerate}

\textbf{Prediction for test point $x^*$:}
\begin{enumerate}
    \item Select the correct tree based on the test point's categorical values.
    \item Traverse the tree using $x'$ until reaching a leaf. Let $c$ and $q$ be its stored midpoint and quantile. 
    \item Reject point if assigned to the leaf but not local: if $\|x^* - c\|_{\infty} > h/2$, return $\infty$; o/w, return $q$.  
\end{enumerate}

\caption{\confextreelcp: Tree-based encoding of local quantiles}
\label{alg:confextree}
\end{algorithm}

\paragraph{Properties of \confextreelcp.} 
The tree constructed by the \confextreelcp\ defines a partitioning of the feature space into disjoint regions $\{\mathcal{X}_g\}_{g \in \mathcal{G}}$. Each $g$ has an associated quantile value $q_{1-\alpha,g}$ computed using only calibration points in $g$. This results in the following finite-sample group-conditional coverage guarantee
\begin{equation} \label{eq: treelcp_guarantee}
\mathbb{P}\big( y \in {C}_{1-\alpha}^\text{Tree}(x^*) \mid x^* \in \mathcal{X}_g\big) \;\ge\; 1-\alpha \qquad \text{for all }g \in \mathcal{G}, 
\end{equation}
as per \citet{vovk2012conditional}. Note that our method overapproximates the group-conditional quantiles as it assigns a quantile of $\infty$ when $x^*$ has $L_\infty$ distance more than $h/2$ from the midpoint of $g$. For this reason, it still satisfies the above guarantee.

Moreover, by construction, the groups created by \confextreelcp\ are local regions of calibration points in the feature space. Hence, we obtain an approximate conditional guarantee, as the tree approximates the conditional quantile $Q_{1-\alpha}(s|x)$ with the granularity of the approximation being controlled by the bandwidth $h$.

Finally,  \confextreelcp\ can be viewed as an instance of LCP using the following kernel
\begin{equation}
H(x, x') = \mathbf{1}(\| x-x'\|_{\infty} \le h \wedge \exists g. x,x' \in \mathcal{X}_g),
\end{equation}
i.e., both points need to belong to the same leaf and have $L_{\infty}$ distance bounded by $h$. Using this kernel, the guarantees of \eqref{eq: lcp_guarantee} also apply to \confextreelcp.


%
\section{Evaluation}
\label{sec: Evaluation}

In this section, we evaluate our method against competing CFX methods, assessing the cost (distance), plausibility and sensitivity of CFXs generated by \confextree. We explore the impact of varying the kernel bandwidth and the user-specified coverage rate, and we verify the formal coverage guarantees of \confex\ methods. We find that \confex\ consistently produces more stable and plausible CFXs across the benchmarks, provided the kernel bandwidth is appropriately chosen.

\paragraph{Experimental setup}{}

For our experiments, two classes of models are considered: multi-layer perceptrons (MLPs) and random forests (RFs). We selected four tabular datasets commonly found in the CFX literature: AdultIncome \citep{kohavi1996adult}, CaliforniaHousing \citep{pace1997sparse}, GiveMeSomeCredit and GermanCredit \citep{hofmann1994statlog}, using a training-calibration-test split of 60\%-20\%-20\% for each. 

To evaluate \confex, we compare our efficient tree-based approach \confextree\ (CTree) against competing uncertainty-aware generators: ECCCo \citep{altmeyer2024faithful}, the only other CFX method which uses CP, and a modified version of Schut \citep{schut2021generating} (called `Greedy' in our table) which uses a single MLP instead of an ensemble, as well as the \citet{wachter2017counterfactual} baseline. For tree-based models, we compare against the popular methods FeatureTweak (FT) \citep{Tolomei_2017}, which searches for possible paths which can change the classification, and FOCUS \citep{lucic2021focusflexibleoptimizablecounterfactual}, which optimises for distance over a differentiable relaxation of the tree models. As baselines, we include \basecfx\ (MinDist) and \confexnaive\ (CNaive). As discussed previously, \confexlcp\ is very expensive due to its ``direct'' (and inefficient) quantile encoding, hence, we did not conduct extensive experiments for it.

\paragraph{Metrics}{} To evaluate the CFXs, we focus on two main dimensions: plausibility and sensitivity. \emph{Plausibility} evaluates whether counterfactuals lie close to the data distribution, and is measured with the Local Outlier Factor (LOF) stratified per target class, with higher scores indicating more realistic examples. \emph{Sensitivity} (Sens) captures robustness to small perturbations of the input instance $x$; counterfactuals with low sensitivity remain consistent under such perturbations.

For each model and generator, we compute metrics from 100 generated CFXs for factual points taken from the test set, plus an additional 100 for the sensitivity metric. This process is repeated twice per dataset, and the metrics obtained are then computed and averaged to ensure statistical reliability. We also record the distance, implausibility, stability, and validity of the method. Further details on the metrics and experimental setup can be found in the appendix.

\paragraph{Evaluation of conformal guarantees} 
In the main setup, CFXs are generated for each test instance, but since their ground truth is unknown, coverage cannot be computed. We therefore run an additional simulated setup, identical to \confex\ in that it finds the \textit{closest test point} whose CP region is a singleton comprising the target class. This way, true labels are known and we can compute the empirical coverage $\mathbb{E}(\mathbf{1}(y \in C_{1-\alpha}(x)))$ over this resampling of the test set. We measure the gap between the observed coverage and the target $1-\alpha$. Note that this resampling considers only CFX-like points and hence breaks exchangeability. So, we expect \confexnaive\ to miss the coverage target and the localised procedures to fare better.

\begin{table*}[t]

\label{Table 1}

\centering
\begin{tabular}{lcccccc}
\toprule
 & \multicolumn{3}{c}{CaliforniaHousing} & \multicolumn{3}{c}{GermanCredit} \\
\cmidrule(lr){2-4} \cmidrule(lr){5-7}
 & Distance & Plausibility & Sens $(10^{-1})$ & Distance & Plausibility & Sens $(10^{-1})$  \\
\midrule
\multicolumn{6}{l}{\textbf{Multi-Layer Perceptron}} \\
MinDist       &   \textbf{0.02 ± 0.00} & 0.38 ± 0.04 & 41.83 ± 8.98  
& 1.69 ± 0.04 & 0.50 ± 0.06 & 0.09 ± 0.01 \\
ECCCo         & 0.35 ± 0.01 & -0.61 ± 0.03 & 0.24 ± 0.01
& 0.94 ± 0.01 & 0.21 ± 0.05 & 0.05 ± 0.02  \\
Greedy         &  1.71 ± 0.18 & -0.98 ± 0.02 & \textbf{0.13 ± 0.00}  
& 0.98 ± 0.05 & -0.02 ± 0.04 & 0.09 ± 0.00  \\
Wachter        & 0.08 ± 0.01 & 0.47 ± 0.07 & 1.57 ± 0.26
& \textbf{0.40 ± 0.01} & 0.77 ± 0.03 & 0.25 ± 0.00  \\
CNaive       & 0.03 ± 0.01 & 0.36 ± 0.04 & 15.70 ± 2.36 
& 1.83 ± 0.05 & 0.23 ± 0.15 & 0.08 ± 0.03   \\
CTree   & 0.21 ± 0.01 & \textbf{0.72 ± 0.02} & 0.26 ± 0.09   
& 2.58 ± 0.00 & \textbf{1.00 ± 0.00} & \textbf{0.00 ± 0.00}   \\
\midrule
\multicolumn{5}{l}{\textbf{Random Forest}} \\
MinDist            & \textbf{0.01 ± 0.00} & 0.45 ± 0.03 & 35.36 ± 15.39 
& 1.69 ± 0.05 & 0.36 ± 0.02 & 0.09 ± 0.01  \\
FT         & 0.11 ± 0.01 & 0.46 ± 0.02 & 0.53 ± 0.06
& 0.57 ± 0.05 & 0.86 ± 0.00 & \textbf{0.09 ± 0.00} \\
FOCUS & 0.09 ± 0.00 & 0.44 ± 0.06 & 5.58 ± 1.73 
 & \textbf{0.55 ± 0.13} & 0.88 ± 0.02 & 0.48 ± 0.02   \\
CNaive            & 0.03 ± 0.01 & 0.47 ± 0.05 & 9.92 ± 2.08 
& 1.71 ± 0.04 & 0.67 ± 0.07 & 0.09 ± 0.02    \\
CTree          & 0.18 ± 0.01 & \textbf{0.65 ± 0.03} & \textbf{0.42 ± 0.16} 
& 2.58 ± 0.70 & \textbf{1.00 ± 0.00} & 0.28 ± 0.26   \\

\bottomrule

\end{tabular}

\caption{Results for CaliforniaHousing and GermanCredit datasets. We set $\alpha=0.1$, and report the best result in terms of plausibility for CONFEX-Tree, which is with bandwidth $0.05$ for both datasets. Note that other methods seem to attain smaller distances than MinDist in some configurations; this is because these methods not always return valid counterfactuals. See the appendix for further discussion and full results.}
\label{Table 1}

\end{table*}

\begin{figure}[t]
\centering
\begin{subfigure}{0.315\linewidth}
    \includegraphics[width=\linewidth]{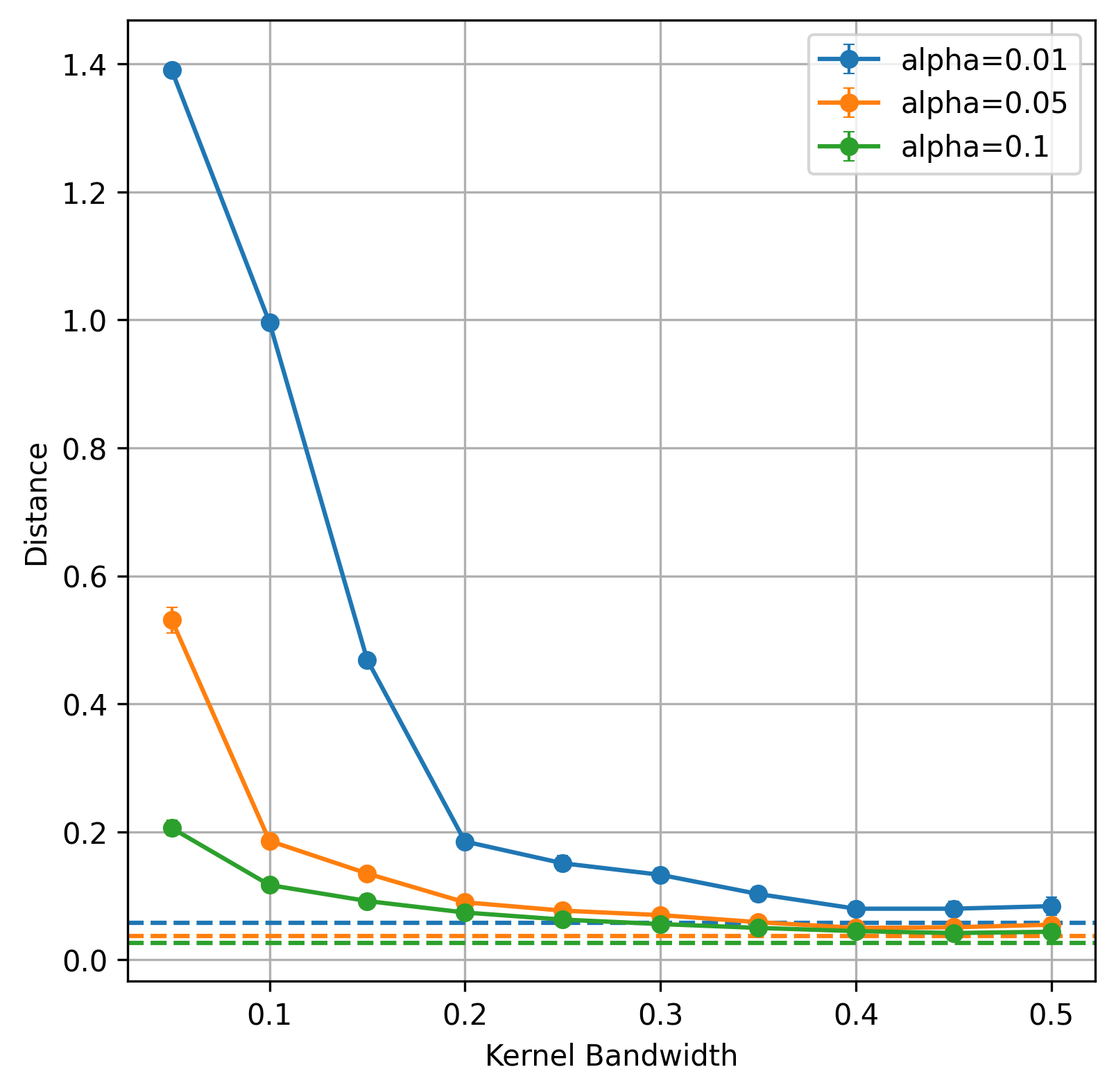}
    \caption{Distance}
\end{subfigure}
\hfill
\begin{subfigure}{0.33\linewidth}
    \includegraphics[width=\linewidth]{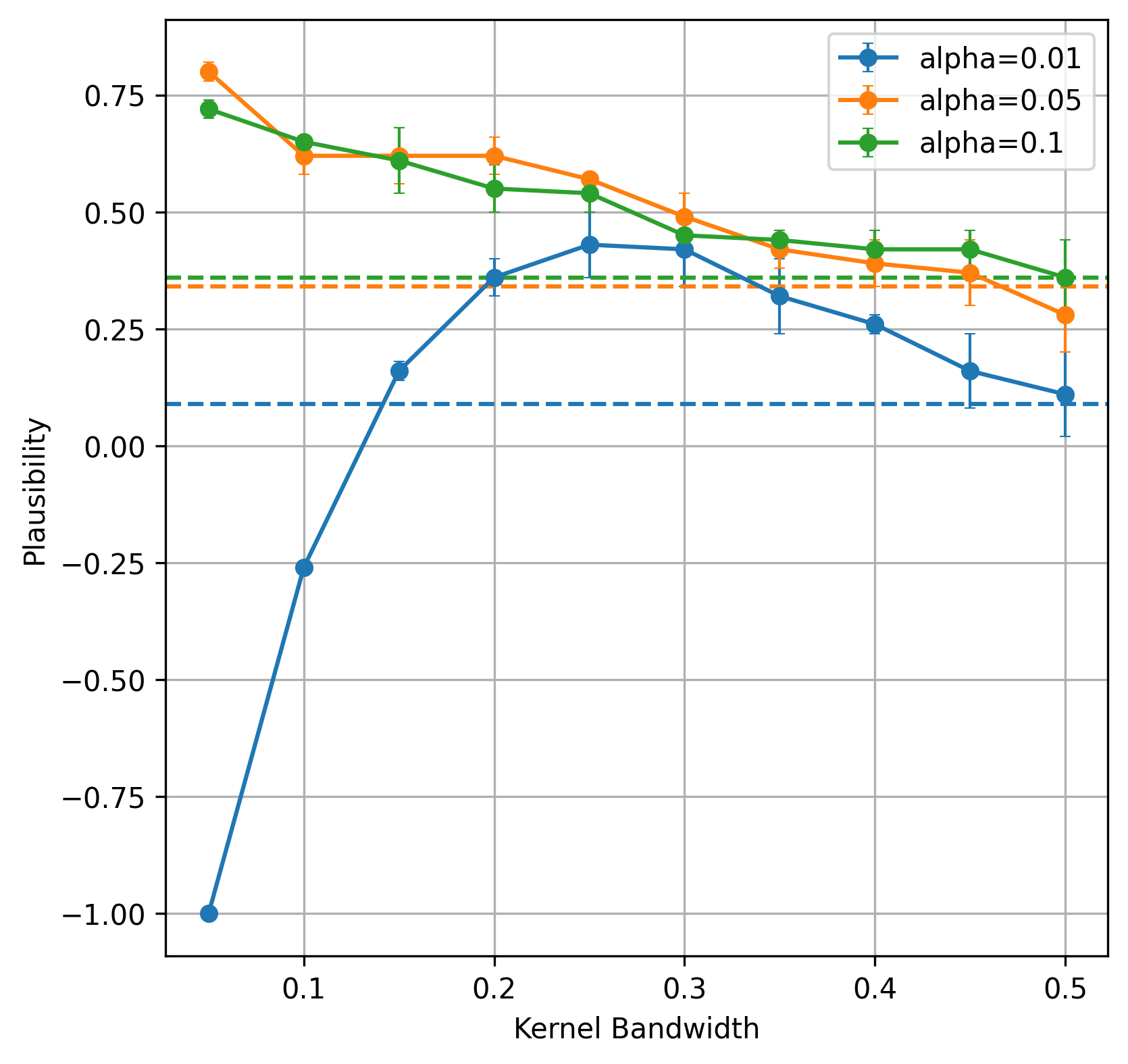}
    \caption{Plausibility}
\end{subfigure}
\hfill
\begin{subfigure}{0.32\linewidth}
    \includegraphics[width=\linewidth]{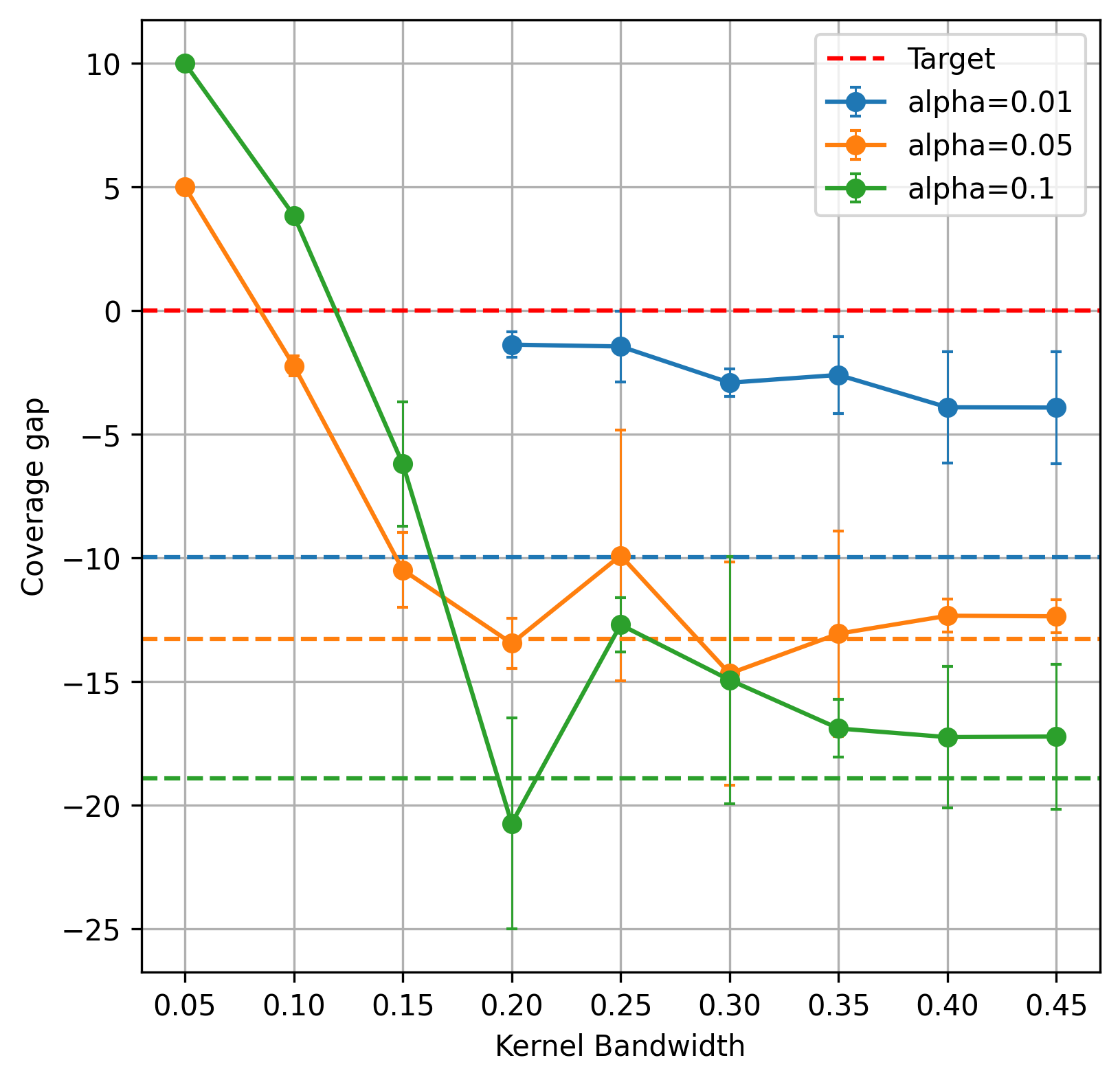}
    \caption{Coverage Gap}
\end{subfigure}
\caption{Effect of coverage rate and kernel bandwidth on metrics for CONFEX-Tree on the CaliforniaHousing dataset. \confexnaive\ is represented by dashed horizontal lines.}
\label{Figure 2}
\end{figure}

\paragraph{Results discussion}
In \cref{Table 1}, we observe that \confextree\ consistently outperforms competing methods by producing more plausible, and in many cases less sensitive explanations. In some instances, this comes with the added benefit of shorter distances, as seen with the CaliforniaHousing dataset. This is in contrast to \confexnaive\ which shows substantially lower plausibility and higher sensitivity, validating the issues illustrated in Figure 1 and further motivating the use of localisation in CP. We observe similar trends for the AdultIncome and GiveMeSomeCredit datasets, but these are reported in the appendix for space reasons.

\cref{Figure 2} illustrates the effect of varying the kernel bandwidth and coverage rate in the \confextree\ method. Increasing the coverage rate $1-\alpha$ leads to larger distances, since prediction sets become more conservative and singleton regions less frequent. Larger bandwidths yield shorter distances but at the cost of lower plausibility, as the notion of locality becomes weaker\footnote{For very small $\alpha$ ($0.01$) and small kernel bandwidths, we observe low plausibility: we conjecture this could be due to the CP method localising on outlier points.}. These observations are consistent with the fact that, as the kernel bandwidth grows, localised CP converges to standard marginal CP, as seen with \confexnaive\ in the figures.

In the (simulated) CFX setting, the Coverage Gap results confirm that vanilla CP (used by \confexnaive) fails to reach the target coverage, while localised CP with a suitably chosen kernel bandwidth succeeds. For small bandwidths (i.e., ``strong'' locality), all three choices of $\alpha$ attain or are close to the target coverage level, but the gap grows as the bandwidth increases and localisation diminishes. 
For $\alpha=0.01$ and small bandwidths, no data is obtained since no test points produced a singleton prediction region (as required by our \confex\ constraints). 
These figures demonstrate that picking a correct bandwidth is crucial for obtaining good plausibility and coverage guarantees. 

\medskip


%

%
\section{Conclusions}
\label{sec: Conclusions}
We introduced a novel MILP-based framework for generating uncertainty-aware counterfactual explanations with formal, distribution-free guarantees. By developing an efficient encoding of localised conformal prediction, we address the critical issue of exchangeability violation in the CFX search process. This allows us to enforce approximate test-conditional guarantees, ensuring the generation of provably reliable, plausible, and robust explanations.

\paragraph{Limitations} Since our approach uses MILP to solve for CFXs, it will struggle scaling to very large models; gradient-based methods like Wachter and ECCCo are less prone to this problem, but they sacrifice guarantees on CFX validity. 
Moreover, CP requires a held-out calibration dataset, which may be problematic when data is scarce. Fortunately, CP guarantees hold regardless of the calibration set size (but small sets will lead to more conservative prediction regions). 

Picking an appropriate kernel bandwidth is an additional task which requires domain knowledge or evaluation on a validation set.


%
{
  \small
  \bibliography{references}
}
\newpage
\appendix
\section{Appendix}
\label{sec: Appendix}
\subsection{Related Works}
Our work integrates three research areas: counterfactual explanations (CFXs), uncertainty quantification in explanations, and the application of conformal prediction (CP) to optimization problems. Counterfactual explanations, introduced by \citet{wachter2017counterfactual}, provide recourse by identifying minimal feature changes to alter a model's prediction. While initial work focused on validity and distance, the field has expanded to include desiderata like plausibility and actionability \citep{verma2020counterfactual, karimi2020survey}. Methodologies have also diversified from gradient-based optimization to tree-specific algorithms \citep{Tolomei_2017, lucic2021focusflexibleoptimizablecounterfactual} and constraint-based methods using Mixed-Integer Linear Programming (MILP) \citep{kanamori2020dace}. However, a critical limitation of many approaches is their failure to account for model uncertainty, which can result in misleading or brittle explanations \citep{schut2021generating}. To address this, prior works have employed Bayesian methods \citep{antoran2020getting} or model ensembles \citep{schut2021generating}. CONFEX contributes a novel, principled alternative by using Conformal Prediction. More relevant is ECCCo \citep{altmeyer2024faithful}, which uses a loss term based on the conformal set size \citep{stutz2022learningoptimalconformalclassifiers} but crucially does not address the violation of the exchangeability assumption inherent in the CFX search process.

\subsection{MILP encoding of localised CP}
The following algorithm Algorithm~\ref{alg: confex-lcp-quantile} computes the LCP quantile value in MILP. To do this, all calibration scores and calibration points must be accessible to the optimiser. Variables are constrained as distances from the test point to each calibration point, and another set of variables compute the corresponding weight according to the L1 kernel. These weights are used alongside calibration scores to identify the desired weighted quantile. This encoding is linear in the size of the calibration set.

\begin{algorithm}
\LinesNumbered
    \Input{Calibration dataset $\{(x_i,y_i)\}_{i=1}^n$, corresponding scores $\{s_i\}_{i=1}^n$, test input $x^*$, L1 localisation kernel with bandwidth $h$, level $\alpha\in(0,1)$}

    \Output{Local quantile $q^{LCP}_{1-\alpha}$}
    
    Sort $\{(x_i,y_i)\}_{i=1}^n$ in ascending order w.r.t.\ scores.

    Add $n$ real variables $d_1,\ldots,d_n$.

    For $i=1,\ldots,n$, add the L1 distance constraint $d_i=\|x_i - x^* \|_1$.

    Add $n$ binary variables $w_1,\ldots,w_n$ as the weights induced by the L1 kernel.

    For $i=1,\ldots,n$, add the constraint $w_i = \mathbf{1}(d_i \leq h)$, implemented for arbitrarily large $M>0$ as
    \[
    d_i \le h + M (1-w_i)\wedge
    d_i \ge h - M w_i
    \]

    Add $n$ binary variables $in_1,\ldots,in_n$; each $in_i$ keeps track if the score $s_i$ is below the quantile.

    Add integer variables $W$ and $W_{1-\alpha}$ denoting, respectively, the sum of all weights and of those weights whose score is below the quantile.

    Add constraints $W=\sum_{i=1}^n w_i$, $W_{1-\alpha}=\sum_{i=1}^n in_i \cdot w_i$ and $W_{1-\alpha}\geq \lceil(1-\alpha)W\rceil$. The latter expresses that the scores below the quantile have probability at least $1-\alpha$. 

    Define $W'_{1-\alpha}=\sum_{i=1}^n (1-in_i) \cdot w_i$ and add constraint $W'_{1-\alpha} \ge  \lfloor \alpha W \rfloor$

    Solve constraints and return $s_k$.
    
    $q^{LCP}_{1-\alpha}$ will be the largest calibration score $s_i$ for which $in_i=1$. To identify it, add an integer variable $k \in \{1,\ldots,n\}$.

    For $i=1,\ldots,n$, add the constraint $in_i = \mathbf{1}(i\leq k)$ using a big-M encoding as done in line 5.

\caption{Localised CP constraints in MILP}
\label{alg: confex-lcp-quantile}
\end{algorithm}

\subsection{Further discussion of \cref{Table 1}}
For GermanCredit, whilst Wachter obtained the closest counterfactuals, had a validity rate of 84\%, demonstrating how gradient-based methods may fail to correctly change prediction to the target class. ECCCo (79\%) and FeatureTweak (52\%) also suffered validity issues. On the other hand, \basecfx always found a valid counterfactual, including satisfying correct categorical and ordinal encoding unlike some of the competing tree generators, and this is reflected with an increased distance.

Note that in all figures, kernel bandwidth is measured as a multiple of the median pairwise distance between all points in the dataset. 

\newpage
\section{Further Evaluation}

\subsection{Experimental Setup}

\textbf{Generators. } For solving MILP instances, we utilise the Gurobi solver, and utilise the Gurobi Machine Learning \cite{GurobiML} library to formulate the trained classifiers as constraints. All generators, except FOCUS (using the CFXplorer package \cite{CFXplorer}) and FeatureTweak (implementation taken from CARLA \cite{pawelczyk2021carla}, and FeatureTweakPy\footnote{\url{https://github.com/upura/featureTweakPy/blob/master/featureTweakPy.py}}), were implemented as part of a Python library to generate CFXs called PyCFX. This library, containing code to reproduce our results, can be accessed at \url{https://github.com/ABilkhoo/pycfx}

\textbf{Model Configuration.} For all datasets, we used a multilayer perceptron (MLP) with 50 hidden units. The batch size was set to 64 for California Housing and German Credit, trained for 100 epochs, and 256 for GiveMeSomeCredit and Adult Income, trained for 50 epochs. For the random forest model, we also evaluated a Random Forest classifier with 5 estimators and number of leaves limited to 500 for the GiveMeSomeCredit and AdultIncome models.

\subsection{Metrics}
\label{appendix: Metric}

In order to evaluate the quality of the generated counterfactual explanations, we adopt a set of quantitative metrics that measure different aspects of their usefulness and reliability. Specifically, we focus on three core dimensions: \emph{plausibility}, \emph{sensitivity}, and \emph{stability}. In addition, we report auxiliary metrics such as the distance of counterfactuals to the original instance, the proportion of failures, and the validity rate of generated explanations. Together, these metrics provide a comprehensive view of both the fidelity and robustness of counterfactual explanations. 

\medskip

\noindent\textbf{Plausibility.}  
A counterfactual explanation should lie close to the underlying data distribution so that it represents a realistic and interpretable alternative. To assess this, we measure plausibility using the Local Outlier Factor (LOF) \citep{breunig2000lof}, which quantifies how isolated a sample is with respect to its nearest neighbours. A LOF score of $+1$ indicates that the counterfactual is consistent with observed data, whereas $-1$ suggest that the counterfactual is implausible. We use the \texttt{scikit-learn} implementation of LOF with \texttt{novelty=True} and $n\_ \text{neighbors} = 20$, stratified by the target class. In practice, we average over 100 test points. 

\medskip

\noindent\textbf{Sensitivity.}  
Beyond plausibility, we also want to assess whether counterfactuals are \emph{robust} to small changes in the input instance. Sensitivity measures how much a counterfactual explanation changes when the original instance $x$ is perturbed within a small neighbourhood. Formally, given an input $x$ and its counterfactual $x_c$, we uniformly sample a perturbed instance $x' \sim U_b(x)$ from the $\ell_2$ ball centred around the factual, compute a new counterfactual $x'_c$. Sensitivity is then defined as the relative deviation between the two counterfactuals, normalised by the cost of the initial counterfactual:
\[
\text{CFX Sensitivity} = \mathbb{E}_{x' \sim U_b(x)} \left[ \frac{\left\| x'_c - x_c \right\|_2}{\left\| x_c - x \right\|_2} \right].
\]
In practice, we sample 4 neighbours from 25 test points to inform our sensitivity metric. Intuitively, low sensitivity indicates that the explanation remains stable when the factual input undergoes small variations, thereby suggesting robustness and consistency. 

In our experiments, we choose the budget $b$ of the uniform sampling to correspond to a ball with 0.1\% of the volume of the feature space. 

\[V_{\text {ball }}=\frac{\pi^{d / 2}}{\Gamma\left(\frac{d}{2}+1\right)} r^d=b V_{\text {total }}\]

where $d$ is the number of non-categorical features in the space. Solving for $r$, 
\[r=\left(\frac{b V_{\text {total }}}{\pi^{d / 2} / \Gamma\left(\frac{d}{2}+1\right)}\right)^{1 / d}\]

This allows the same budget to be used across datasets with differing numbers of features. When sampling neighbours, we do not change categorical values and we fix ordinal values to their closest valid value.

\medskip

\noindent\textbf{Stability.}  
Complementary to sensitivity, stability measures how consistent the counterfactual is under perturbations applied directly to the counterfactual itself. That is, we perturb $x_c$ within a budgeted neighbourhood and evaluate the variance in the model predictions across these perturbed samples. Following an adaptation of \citep{dutta2022robust}, stability is computed as:
\[
\begin{aligned}
&\text{CFX Stability} =\frac{1}{K} \sum_{x^{\prime} \in N_x} \hat f\left(x^{\prime}\right)_{y^+}-
\sqrt{\frac{1}{K} \sum_{x^{\prime} \in N_x}\left(\hat f\left(x^{\prime}\right)_{y^+}-\frac{1}{K} \sum_{x^{\prime} \in N_x} \hat f\left(x^{\prime}\right)_{y^+}\right)^2}, \\
& \quad \text{where } N_x \text{ is a set of } K \text{ points sampled as } x' \sim U_b(x_c).
\end{aligned}
\]

where $\hat f\left(x^{\prime}\right)_{y^+}$ refers to the predicted probability of the target class. The metric neighbours a large mean value for the predicted probability of sampled neighbours, whilst penalising variations in these values by subtracting the standard deviation to ensure that that mean is not a combination of very high and very low values. Similarly to the Sensivity metric, $U_b(x_c)$ denotes sampling from the $\ell_2$ ball centred around the counterfactual, computing the radius in the same way, taking the budget to represent 0.1\% of the total feature volume.

Stability is high when the predictions across perturbed counterfactuals remain close to each other, which indicates that the explanation is not overly sensitive to minor fluctuations in its actualisation.  

\medskip

\noindent\textbf{Auxiliary metrics.}  
In addition to the three core dimensions, we report the following supplementary measures:
\begin{itemize}
    \item \textit{Distance:} the average L1 distance between the original instance and the counterfactual,  
    \[
    \text{Distance} = \mathbb{E}\big(\lVert x' - x \rVert_1 \big),
    \]
    which quantifies the minimality of the intervention required.
    \item \textit{Validity:} the proportion of counterfactuals that successfully change the prediction to the desired class,  
    \[
    \text{Validity} = \mathbb{E}(\mathsf{1}\{\hat f(x')=y^+\}).
    \]
     For example, invalidity could be due to numerical artefacts in encoding the models in MILP, or failure for SGD procedures to converge to a flipped class. We report whenever a method a method produces less than 90\% validity, and exclude invalid CFXs from the computation of other metrics.
    \item \textit{Failure rate:} the proportion of runs where the generator fails to produce a counterfactual, for example due to infeasible constraints in optimisation-based methods such as MILP.
    \item \textit{Implausibility:} The average distance from the counterfactual to the closest 10\% of points of the target class, similar to \cite{altmeyer2024faithful}. 
\end{itemize}

\subsubsection {Conditional coverage results}
In the additional results, we furthermore evaluate the performance of different conformal CFX generators under four evaluation settings: marginal coverage, class-conditional coverage, random binning, and counterfactual similarity. In the paper we discussed the counterfactual simulation, however we also evaluate the marginal coverage over a test set, average class-conditional coverage, average coverage over a random paritioning of the test set into 3 bins. We report the coverage gap: the difference between the target coverage and empirical coverage, in percentage points.

\newpage
\subsection{California Housing}

We use the California Housing dataset~\cite{pace1997sparse} from the StatLib repository through scikit-learn's \texttt{sklearn.datasets.fetch\_california\_housing} function\footnote{\url{https://www.dcc.fc.up.pt/~ltorgo/Regression/cal_housing.html}}. The original regression problem was changed into a binary classification task by categorizing houses based on whether the median income exceeds \$20,000 (42\% above, 58\% below). The dataset contains 8 numeric features, which we scaled to the range $(0, 1)$ using MinMax scaling.

\subsubsection{Model evaluation results}

\begin{table}[H]
\centering
\renewcommand{\arraystretch}{1.2}
\setlength{\tabcolsep}{8pt}
\begin{tabular}{lcccc}
\toprule
Repeat  & Accuracy (\%) & Precision (\%) & F1 Score (\%) \\
\midrule
repeat0, MLP & 83.58 & 83.61 & 83.59 \\
repeat1, MLP & 82.95 & 83.59 & 82.95 \\
repeat0, RF & 78.05 & 80.60 &  77.83 \\
repeat1, RF & 78.10 & 80.60 & 77.90 \\
\bottomrule
\end{tabular}
\caption{Model evaluation results, CaliforniaHousing.}
\end{table}

\subsubsection{Plots}

\begin{figure}[H]
\centering
\begin{subfigure}{0.315\linewidth}
    \includegraphics[width=\linewidth]{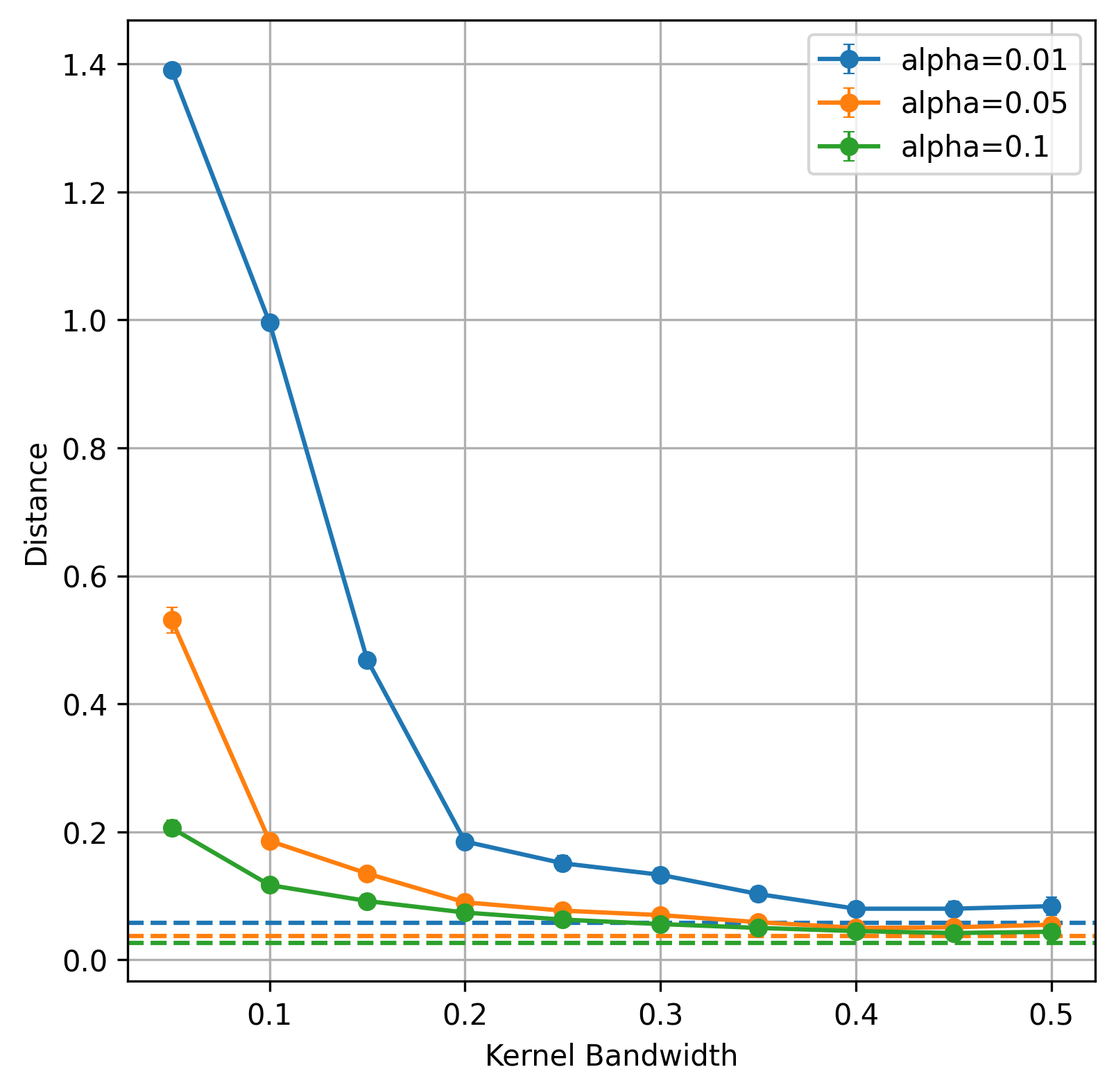}
    \caption{Distance}
\end{subfigure}
\hfill
\begin{subfigure}{0.33\linewidth}
    \includegraphics[width=\linewidth]{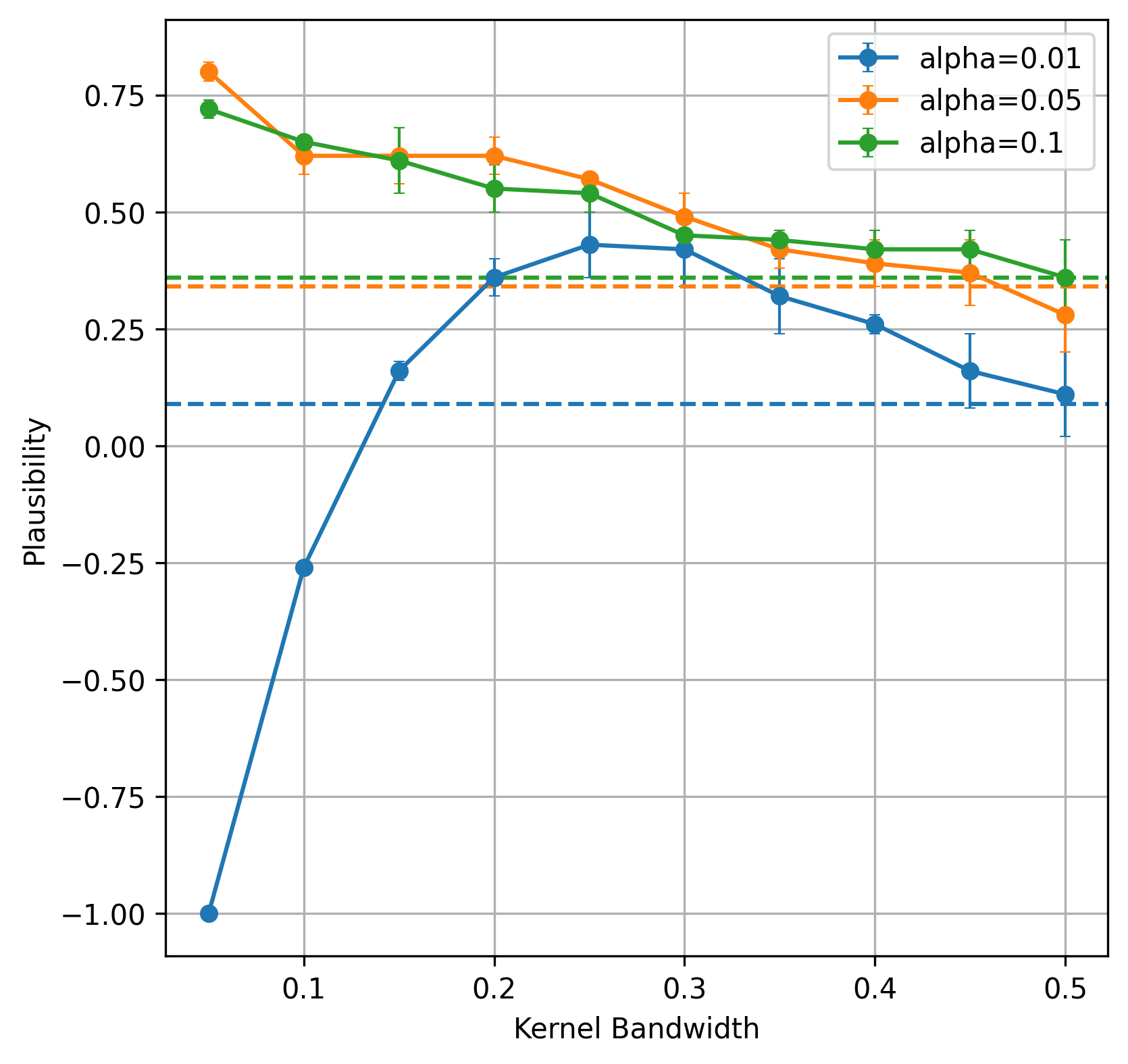}
    \caption{Plausibility}
\end{subfigure}
\hfill
\begin{subfigure}{0.32\linewidth}
    \includegraphics[width=\linewidth]{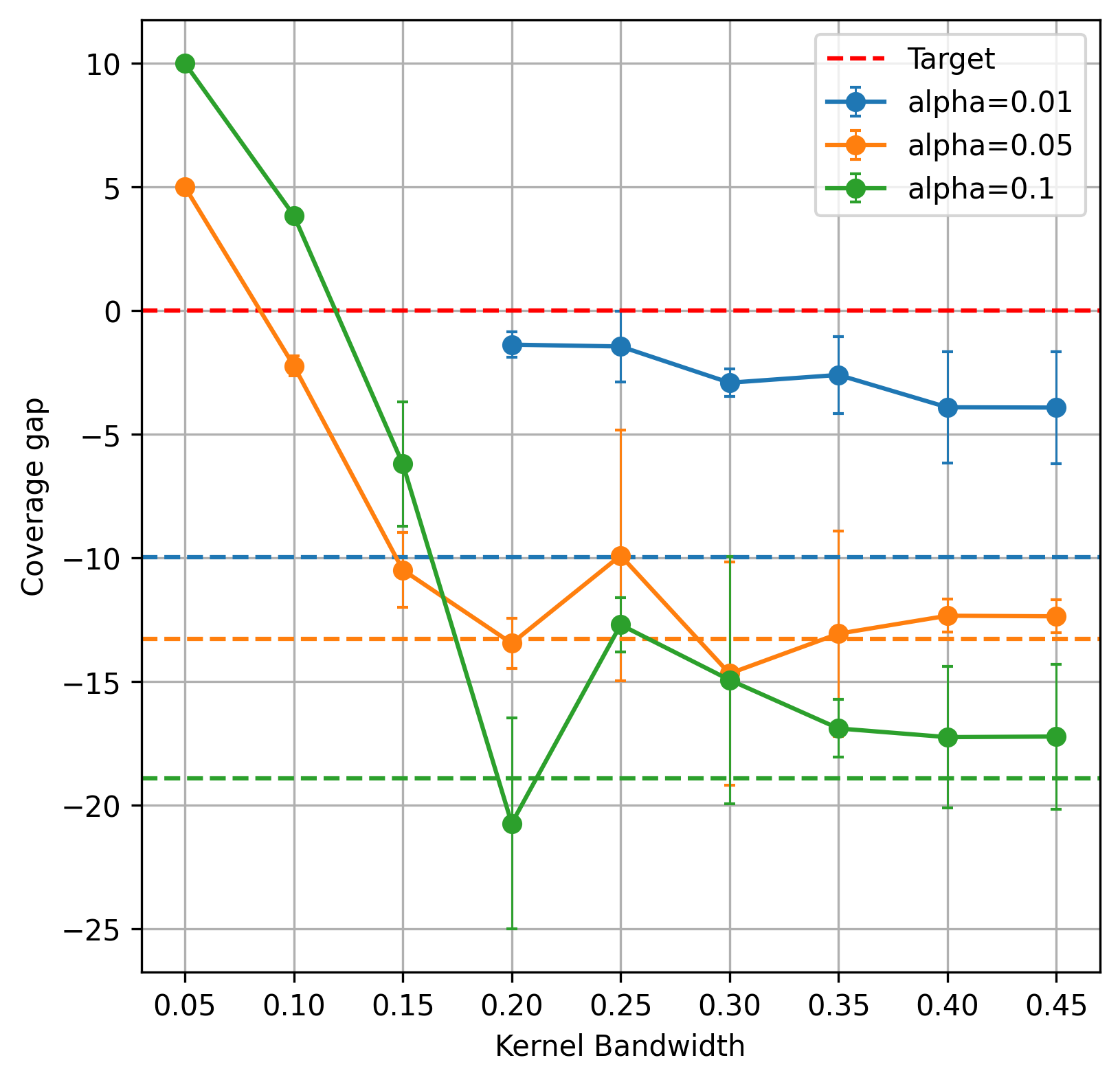}
    \caption{Coverage Gap}
\end{subfigure}
\caption{Effect of coverage rate and kernel bandwidth on metrics for CONFEX-Tree on the CaliforniaHousing dataset, MLP. \confexnaive\ is represented by dashed horizontal lines.}
\label{Figure 2}
\end{figure}

\begin{figure}[H]
\centering
\begin{subfigure}{0.315\linewidth}
    \includegraphics[width=\linewidth]{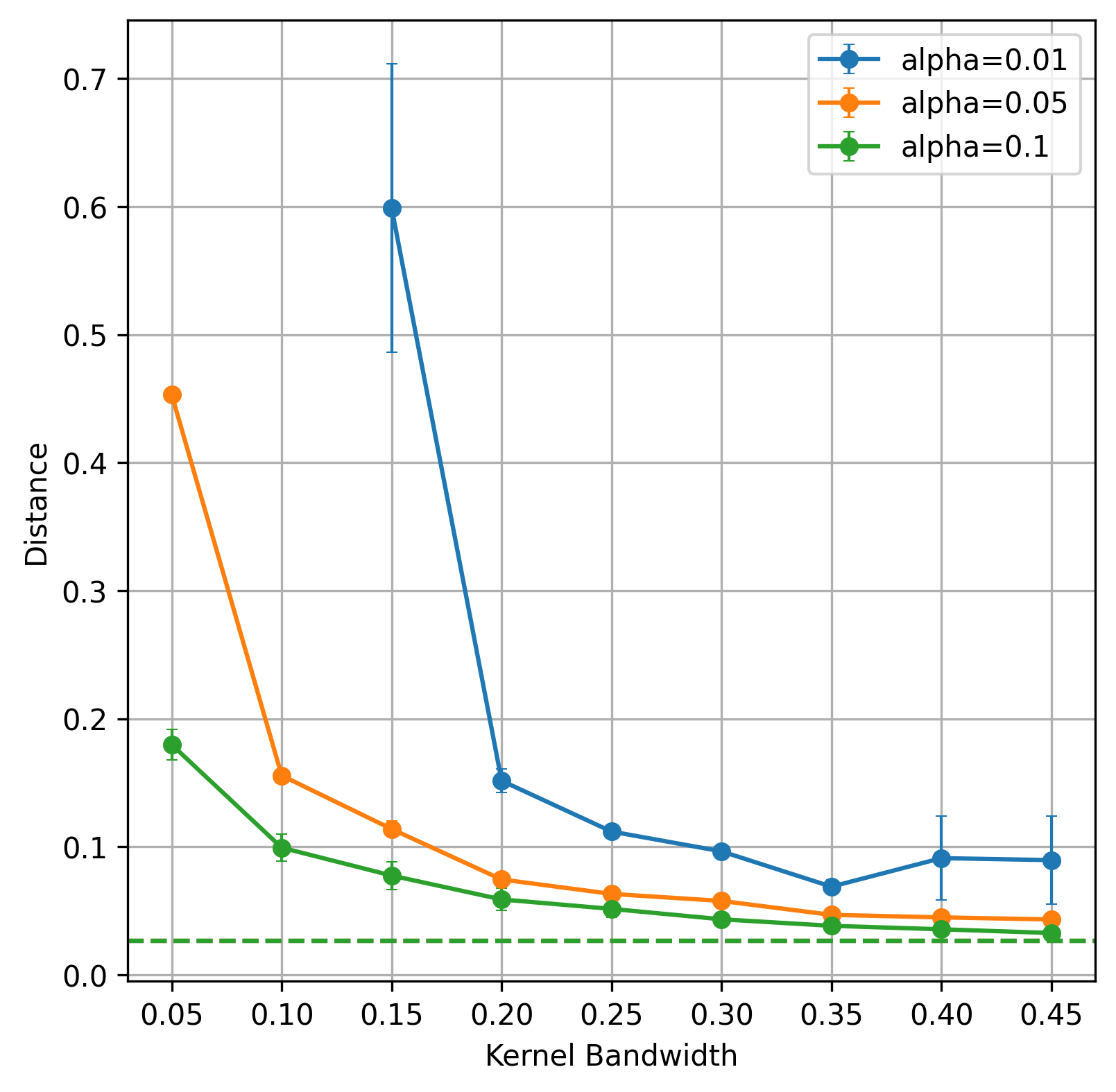}
    \caption{Distance}
\end{subfigure}
\hfill
\begin{subfigure}{0.33\linewidth}
    \includegraphics[width=\linewidth]{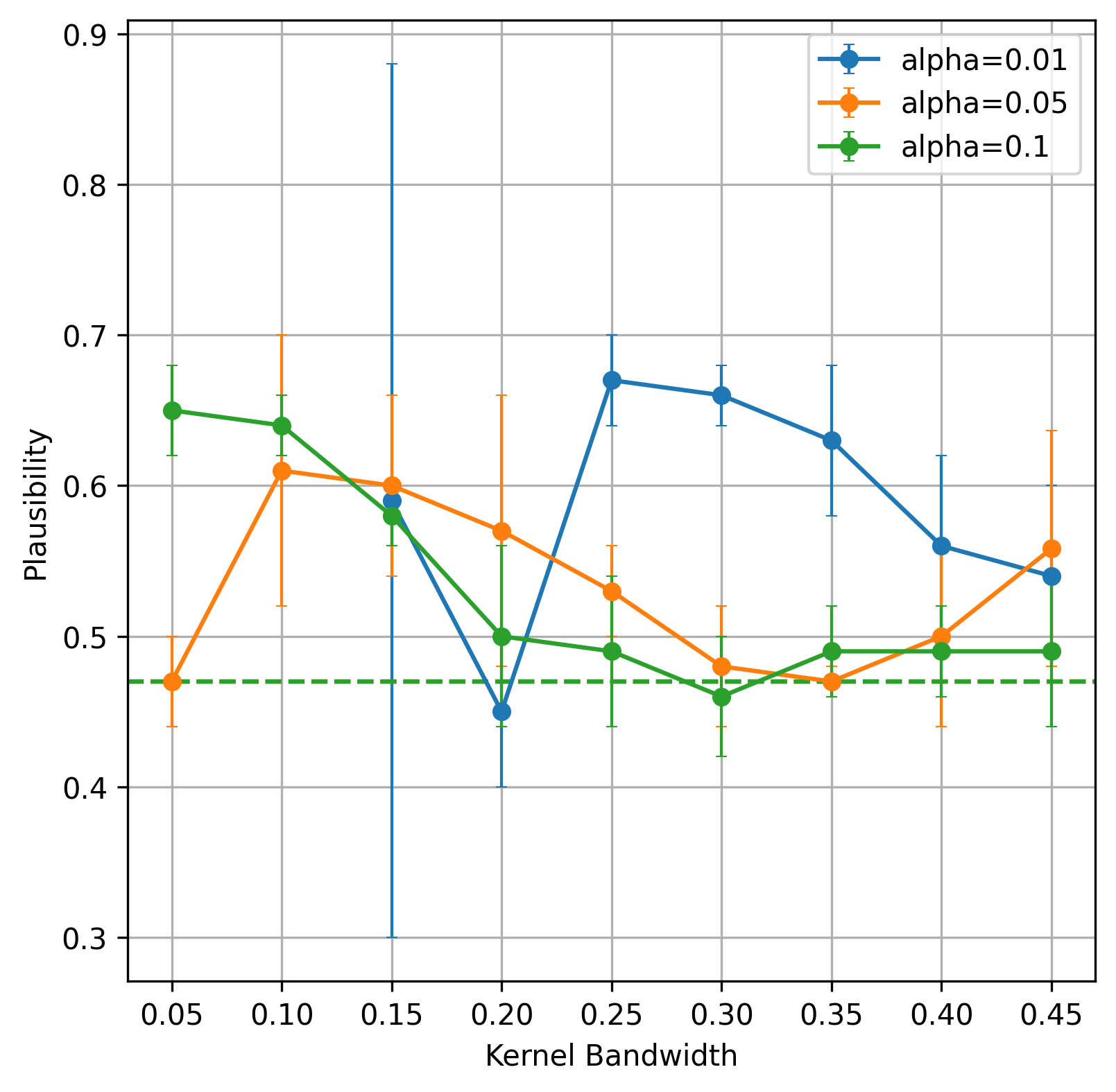}
    \caption{Plausibility}
\end{subfigure}
\hfill
\begin{subfigure}{0.32\linewidth}
    \includegraphics[width=\linewidth]{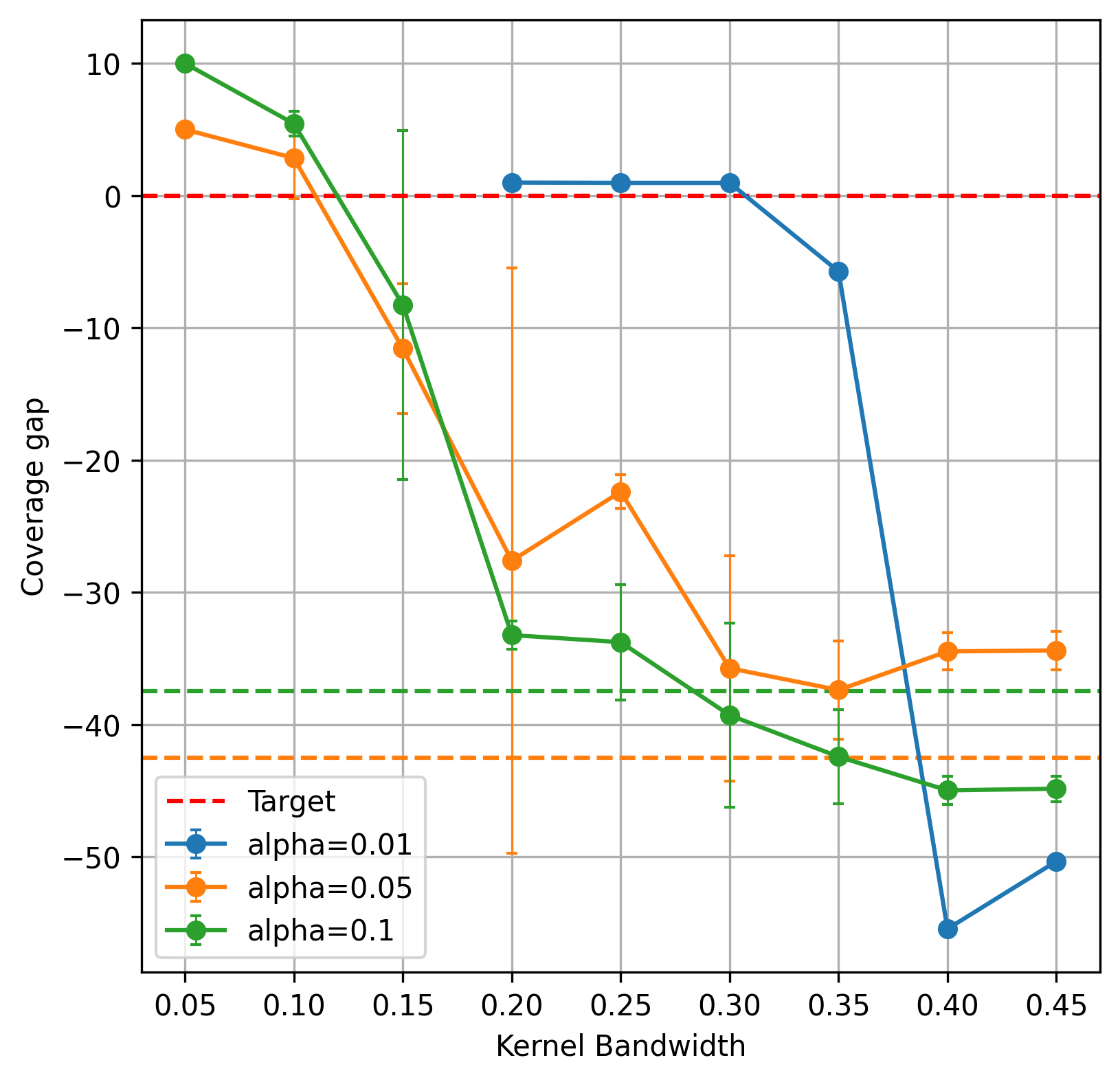}
    \caption{Coverage Gap}
\end{subfigure}
\caption{Effect of coverage rate and kernel bandwidth on metrics for CONFEX-Tree on the CaliforniaHousing dataset, RandomForest. \confexnaive\ is represented by dashed horizontal lines.}
\label{Figure 2}
\end{figure}

\subsubsection{CFX generation results}

\begin{table}[H]
\begin{tabular}{llllll}
\toprule
Generator & Distance & Plausibility & Implausibility & Sensitivity $(10^{-1})$ & Stability \\
\midrule
\textbf{MLP} &  &  &  &  &  \\
MinDist & 0.02 ± 0.00 & 0.38 ± 0.04 & 0.21 ± 0.00 & 41.83 ± 8.98 & 0.06 ± 0.02 \\
Wachter & 0.08 ± 0.01 & 0.47 ± 0.07 & 0.20 ± 0.00 & 1.57 ± 0.26 & 0.07 ± 0.02 \\
Greedy & 1.71 ± 0.18 & -0.98 ± 0.02 & 0.82 ± 0.06 & 0.13 ± 0.00 & 0.46 ± 0.03 \\
ConfexNaive &  &  &  &  &  \\
$\ \ \alpha = 0.01$ & 0.06 ± 0.01 & 0.09 ± 0.05 & 0.22 ± 0.00 & 3.27 ± 0.19 & 0.07 ± 0.02 \\
$\ \ \alpha = 0.05$ & 0.04 ± 0.01 & 0.34 ± 0.00 & 0.21 ± 0.00 & 8.01 ± 1.58 & 0.06 ± 0.02 \\
$\ \ \alpha = 0.1$ & 0.03 ± 0.01 & 0.36 ± 0.04 & 0.21 ± 0.00 & 15.70 ± 2.36 & 0.06 ± 0.02 \\
ECCCo &  &  &  &  &  \\
$\ \ \alpha = 0.01$ & 0.37 ± 0.02 & -0.67 ± 0.01 & 0.20 ± 0.00 & 0.22 ± 0.02 & 0.33 ± 0.05 \\
$\ \ \alpha = 0.05$ & 0.36 ± 0.01 & -0.62 ± 0.04 & 0.20 ± 0.00 & 0.24 ± 0.01 & 0.32 ± 0.06 \\
$\ \ \alpha = 0.1$ & 0.35 ± 0.01 & -0.61 ± 0.03 & 0.20 ± 0.00 & 0.24 ± 0.01 & 0.32 ± 0.06 \\
ConfexTree, $\alpha = 0.01$ &  &  &  &  &  \\
$\ \ \text{bw} = 0.05$ & 1.39 ± 0.01 & -1.00 ± 0.00 & 0.39 ± 0.00 & 0.00 ± 0.00 & 1.00 ± 0.00 \\
$\ \ \text{bw} = 0.1$ & 1.00 ± 0.00 & -0.26 ± 0.00 & 0.16 ± 0.01 & 0.02 ± 0.01 & 0.06 ± 0.03 \\
$\ \ \text{bw} = 0.15$ & 0.47 ± 0.01 & 0.16 ± 0.02 & 0.16 ± 0.01 & 0.08 ± 0.01 & 0.09 ± 0.01 \\
$\ \ \text{bw} = 0.2$ & 0.19 ± 0.01 & 0.36 ± 0.04 & 0.16 ± 0.00 & 0.58 ± 0.34 & 0.10 ± 0.02 \\
$\ \ \text{bw} = 0.25$ & 0.15 ± 0.01 & 0.43 ± 0.07 & 0.16 ± 0.00 & 0.85 ± 0.33 & 0.10 ± 0.02 \\
$\ \ \text{bw} = 0.3$ & 0.13 ± 0.01 & 0.42 ± 0.08 & 0.18 ± 0.00 & 1.17 ± 0.37 & 0.09 ± 0.02 \\
$\ \ \text{bw} = 0.35$ & 0.10 ± 0.01 & 0.32 ± 0.08 & 0.18 ± 0.00 & 1.87 ± 0.78 & 0.09 ± 0.03 \\
$\ \ \text{bw} = 0.4$ & 0.08 ± 0.01 & 0.26 ± 0.02 & 0.19 ± 0.00 & 2.54 ± 0.53 & 0.08 ± 0.03 \\
$\ \ \text{bw} = 0.45$ & 0.08 ± 0.01 & 0.16 ± 0.08 & 0.20 ± 0.00 & 2.41 ± 0.03 & 0.08 ± 0.03 \\
ConfexTree, $\alpha = 0.05$ &  &  &  &  &  \\
$\ \ \text{bw} = 0.05$ & 0.53 ± 0.02 & 0.80 ± 0.02 & 0.15 ± 0.00 & 0.09 ± 0.07 & 0.10 ± 0.02 \\
$\ \ \text{bw} = 0.1$ & 0.19 ± 0.01 & 0.62 ± 0.04 & 0.16 ± 0.00 & 0.44 ± 0.15 & 0.10 ± 0.03 \\
$\ \ \text{bw} = 0.15$ & 0.13 ± 0.01 & 0.62 ± 0.06 & 0.17 ± 0.00 & 1.09 ± 0.67 & 0.10 ± 0.03 \\
$\ \ \text{bw} = 0.2$ & 0.09 ± 0.01 & 0.62 ± 0.04 & 0.18 ± 0.00 & 2.54 ± 0.93 & 0.08 ± 0.02 \\
$\ \ \text{bw} = 0.25$ & 0.08 ± 0.01 & 0.57 ± 0.01 & 0.18 ± 0.00 & 3.90 ± 0.31 & 0.08 ± 0.02 \\
$\ \ \text{bw} = 0.3$ & 0.07 ± 0.01 & 0.49 ± 0.05 & 0.19 ± 0.00 & 4.76 ± 0.76 & 0.08 ± 0.02 \\
$\ \ \text{bw} = 0.35$ & 0.06 ± 0.01 & 0.42 ± 0.04 & 0.19 ± 0.00 & 6.04 ± 0.36 & 0.07 ± 0.02 \\
$\ \ \text{bw} = 0.4$ & 0.05 ± 0.01 & 0.39 ± 0.05 & 0.20 ± 0.00 & 7.58 ± 1.03 & 0.07 ± 0.02 \\
$\ \ \text{bw} = 0.45$ & 0.05 ± 0.01 & 0.37 ± 0.07 & 0.20 ± 0.00 & 7.78 ± 1.44 & 0.07 ± 0.02 \\
ConfexTree, $\alpha = 0.1$ &  &  &  &  &  \\
$\ \ \text{bw} = 0.05$ & 0.21 ± 0.01 & 0.72 ± 0.02 & 0.16 ± 0.00 & 0.26 ± 0.09 & 0.10 ± 0.03 \\
$\ \ \text{bw} = 0.1$ & 0.12 ± 0.01 & 0.65 ± 0.01 & 0.18 ± 0.00 & 1.40 ± 0.98 & 0.09 ± 0.02 \\
$\ \ \text{bw} = 0.15$ & 0.09 ± 0.01 & 0.61 ± 0.07 & 0.18 ± 0.00 & 1.94 ± 1.36 & 0.08 ± 0.02 \\
$\ \ \text{bw} = 0.2$ & 0.07 ± 0.01 & 0.55 ± 0.05 & 0.19 ± 0.00 & 3.78 ± 0.82 & 0.07 ± 0.02 \\
$\ \ \text{bw} = 0.25$ & 0.06 ± 0.01 & 0.54 ± 0.04 & 0.19 ± 0.00 & 6.10 ± 1.08 & 0.07 ± 0.02 \\
$\ \ \text{bw} = 0.3$ & 0.06 ± 0.01 & 0.45 ± 0.01 & 0.19 ± 0.00 & 6.70 ± 1.35 & 0.07 ± 0.02 \\
$\ \ \text{bw} = 0.35$ & 0.05 ± 0.01 & 0.44 ± 0.02 & 0.19 ± 0.00 & 9.81 ± 0.27 & 0.07 ± 0.02 \\
$\ \ \text{bw} = 0.4$ & 0.04 ± 0.01 & 0.42 ± 0.04 & 0.20 ± 0.00 & 12.14 ± 1.80 & 0.07 ± 0.02 \\
$\ \ \text{bw} = 0.45$ & 0.04 ± 0.01 & 0.42 ± 0.04 & 0.20 ± 0.00 & 14.51 ± 0.21 & 0.07 ± 0.02 \\
\bottomrule
\end{tabular}

\caption{CFX generation results, CaliforniaHousing, MLP.}
\end{table}

\begin{table}[H]

\begin{tabular}{llllll}
\toprule
Generator & Distance & Plausibility & Implausibility & Sensitivity $(10^{-1})$ & Stability \\
\midrule
\textbf{RandomForest} &  &  &  &  &  \\
MinDist & 0.01 ± 0.00 & 0.45 ± 0.03 & 0.20 ± 0.00 & 35.36 ± 15.39 & 0.23 ± 0.02 \\
ConfexNaive &  &  &  &  &  \\
$\ \ \alpha = 0.01$ & 0.03 ± 0.01 & 0.47 ± 0.05 & 0.20 ± 0.00 & 9.92 ± 2.08 & 0.24 ± 0.02 \\
$\ \ \alpha = 0.05$ & 0.03 ± 0.01 & 0.47 ± 0.05 & 0.20 ± 0.00 & 9.92 ± 2.08 & 0.24 ± 0.02 \\
$\ \ \alpha = 0.1$ & 0.03 ± 0.01 & 0.47 ± 0.05 & 0.20 ± 0.00 & 9.92 ± 2.08 & 0.24 ± 0.02 \\
ConfexTree, $\alpha = 0.01$ &  &  &  &  &  \\
$\ \ \text{bw} = 0.05$ & nan ± nan & nan ± nan & nan ± nan & nan ± nan & nan ± nan \\
$\ \ \text{bw} = 0.1$ & nan ± nan & nan ± nan & nan ± nan & nan ± nan & nan ± nan \\
$\ \ \text{bw} = 0.15$ & 0.60 ± 0.11 & 0.59 ± 0.29 & 0.15 ± 0.00 & 0.10 ± 0.03 & 0.26 ± 0.02 \\
$\ \ \text{bw} = 0.2$ & 0.15 ± 0.01 & 0.45 ± 0.05 & 0.16 ± 0.00 & 0.68 ± 0.19 & 0.24 ± 0.02 \\
$\ \ \text{bw} = 0.25$ & 0.11 ± 0.00 & 0.67 ± 0.03 & 0.16 ± 0.00 & 2.47 ± 0.02 & 0.24 ± 0.01 \\
$\ \ \text{bw} = 0.3$ & 0.10 ± 0.00 & 0.66 ± 0.02 & 0.17 ± 0.00 & 2.97 ± 0.17 & 0.24 ± 0.02 \\
$\ \ \text{bw} = 0.35$ & 0.07 ± 0.01 & 0.63 ± 0.05 & 0.18 ± 0.00 & 4.79 ± 0.89 & 0.24 ± 0.02 \\
$\ \ \text{bw} = 0.4$ & 0.09 ± 0.03 & 0.56 ± 0.06 & 0.18 ± 0.00 & 5.21 ± 0.18 & 0.23 ± 0.02 \\
$\ \ \text{bw} = 0.45$ & 0.09 ± 0.03 & 0.54 ± 0.06 & 0.18 ± 0.00 & 5.51 ± 0.32 & 0.23 ± 0.03 \\
ConfexTree, $\alpha = 0.05$ &  &  &  &  &  \\
$\ \ \text{bw} = 0.05$ & 0.45 ± 0.00 & 0.47 ± 0.03 & 0.16 ± 0.00 & 0.10 ± 0.06 & 0.26 ± 0.03 \\
$\ \ \text{bw} = 0.1$ & 0.16 ± 0.00 & 0.61 ± 0.09 & 0.17 ± 0.00 & 0.50 ± 0.12 & 0.23 ± 0.01 \\
$\ \ \text{bw} = 0.15$ & 0.11 ± 0.01 & 0.60 ± 0.06 & 0.17 ± 0.00 & 1.01 ± 0.46 & 0.23 ± 0.01 \\
$\ \ \text{bw} = 0.2$ & 0.07 ± 0.00 & 0.57 ± 0.09 & 0.18 ± 0.00 & 3.21 ± 2.37 & 0.23 ± 0.02 \\
$\ \ \text{bw} = 0.25$ & 0.06 ± 0.00 & 0.53 ± 0.03 & 0.19 ± 0.00 & 4.48 ± 0.79 & 0.23 ± 0.02 \\
$\ \ \text{bw} = 0.3$ & 0.06 ± 0.00 & 0.48 ± 0.04 & 0.19 ± 0.00 & 4.20 ± 0.32 & 0.23 ± 0.02 \\
$\ \ \text{bw} = 0.35$ & 0.05 ± 0.00 & 0.47 ± 0.01 & 0.19 ± 0.00 & 10.76 ± 0.74 & 0.23 ± 0.02 \\
$\ \ \text{bw} = 0.4$ & 0.04 ± 0.00 & 0.50 ± 0.06 & 0.19 ± 0.00 & 8.45 ± 1.94 & 0.23 ± 0.02 \\
$\ \ \text{bw} = 0.45$ & 0.04 ± 0.00 & 0.56 ± 0.08 & 0.19 ± 0.00 & 12.62 ± 2.07 & 0.23 ± 0.02 \\
ConfexTree, $\alpha = 0.1$ &  &  &  &  &  \\
$\ \ \text{bw} = 0.05$ & 0.18 ± 0.01 & 0.65 ± 0.03 & 0.17 ± 0.00 & 0.42 ± 0.16 & 0.24 ± 0.01 \\
$\ \ \text{bw} = 0.1$ & 0.10 ± 0.01 & 0.64 ± 0.02 & 0.18 ± 0.01 & 2.29 ± 1.65 & 0.23 ± 0.02 \\
$\ \ \text{bw} = 0.15$ & 0.08 ± 0.01 & 0.58 ± 0.02 & 0.18 ± 0.00 & 2.01 ± 1.26 & 0.23 ± 0.02 \\
$\ \ \text{bw} = 0.2$ & 0.06 ± 0.01 & 0.50 ± 0.06 & 0.19 ± 0.00 & 4.65 ± 3.14 & 0.23 ± 0.03 \\
$\ \ \text{bw} = 0.25$ & 0.05 ± 0.01 & 0.49 ± 0.05 & 0.19 ± 0.00 & 7.11 ± 1.87 & 0.23 ± 0.03 \\
$\ \ \text{bw} = 0.3$ & 0.04 ± 0.01 & 0.46 ± 0.04 & 0.19 ± 0.00 & 6.96 ± 1.08 & 0.23 ± 0.03 \\
$\ \ \text{bw} = 0.35$ & 0.04 ± 0.00 & 0.49 ± 0.03 & 0.19 ± 0.00 & 11.79 ± 1.67 & 0.23 ± 0.03 \\
$\ \ \text{bw} = 0.4$ & 0.04 ± 0.00 & 0.49 ± 0.03 & 0.19 ± 0.00 & 13.39 ± 1.87 & 0.23 ± 0.02 \\
$\ \ \text{bw} = 0.45$ & 0.03 ± 0.00 & 0.49 ± 0.05 & 0.19 ± 0.00 & 14.91 ± 2.71 & 0.23 ± 0.02 \\
FeatureTweak & 0.11 ± 0.01 & 0.46 ± 0.02 & 0.19 ± 0.01 & 0.53 ± 0.06 & 0.26 ± 0.02 \\
FOCUS & 0.09 ± 0.00 & 0.44 ± 0.06 & 0.20 ± 0.00 & 5.58 ± 1.73 & 0.26 ± 0.02 \\
\bottomrule
\end{tabular}

\caption{CFX generation results, CaliforniaHousing, RandomForest. Methods with nan values had 100\% failures. Validity 55\% for FeatureTweak.}
\end{table}

\subsubsection{Conformal evaluation results}
\begin{table}[H]
\begin{tabular}{lllll}
\toprule
Generator & Marginal CovGap & Binning CovGap & Class Cond CovGap & Simulated CovGap \\
\midrule
\textbf{MLP} &  &  &  &  \\
ConfexNaive &  &  &  &  \\
$\ \ \alpha = 0.01$ & 0.99 ± 0.01 & -0.59 ± 1.20 & -0.59 ± 1.15 & -9.96 ± 0.67 \\
$\ \ \alpha = 0.05$ & 0.96 ± 0.03 & -0.05 ± 3.53 & -0.01 ± 3.43 & -13.28 ± 7.90 \\
$\ \ \alpha = 0.1$ & 0.92 ± 0.03 & 0.06 ± 3.45 & 0.04 ± 3.32 & -18.90 ± 9.76 \\
ConfexTree, $\alpha = 0.01$ &  &  &  &  \\
$\ \ \text{bw} = 0.05$ & 1.00 ± 0.00 & 1.00 ± 0.00 & 1.00 ± 0.00 & nan ± nan \\
$\ \ \text{bw} = 0.1$ & 1.00 ± 0.00 & 1.00 ± 0.00 & 1.00 ± 0.00 & nan ± nan \\
$\ \ \text{bw} = 0.15$ & 1.00 ± 0.00 & 1.00 ± 0.00 & 1.00 ± 0.00 & nan ± nan \\
$\ \ \text{bw} = 0.2$ & 1.00 ± 0.00 & 0.91 ± 0.03 & 0.90 ± 0.03 & -1.37 ± 0.51 \\
$\ \ \text{bw} = 0.25$ & 1.00 ± 0.00 & 0.82 ± 0.07 & 0.81 ± 0.07 & -1.45 ± 1.44 \\
$\ \ \text{bw} = 0.3$ & 1.00 ± 0.00 & 0.75 ± 0.06 & 0.73 ± 0.07 & -2.91 ± 0.57 \\
$\ \ \text{bw} = 0.35$ & 1.00 ± 0.01 & 0.63 ± 0.12 & 0.61 ± 0.14 & -2.60 ± 1.56 \\
$\ \ \text{bw} = 0.4$ & 1.00 ± 0.00 & 0.37 ± 0.07 & 0.33 ± 0.09 & -3.91 ± 2.24 \\
$\ \ \text{bw} = 0.45$ & 1.00 ± 0.00 & 0.34 ± 0.07 & 0.31 ± 0.09 & -3.92 ± 2.26 \\
ConfexTree, $\alpha = 0.05$ &  &  &  &  \\
$\ \ \text{bw} = 0.05$ & 1.00 ± 0.00 & 5.00 ± 0.00 & 5.00 ± 0.00 & 5.00 ± 0.00 \\
$\ \ \text{bw} = 0.1$ & 1.00 ± 0.00 & 4.91 ± 0.00 & 4.90 ± 0.00 & -2.23 ± 0.39 \\
$\ \ \text{bw} = 0.15$ & 1.00 ± 0.00 & 4.80 ± 0.05 & 4.79 ± 0.05 & -10.48 ± 1.51 \\
$\ \ \text{bw} = 0.2$ & 1.00 ± 0.00 & 4.63 ± 0.02 & 4.62 ± 0.02 & -13.45 ± 1.01 \\
$\ \ \text{bw} = 0.25$ & 0.99 ± 0.00 & 4.28 ± 0.06 & 4.26 ± 0.05 & -9.90 ± 5.07 \\
$\ \ \text{bw} = 0.3$ & 0.99 ± 0.00 & 3.78 ± 0.20 & 3.73 ± 0.19 & -14.67 ± 4.52 \\
$\ \ \text{bw} = 0.35$ & 0.96 ± 0.01 & 1.69 ± 0.19 & 1.54 ± 0.14 & -13.06 ± 4.16 \\
$\ \ \text{bw} = 0.4$ & 0.95 ± 0.00 & 0.40 ± 0.16 & 0.18 ± 0.10 & -12.33 ± 0.67 \\
$\ \ \text{bw} = 0.45$ & 0.95 ± 0.00 & 0.38 ± 0.16 & 0.15 ± 0.10 & -12.36 ± 0.67 \\
ConfexTree, $\alpha = 0.1$ &  &  &  &  \\
$\ \ \text{bw} = 0.05$ & 1.00 ± 0.00 & 10.00 ± 0.00 & 10.00 ± 0.00 & 10.00 ± 0.00 \\
$\ \ \text{bw} = 0.1$ & 1.00 ± 0.00 & 9.83 ± 0.02 & 9.82 ± 0.02 & 3.85 ± 0.00 \\
$\ \ \text{bw} = 0.15$ & 1.00 ± 0.00 & 9.72 ± 0.07 & 9.71 ± 0.07 & -6.19 ± 2.52 \\
$\ \ \text{bw} = 0.2$ & 0.99 ± 0.00 & 9.03 ± 0.09 & 9.01 ± 0.07 & -20.73 ± 4.27 \\
$\ \ \text{bw} = 0.25$ & 0.98 ± 0.00 & 8.31 ± 0.33 & 8.26 ± 0.31 & -12.70 ± 1.10 \\
$\ \ \text{bw} = 0.3$ & 0.98 ± 0.00 & 7.36 ± 0.56 & 7.27 ± 0.53 & -14.93 ± 5.00 \\
$\ \ \text{bw} = 0.35$ & 0.92 ± 0.00 & 2.33 ± 0.12 & 1.98 ± 0.21 & -16.89 ± 1.16 \\
$\ \ \text{bw} = 0.4$ & 0.91 ± 0.00 & 1.52 ± 0.07 & 1.15 ± 0.02 & -17.24 ± 2.86 \\
$\ \ \text{bw} = 0.45$ & 0.91 ± 0.00 & 1.49 ± 0.09 & 1.11 ± 0.00 & -17.22 ± 2.93 \\
\bottomrule
\end{tabular}
\caption{Conformal evaluation results, CaliforniaHousing, MLP}
\end{table}

\begin{table}[H]
\begin{tabular}{lllll}
\toprule
Generator & Marginal CovGap & Binning CovGap & Class Cond CovGap & Simulated CovGap \\
\midrule
\textbf{RandomForest} &  &  &  &  \\
ConfexNaive &  &  &  &  \\
$\ \ \alpha = 0.01$ & 1.00 ± 0.00 & 1.00 ± 0.00 & 1.00 ± 0.00 & nan ± nan \\
$\ \ \alpha = 0.05$ & 0.94 ± 0.01 & -0.24 ± 0.13 & -0.52 ± 0.14 & -42.48 ± 0.00 \\
$\ \ \alpha = 0.1$ & 0.94 ± 0.01 & 4.76 ± 0.13 & 4.48 ± 0.14 & -37.48 ± 0.00 \\
ConfexTree, $\alpha = 0.01$ &  &  &  &  \\
$\ \ \text{bw} = 0.05$ & 1.00 ± 0.00 & 1.00 ± 0.00 & 1.00 ± 0.00 & nan ± nan \\
$\ \ \text{bw} = 0.1$ & 1.00 ± 0.00 & 1.00 ± 0.00 & 1.00 ± 0.00 & nan ± nan \\
$\ \ \text{bw} = 0.15$ & 1.00 ± 0.00 & 1.00 ± 0.00 & 1.00 ± 0.00 & nan ± nan \\
$\ \ \text{bw} = 0.2$ & 1.00 ± 0.00 & 0.99 ± 0.02 & 0.99 ± 0.02 & 0.99 ± 0.02 \\
$\ \ \text{bw} = 0.25$ & 1.00 ± 0.00 & 0.98 ± 0.03 & 0.98 ± 0.03 & 0.96 ± 0.05 \\
$\ \ \text{bw} = 0.3$ & 1.00 ± 0.00 & 0.98 ± 0.03 & 0.98 ± 0.03 & 0.96 ± 0.05 \\
$\ \ \text{bw} = 0.35$ & 1.00 ± 0.00 & 0.98 ± 0.03 & 0.98 ± 0.03 & -5.73 ± nan \\
$\ \ \text{bw} = 0.4$ & 1.00 ± 0.00 & 0.81 ± 0.27 & 0.81 ± 0.27 & -55.44 ± nan \\
$\ \ \text{bw} = 0.45$ & 1.00 ± 0.00 & 0.81 ± 0.27 & 0.81 ± 0.27 & -50.33 ± nan \\
ConfexTree, $\alpha = 0.05$ &  &  &  &  \\
$\ \ \text{bw} = 0.05$ & 1.00 ± 0.00 & 5.00 ± 0.00 & 5.00 ± 0.00 & 5.00 ± 0.00 \\
$\ \ \text{bw} = 0.1$ & 1.00 ± 0.00 & 4.93 ± 0.07 & 4.93 ± 0.07 & 2.83 ± 3.03 \\
$\ \ \text{bw} = 0.15$ & 1.00 ± 0.00 & 4.90 ± 0.02 & 4.89 ± 0.02 & -11.57 ± 4.90 \\
$\ \ \text{bw} = 0.2$ & 1.00 ± 0.00 & 4.77 ± 0.13 & 4.76 ± 0.14 & -27.61 ± 22.13 \\
$\ \ \text{bw} = 0.25$ & 1.00 ± 0.00 & 4.44 ± 0.16 & 4.42 ± 0.17 & -22.39 ± 1.28 \\
$\ \ \text{bw} = 0.3$ & 0.99 ± 0.01 & 3.89 ± 0.32 & 3.86 ± 0.34 & -35.73 ± 8.51 \\
$\ \ \text{bw} = 0.35$ & 0.97 ± 0.01 & 2.60 ± 0.37 & 2.50 ± 0.38 & -37.38 ± 3.72 \\
$\ \ \text{bw} = 0.4$ & 0.97 ± 0.01 & 2.56 ± 0.20 & 2.46 ± 0.21 & -34.46 ± 1.40 \\
$\ \ \text{bw} = 0.45$ & 0.97 ± 0.01 & 2.55 ± 0.22 & 2.44 ± 0.22 & -34.39 ± 1.44 \\
ConfexTree, $\alpha = 0.1$ &  &  &  &  \\
$\ \ \text{bw} = 0.05$ & 1.00 ± 0.00 & 10.00 ± 0.00 & 10.00 ± 0.00 & 10.00 ± 0.00 \\
$\ \ \text{bw} = 0.1$ & 1.00 ± 0.00 & 9.89 ± 0.06 & 9.89 ± 0.05 & 5.43 ± 0.94 \\
$\ \ \text{bw} = 0.15$ & 1.00 ± 0.00 & 9.67 ± 0.27 & 9.65 ± 0.29 & -8.27 ± 13.19 \\
$\ \ \text{bw} = 0.2$ & 1.00 ± 0.00 & 9.38 ± 0.07 & 9.37 ± 0.07 & -33.23 ± 1.04 \\
$\ \ \text{bw} = 0.25$ & 0.99 ± 0.01 & 8.95 ± 0.09 & 8.93 ± 0.10 & -33.76 ± 4.37 \\
$\ \ \text{bw} = 0.3$ & 0.99 ± 0.01 & 8.09 ± 0.28 & 8.06 ± 0.31 & -39.29 ± 6.97 \\
$\ \ \text{bw} = 0.35$ & 0.94 ± 0.00 & 4.52 ± 0.17 & 4.32 ± 0.15 & -42.42 ± 3.56 \\
$\ \ \text{bw} = 0.4$ & 0.94 ± 0.01 & 3.93 ± 0.02 & 3.71 ± 0.02 & -44.97 ± 1.06 \\
$\ \ \text{bw} = 0.45$ & 0.94 ± 0.01 & 3.77 ± 0.14 & 3.54 ± 0.15 & -44.84 ± 0.96 \\
\bottomrule
\end{tabular}
\caption{Conformal evaluation results, CaliforniaHousing, RandomForest}
\end{table}

\newpage
\subsection{German Credit} We use the German Credit dataset from the UCI Machine Learning Repository~\cite{hofmann1994statlog}, with a cleaned version obtained through Kaggle\footnote{\url{https://www.kaggle.com/datasets/uciml/german-credit/data}}. The preprocessing included: (i) scaling numeric features (Age, Credit amount, Duration) to $(0, 1)$ using MinMax scaling, (ii) ordinal encoding of categorical features (job, savings account, checking account), then normalised. The Purpose feature was dropped.

\subsubsection{Model evaluation results}

\begin{table}[ht!]
\centering
\renewcommand{\arraystretch}{1.2}
\setlength{\tabcolsep}{8pt}
\begin{tabular}{lcccc}
\toprule
Repeat  & Accuracy (\%) & Precision (\%) & F1 Score (\%) \\
\midrule
repeat0,RF  & 70.00 & 68.27 & 68.77 \\
repeat1,RF  & 69.50 & 68.31 & 68.76 \\
repeat0,MLP & 72.00 & 72.00 & 72.00 \\
repeat1,MLP & 71.00 & 70.01 & 70.39 \\
\bottomrule
\end{tabular}
\caption{Model evaluation results, GermanCredit.}
\end{table}

\subsubsection{Plots}

\begin{figure}[H]
\centering
\begin{subfigure}{0.315\linewidth}
    \includegraphics[width=\linewidth]{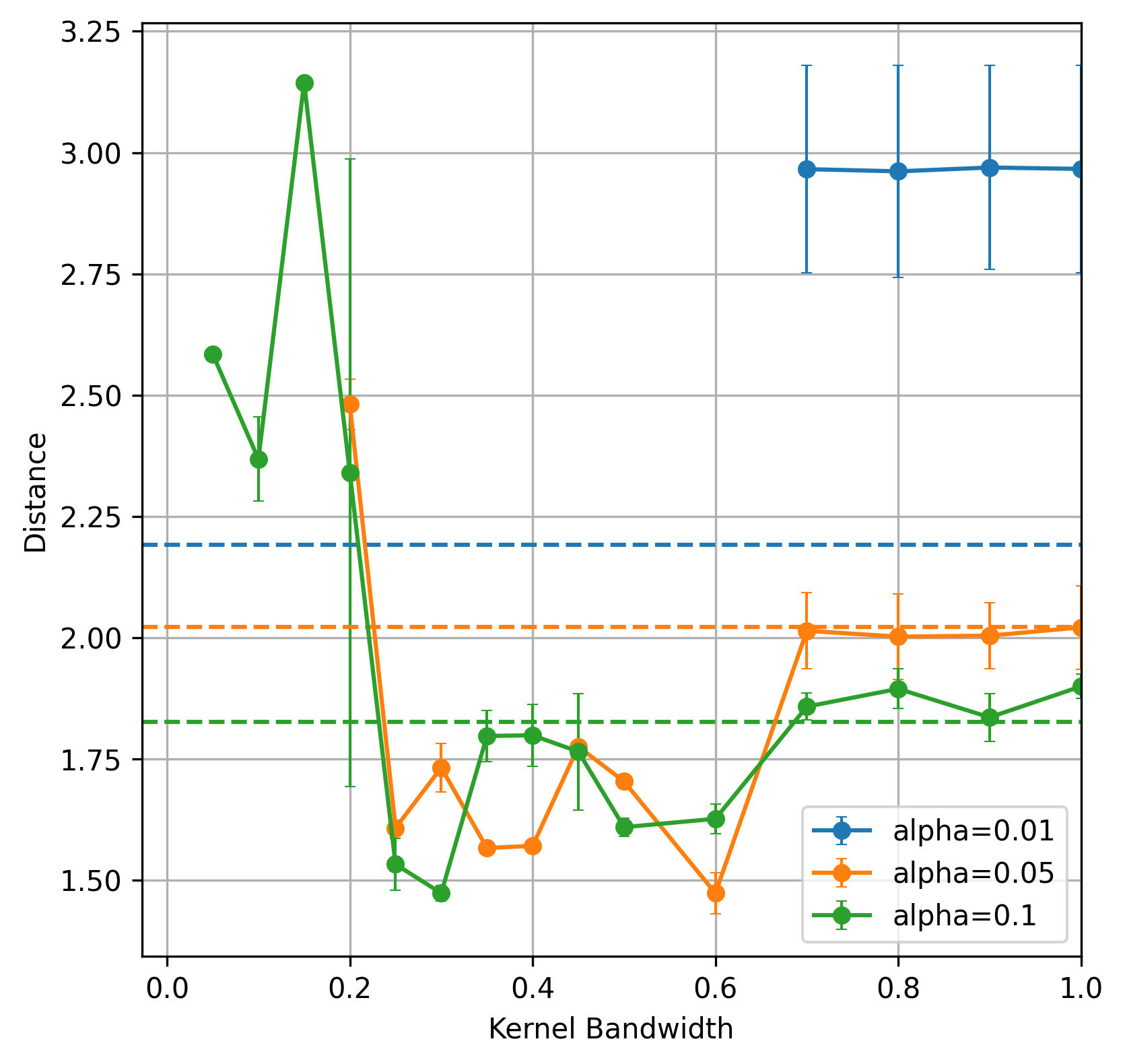}
    \caption{Distance}
\end{subfigure}
\hfill
\begin{subfigure}{0.33\linewidth}
    \includegraphics[width=\linewidth]{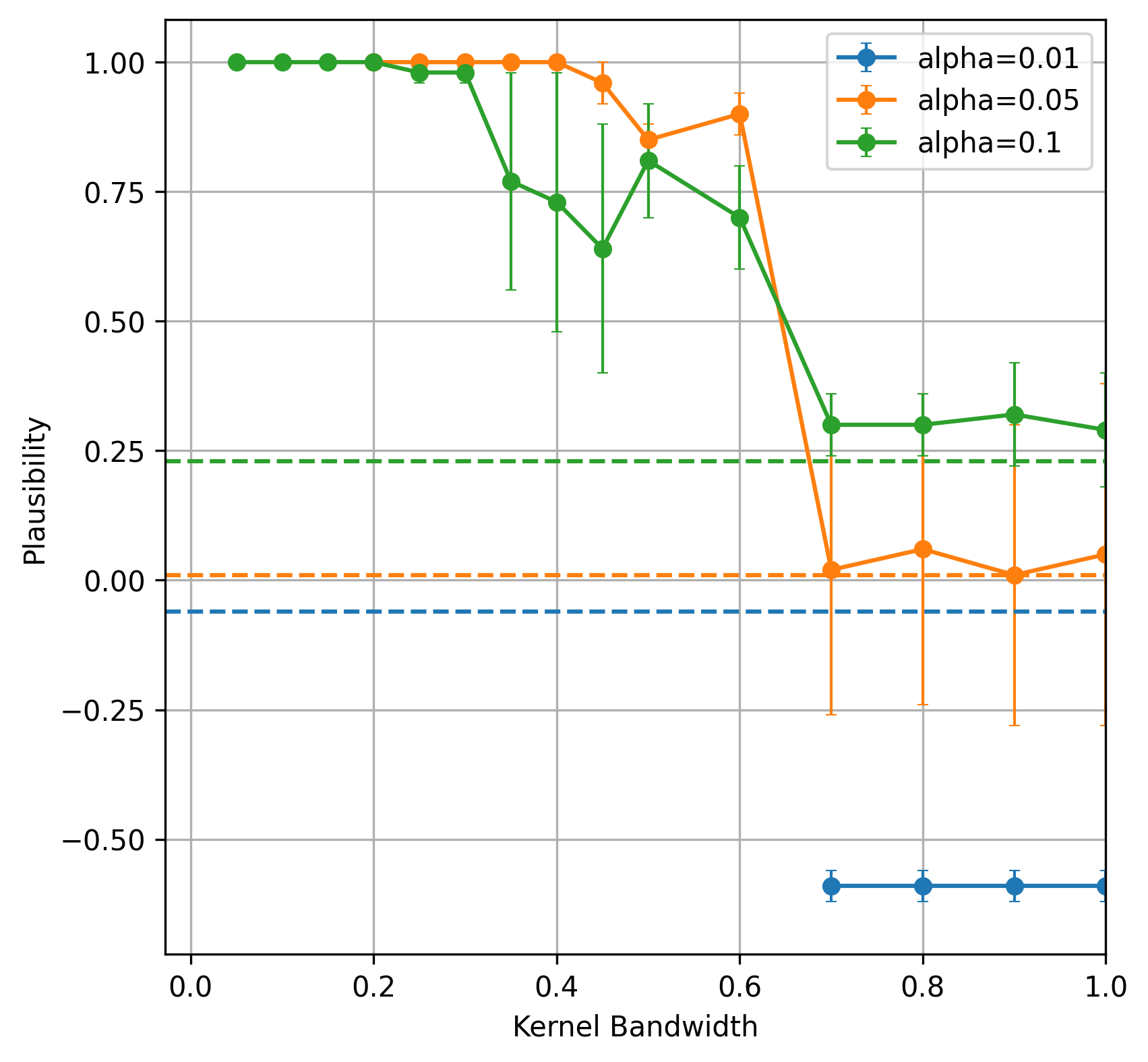}
    \caption{Plausibility}
\end{subfigure}
\hfill
\begin{subfigure}{0.32\linewidth}
    \includegraphics[width=\linewidth]{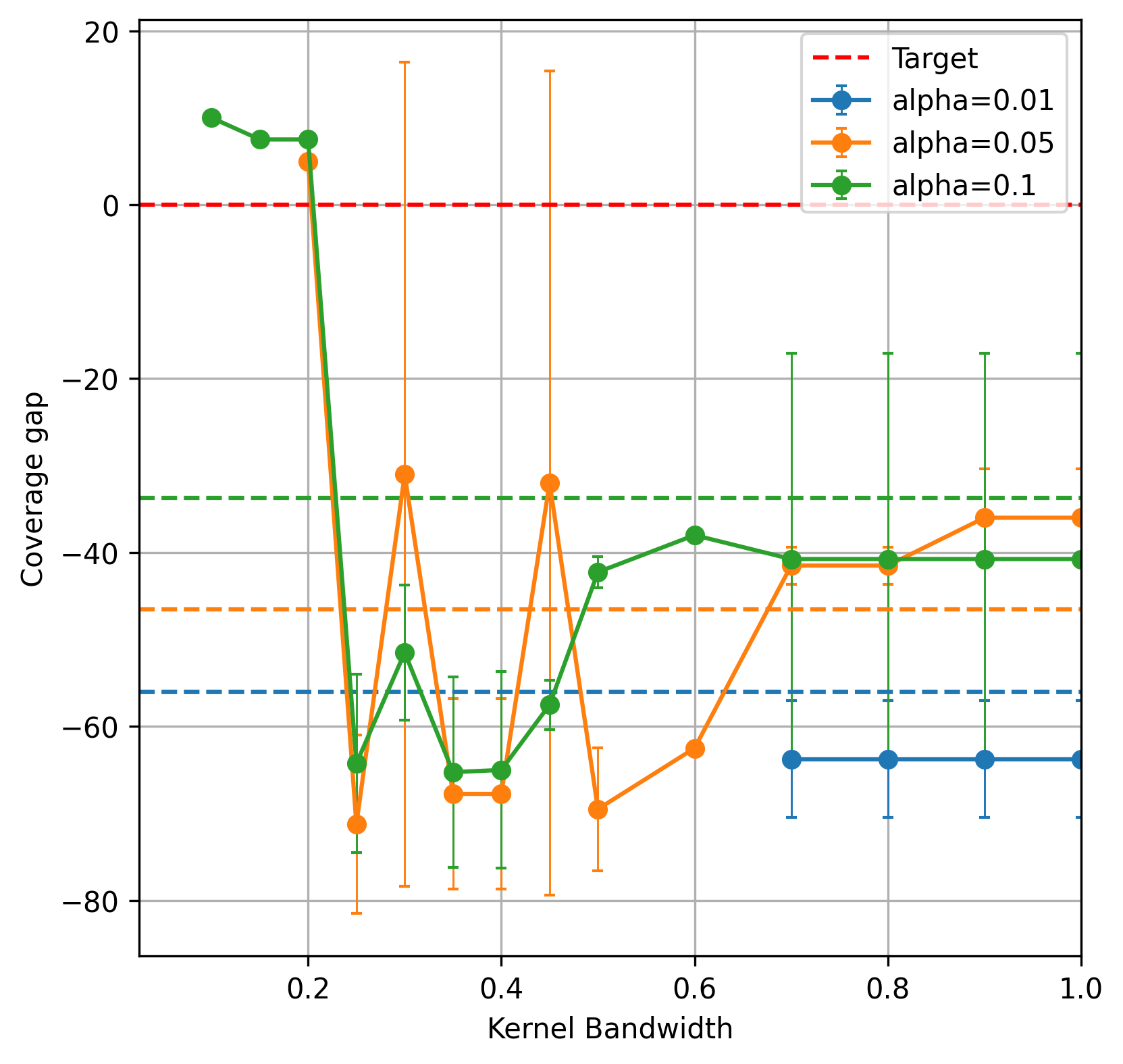}
    \caption{Coverage Gap}
\end{subfigure}
\caption{Effect of coverage rate and kernel bandwidth on metrics for CONFEX-Tree on the GermanCredit dataset, MLP. \confexnaive\ is represented by dashed horizontal lines.}
\label{Figure 2}
\end{figure}

\begin{figure}[H]
\centering
\begin{subfigure}{0.315\linewidth}
    \includegraphics[width=\linewidth]{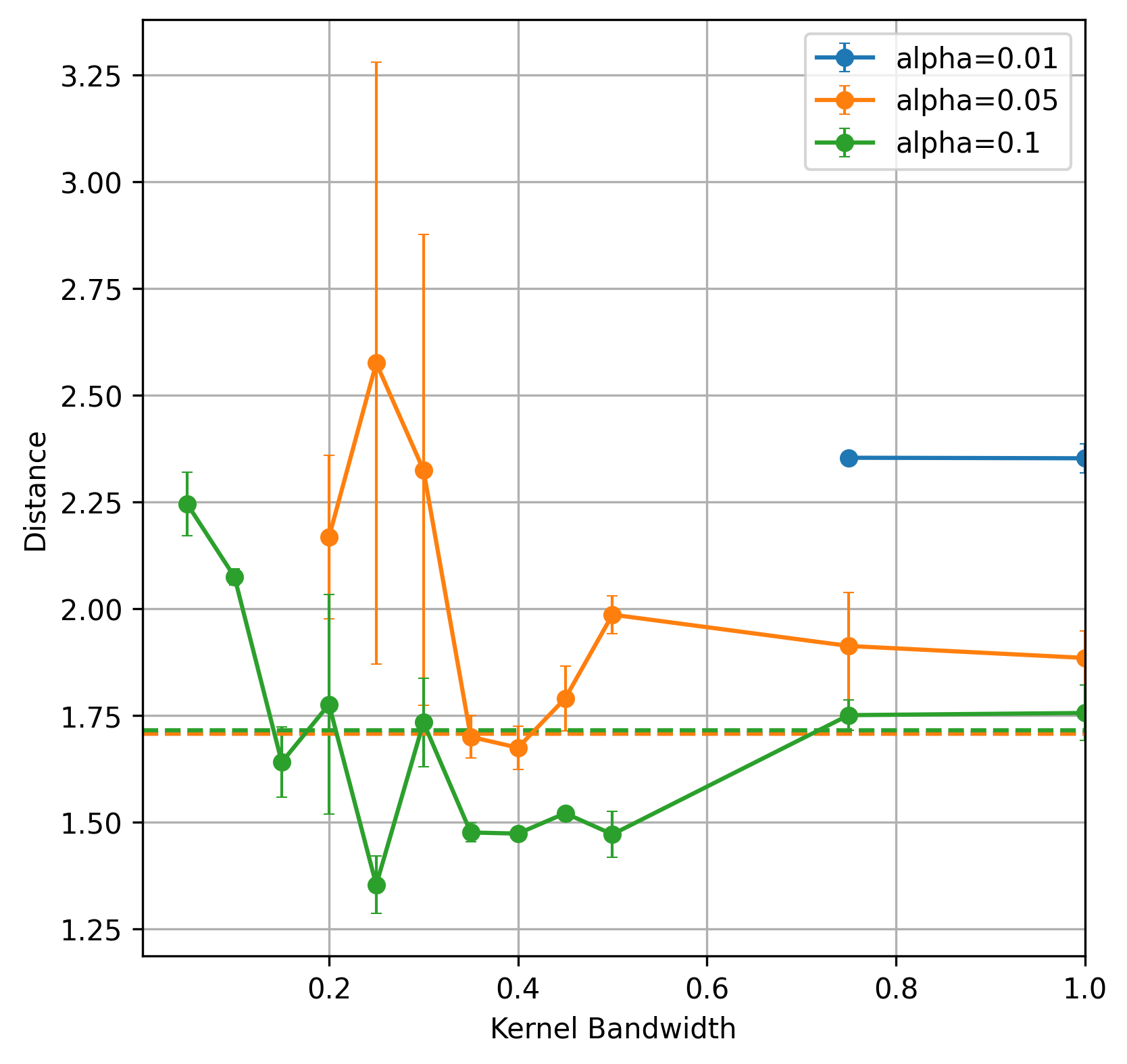}
    \caption{Distance}
\end{subfigure}
\hfill
\begin{subfigure}{0.33\linewidth}
    \includegraphics[width=\linewidth]{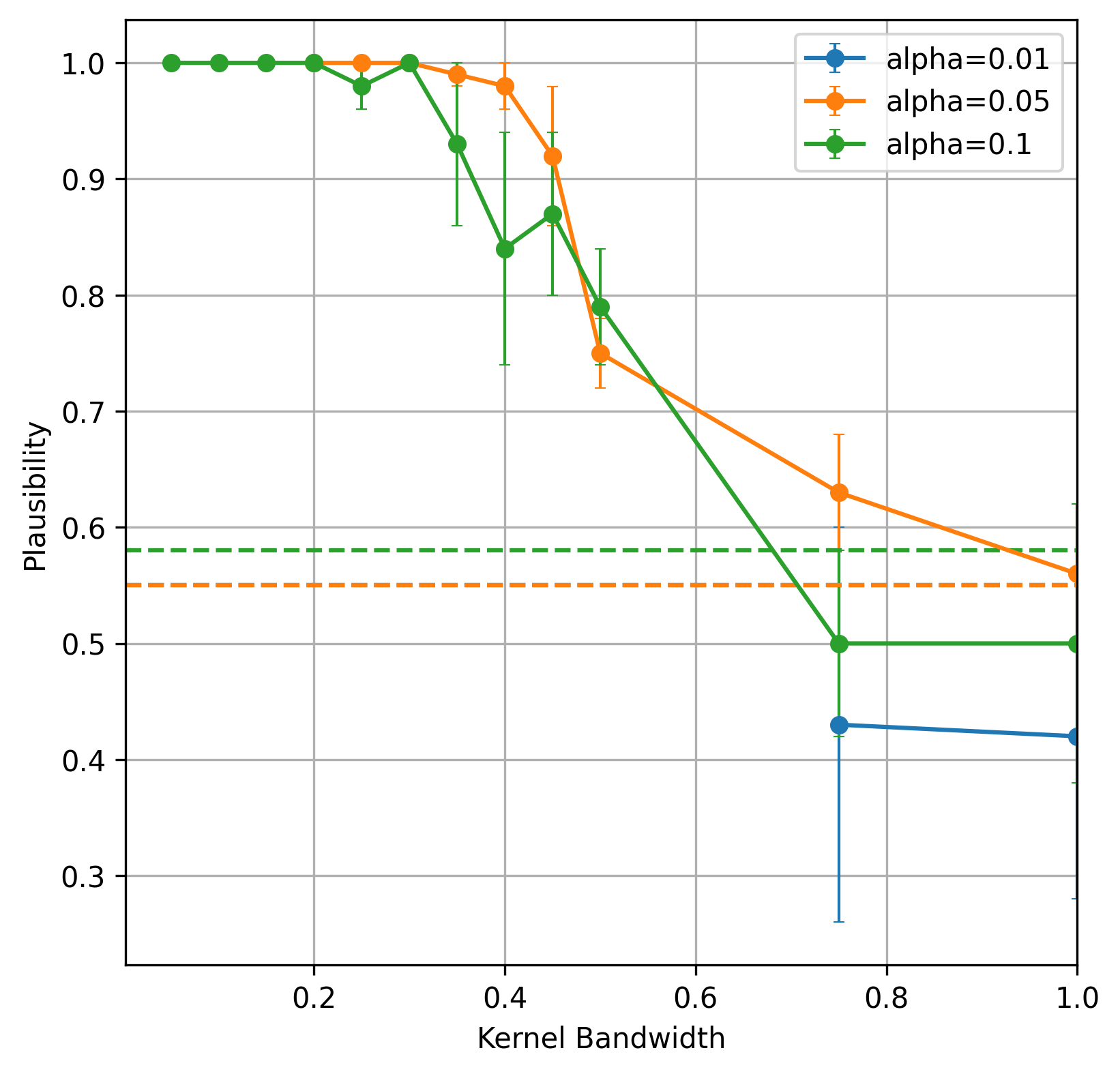}
    \caption{Plausibility}
\end{subfigure}
\hfill
\begin{subfigure}{0.32\linewidth}
    \includegraphics[width=\linewidth]{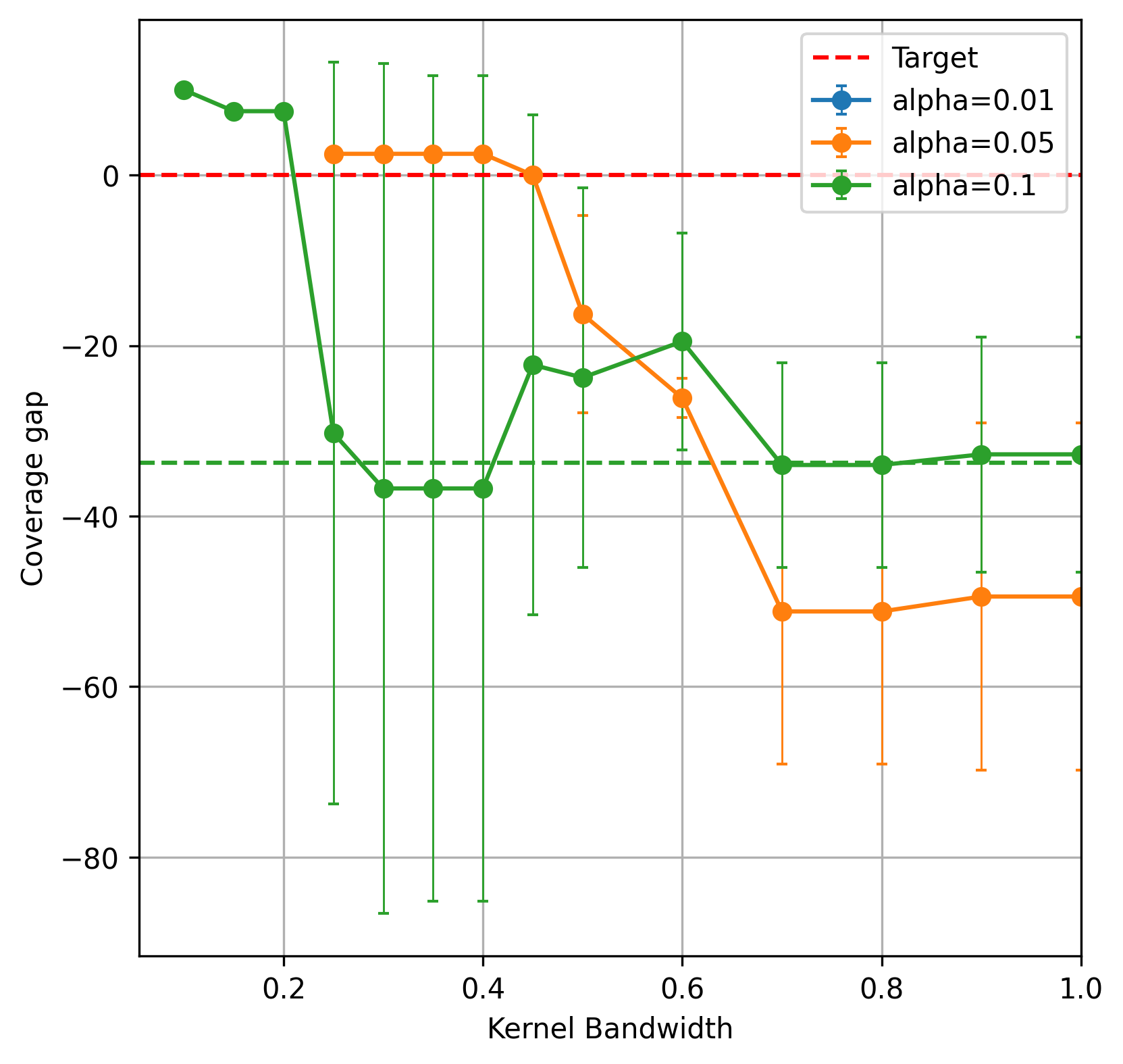}
    \caption{Coverage Gap}
\end{subfigure}
\caption{Effect of coverage rate and kernel bandwidth on metrics for CONFEX-Tree on the GermanCredit dataset, RandomForest. \confexnaive\ is represented by dashed horizontal lines.}
\label{Figure 2}
\end{figure}

\subsubsection{CFX generation results}

\begin{table}[H]

\begin{tabular}{llllll}
\toprule
Generator & Distance & Plausibility & Implausibility & Sensitivity $(10^{-1})$ & Stability \\
\midrule
\textbf{MLP} &  &  &  &  &  \\
MinDist & 1.69 ± 0.04 & 0.50 ± 0.06 & 0.73 ± 0.01 & 0.09 ± 0.01 & 0.58 ± 0.01 \\
Wachter & 0.40 ± 0.01 & 0.77 ± 0.03 & 0.59 ± 0.00 & 0.25 ± 0.00 & 0.22 ± 0.01 \\
Greedy & 0.98 ± 0.05 & -0.02 ± 0.04 & 0.80 ± 0.02 & 0.09 ± 0.00 & 0.67 ± 0.03 \\
ConfexNaive &  &  &  &  &  \\
$\ \ \alpha = 0.01$ & 2.19 ± 0.10 & -0.06 ± 0.24 & 0.83 ± 0.02 & 0.03 ± 0.00 & 0.96 ± 0.03 \\
$\ \ \alpha = 0.05$ & 2.02 ± 0.00 & 0.01 ± 0.21 & 0.79 ± 0.02 & 0.06 ± 0.01 & 0.84 ± 0.06 \\
$\ \ \alpha = 0.1$ & 1.83 ± 0.05 & 0.23 ± 0.15 & 0.75 ± 0.02 & 0.08 ± 0.03 & 0.70 ± 0.03 \\
ECCCo &  &  &  &  &  \\
$\ \ \alpha = 0.01$ & 1.01 ± 0.03 & 0.16 ± 0.04 & 0.78 ± 0.02 & 0.05 ± 0.02 & 0.73 ± 0.01 \\
$\ \ \alpha = 0.05$ & 0.98 ± 0.03 & 0.14 ± 0.10 & 0.77 ± 0.02 & 0.05 ± 0.02 & 0.72 ± 0.02 \\
$\ \ \alpha = 0.1$ & 0.94 ± 0.01 & 0.21 ± 0.05 & 0.75 ± 0.01 & 0.05 ± 0.02 & 0.72 ± 0.02 \\ 
ConfexTree, $\alpha = 0.01$ &  &  &  &  &  \\
$\ \ \text{bw} = 0.05$ & nan ± nan & nan ± nan & nan ± nan & nan ± nan & nan ± nan \\
$\ \ \text{bw} = 0.6$ & nan ± nan & nan ± nan & nan ± nan & nan ± nan & nan ± nan \\
$\ \ \text{bw} = 0.7$ & 2.97 ± 0.21 & -0.59 ± 0.03 & 0.84 ± 0.07 & 0.03 ± 0.01 & 0.98 ± 0.00 \\
$\ \ \text{bw} = 0.8$ & 2.96 ± 0.22 & -0.59 ± 0.03 & 0.84 ± 0.08 & 0.02 ± 0.01 & 0.98 ± 0.00 \\
$\ \ \text{bw} = 0.9$ & 2.97 ± 0.21 & -0.59 ± 0.03 & 0.84 ± 0.07 & 0.02 ± 0.01 & 0.98 ± 0.00 \\
$\ \ \text{bw} = 1$ & 2.97 ± 0.21 & -0.59 ± 0.03 & 0.84 ± 0.07 & 0.02 ± 0.01 & 0.98 ± 0.00 \\
$\ \ \text{bw} = 1.2$ & 2.97 ± 0.21 & -0.60 ± 0.04 & 0.84 ± 0.07 & 0.03 ± 0.01 & 0.98 ± 0.00 \\
$\ \ \text{bw} = 1.4$ & 2.97 ± 0.21 & -0.60 ± 0.04 & 0.84 ± 0.07 & 0.03 ± 0.01 & 0.98 ± 0.00 \\
$\ \ \text{bw} = 1.6$ & 2.97 ± 0.21 & -0.60 ± 0.04 & 0.84 ± 0.07 & 0.03 ± 0.01 & 0.98 ± 0.00 \\
ConfexTree, $\alpha = 0.05$ &  &  &  &  &  \\
$\ \ \text{bw} = 0.05$ & nan ± nan & nan ± nan & nan ± nan & nan ± nan & nan ± nan \\
$\ \ \text{bw} = 0.15$ & nan ± nan & nan ± nan & nan ± nan & nan ± nan & nan ± nan \\
$\ \ \text{bw} = 0.2$ & 2.48 ± 0.05 & 1.00 ± 0.00 & 0.27 ± 0.02 & 0.01 ± 0.01 & 0.89 ± 0.00 \\
$\ \ \text{bw} = 0.3$ & 1.73 ± 0.05 & 1.00 ± 0.00 & 0.43 ± 0.00 & 0.01 ± 0.00 & 0.80 ± 0.00 \\
$\ \ \text{bw} = 0.4$ & 1.57 ± 0.01 & 1.00 ± 0.00 & 0.43 ± 0.03 & 0.03 ± 0.01 & 0.65 ± 0.01 \\
$\ \ \text{bw} = 0.5$ & 1.70 ± 0.01 & 0.85 ± 0.03 & 0.59 ± 0.00 & 0.06 ± 0.00 & 0.58 ± 0.00 \\
$\ \ \text{bw} = 0.6$ & 1.47 ± 0.04 & 0.90 ± 0.04 & 0.61 ± 0.01 & 0.06 ± 0.01 & 0.62 ± 0.01 \\
$\ \ \text{bw} = 0.7$ & 2.01 ± 0.08 & 0.02 ± 0.28 & 0.79 ± 0.04 & 0.06 ± 0.00 & 0.82 ± 0.06 \\
$\ \ \text{bw} = 0.8$ & 2.00 ± 0.09 & 0.06 ± 0.30 & 0.79 ± 0.05 & 0.06 ± 0.00 & 0.82 ± 0.06 \\
$\ \ \text{bw} = 1$ & 2.02 ± 0.09 & 0.05 ± 0.33 & 0.79 ± 0.05 & 0.05 ± 0.00 & 0.81 ± 0.07 \\
$\ \ \text{bw} = 1.2$ & 2.00 ± 0.08 & 0.04 ± 0.32 & 0.79 ± 0.05 & 0.05 ± 0.00 & 0.81 ± 0.07 \\
$\ \ \text{bw} = 1.4$ & 2.01 ± 0.08 & 0.05 ± 0.31 & 0.80 ± 0.05 & 0.05 ± 0.00 & 0.81 ± 0.06 \\
$\ \ \text{bw} = 1.6$ & 2.01 ± 0.08 & 0.05 ± 0.31 & 0.80 ± 0.05 & 0.05 ± 0.00 & 0.81 ± 0.06 \\
ConfexTree, $\alpha = 0.1$ &  &  &  &  &  \\
$\ \ \text{bw} = 0.05$ & 2.58 ± 0.00 & 1.00 ± 0.00 & 0.34 ± 0.00 & 0.00 ± 0.00 & 0.46 ± 0.02 \\
$\ \ \text{bw} = 0.1$ & 2.37 ± 0.09 & 1.00 ± 0.00 & 0.24 ± 0.01 & 0.01 ± 0.00 & 0.83 ± 0.03 \\
$\ \ \text{bw} = 0.2$ & 2.34 ± 0.65 & 1.00 ± 0.00 & 0.38 ± 0.01 & 0.01 ± 0.00 & 0.62 ± 0.00 \\
$\ \ \text{bw} = 0.4$ & 1.80 ± 0.06 & 0.73 ± 0.25 & 0.67 ± 0.01 & 0.06 ± 0.02 & 0.57 ± 0.08 \\
$\ \ \text{bw} = 0.6$ & 1.63 ± 0.03 & 0.70 ± 0.10 & 0.65 ± 0.01 & 0.08 ± 0.00 & 0.62 ± 0.05 \\
$\ \ \text{bw} = 0.8$ & 1.89 ± 0.04 & 0.30 ± 0.06 & 0.75 ± 0.01 & 0.06 ± 0.02 & 0.73 ± 0.04 \\
$\ \ \text{bw} = 1$ & 1.90 ± 0.03 & 0.29 ± 0.11 & 0.75 ± 0.01 & 0.07 ± 0.02 & 0.73 ± 0.05 \\
$\ \ \text{bw} = 1.2$ & 1.86 ± 0.03 & 0.31 ± 0.09 & 0.74 ± 0.01 & 0.07 ± 0.01 & 0.73 ± 0.04 \\
$\ \ \text{bw} = 1.4$ & 1.86 ± 0.03 & 0.28 ± 0.10 & 0.75 ± 0.02 & 0.07 ± 0.02 & 0.73 ± 0.05 \\
$\ \ \text{bw} = 1.6$ & 1.86 ± 0.03 & 0.28 ± 0.10 & 0.75 ± 0.02 & 0.07 ± 0.02 & 0.73 ± 0.05 \\
\bottomrule
\end{tabular}

\caption{CFX generation results, GermanCredit, MLP. Methods with nan values had 100\% failures. Validity 84\% for all ECCCo methods, 78.5\% for Greedy, 84\% for Wachter.}
\end{table}

\begin{table}[H]

\begin{tabular}{llllll}
\toprule
Generator & Distance & Plausibility & Implausibility & Sensitivity $(10^{-1})$ & Stability \\
\midrule
\textbf{RandomForest} &  &  &  &  &  \\
MinDist & 1.69 ± 0.05 & 0.34 ± 0.12 & 0.77 ± 0.02 & 0.09 ± 0.01 & 0.37 ± 0.00 \\
ConfexNaive &  &  &  &  &  \\
$\ \ \alpha = 0.01$ & 1.71 ± 0.04 & 0.55 ± 0.11 & 0.71 ± 0.01 & 0.09 ± 0.02 & 0.45 ± 0.01 \\
$\ \ \alpha = 0.05$ & 1.71 ± 0.04 & 0.55 ± 0.11 & 0.71 ± 0.01 & 0.09 ± 0.02 & 0.45 ± 0.01 \\
$\ \ \alpha = 0.1$ & 1.72 ± 0.04 & 0.58 ± 0.08 & 0.71 ± 0.00 & 0.09 ± 0.02 & 0.45 ± 0.01 \\
ConfexTree, $\alpha = 0.01$ &  &  &  &  &  \\
$\ \ \text{bw} = 0.05$ & nan ± nan & nan ± nan & nan ± nan & nan ± nan & nan ± nan \\
$\ \ \text{bw} = 0.1$ & nan ± nan & nan ± nan & nan ± nan & nan ± nan & nan ± nan \\
$\ \ \text{bw} = 0.15$ & nan ± nan & nan ± nan & nan ± nan & nan ± nan & nan ± nan \\
$\ \ \text{bw} = 0.2$ & nan ± nan & nan ± nan & nan ± nan & nan ± nan & nan ± nan \\
$\ \ \text{bw} = 0.25$ & nan ± nan & nan ± nan & nan ± nan & nan ± nan & nan ± nan \\
$\ \ \text{bw} = 0.3$ & nan ± nan & nan ± nan & nan ± nan & nan ± nan & nan ± nan \\
$\ \ \text{bw} = 0.35$ & nan ± nan & nan ± nan & nan ± nan & nan ± nan & nan ± nan \\
$\ \ \text{bw} = 0.4$ & nan ± nan & nan ± nan & nan ± nan & nan ± nan & nan ± nan \\
$\ \ \text{bw} = 0.45$ & nan ± nan & nan ± nan & nan ± nan & nan ± nan & nan ± nan \\
$\ \ \text{bw} = 0.5$ & nan ± nan & nan ± nan & nan ± nan & nan ± nan & nan ± nan \\
$\ \ \text{bw} = 0.75$ & 2.35 ± 0.01 & 0.43 ± 0.17 & 0.65 ± 0.01 & 0.07 ± 0.02 & 0.44 ± 0.01 \\
$\ \ \text{bw} = 1$ & 2.35 ± 0.03 & 0.42 ± 0.14 & 0.65 ± 0.02 & 0.08 ± 0.02 & 0.44 ± 0.01 \\
ConfexTree, $\alpha = 0.05$ &  &  &  &  &  \\
$\ \ \text{bw} = 0.05$ & nan ± nan & nan ± nan & nan ± nan & nan ± nan & nan ± nan \\
$\ \ \text{bw} = 0.1$ & nan ± nan & nan ± nan & nan ± nan & nan ± nan & nan ± nan \\
$\ \ \text{bw} = 0.15$ & nan ± nan & nan ± nan & nan ± nan & nan ± nan & nan ± nan \\
$\ \ \text{bw} = 0.2$ & 2.17 ± 0.19 & 1.00 ± 0.00 & 0.34 ± 0.06 & 0.02 ± 0.00 & 0.47 ± 0.27 \\
$\ \ \text{bw} = 0.25$ & 2.58 ± 0.70 & 1.00 ± 0.00 & 0.43 ± 0.09 & 0.28 ± 0.26 & 0.30 ± 0.01 \\
$\ \ \text{bw} = 0.3$ & 2.33 ± 0.55 & 1.00 ± 0.00 & 0.41 ± 0.09 & 0.06 ± 0.01 & 0.32 ± 0.01 \\
$\ \ \text{bw} = 0.35$ & 1.70 ± 0.05 & 0.99 ± 0.01 & 0.50 ± 0.07 & 0.15 ± 0.10 & 0.31 ± 0.00 \\
$\ \ \text{bw} = 0.4$ & 1.67 ± 0.05 & 0.98 ± 0.02 & 0.50 ± 0.08 & 0.10 ± 0.06 & 0.31 ± 0.00 \\
$\ \ \text{bw} = 0.45$ & 1.79 ± 0.08 & 0.92 ± 0.06 & 0.48 ± 0.07 & 0.07 ± 0.02 & 0.36 ± 0.01 \\
$\ \ \text{bw} = 0.5$ & 1.99 ± 0.04 & 0.75 ± 0.03 & 0.55 ± 0.03 & 0.05 ± 0.00 & 0.39 ± 0.01 \\
$\ \ \text{bw} = 0.75$ & 1.91 ± 0.13 & 0.63 ± 0.05 & 0.69 ± 0.01 & 0.09 ± 0.00 & 0.50 ± 0.03 \\
$\ \ \text{bw} = 1$ & 1.88 ± 0.06 & 0.56 ± 0.06 & 0.73 ± 0.01 & 0.09 ± 0.02 & 0.47 ± 0.02 \\
ConfexTree, $\alpha = 0.1$ &  &  &  &  &  \\
$\ \ \text{bw} = 0.05$ & 2.25 ± 0.07 & 1.00 ± 0.00 & 0.35 ± 0.01 & 0.01 ± 0.01 & 0.33 ± 0.21 \\
$\ \ \text{bw} = 0.1$ & 2.07 ± 0.02 & 1.00 ± 0.00 & 0.34 ± 0.00 & 0.00 ± 0.00 & 0.21 ± 0.03 \\
$\ \ \text{bw} = 0.15$ & 1.64 ± 0.08 & 1.00 ± 0.00 & 0.37 ± 0.01 & 0.05 ± 0.02 & 0.21 ± 0.01 \\
$\ \ \text{bw} = 0.2$ & 1.78 ± 0.26 & 1.00 ± 0.00 & 0.38 ± 0.03 & 0.03 ± 0.02 & 0.21 ± 0.03 \\
$\ \ \text{bw} = 0.25$ & 1.35 ± 0.07 & 0.98 ± 0.02 & 0.49 ± 0.02 & 0.13 ± 0.07 & 0.33 ± 0.03 \\
$\ \ \text{bw} = 0.3$ & 1.73 ± 0.10 & 1.00 ± 0.00 & 0.47 ± 0.03 & 0.13 ± 0.07 & 0.34 ± 0.02 \\
$\ \ \text{bw} = 0.35$ & 1.48 ± 0.02 & 0.93 ± 0.07 & 0.56 ± 0.01 & 0.07 ± 0.02 & 0.33 ± 0.00 \\
$\ \ \text{bw} = 0.4$ & 1.47 ± 0.01 & 0.84 ± 0.10 & 0.57 ± 0.01 & 0.08 ± 0.01 & 0.33 ± 0.01 \\
$\ \ \text{bw} = 0.45$ & 1.52 ± 0.01 & 0.87 ± 0.07 & 0.57 ± 0.00 & 0.08 ± 0.01 & 0.34 ± 0.01 \\
$\ \ \text{bw} = 0.5$ & 1.47 ± 0.05 & 0.79 ± 0.05 & 0.62 ± 0.00 & 0.10 ± 0.01 & 0.35 ± 0.03 \\
$\ \ \text{bw} = 0.75$ & 1.75 ± 0.04 & 0.50 ± 0.08 & 0.71 ± 0.00 & 0.10 ± 0.01 & 0.43 ± 0.03 \\
$\ \ \text{bw} = 1$ & 1.76 ± 0.07 & 0.50 ± 0.12 & 0.72 ± 0.00 & 0.09 ± 0.02 & 0.43 ± 0.02 \\
FeatureTweak & 0.52 ± 0.05 & 0.82 ± 0.02 & 0.57 ± 0.03 & 0.09 ± 0.00 & 0.18 ± 0.01 \\
FOCUS & 0.54 ± 0.13 & 0.83 ± 0.01 & 0.58 ± 0.02 & 0.48 ± 0.02 & 0.26 ± 0.01 \\
\bottomrule
\end{tabular}

\caption{CFX generation results, GermanCredit, RandomForest. Methods with nan values had 100\% failures. Validity 52\% for FeatureTweak.}
\end{table}

\subsubsection{Conformal evaluation results}

\begin{table}[H]
\begin{tabular}{lllll}
\toprule
Generator & Marginal CovGap & Binning CovGap & Class Cond CovGap & Simulated CovGap \\
\midrule
\textbf{MLP} &  &  &  &  \\
ConfexNaive &  &  &  &  \\
$\ \ \alpha = 0.01$ & 1.00 ± 0.00 & -1.00 ± 1.16 & -0.50 ± 0.71 & -56.00 ± 4.95 \\
$\ \ \alpha = 0.05$ & 0.92 ± 0.04 & -1.05 ± 0.07 & 0.75 ± 0.35 & -46.50 ± 0.71 \\
$\ \ \alpha = 0.1$ & 0.88 ± 0.04 & -7.27 ± 0.51 & -2.50 ± 0.71 & -33.75 ± 4.60 \\
ConfexTree, $\alpha = 0.01$ &  &  &  &  \\
$\ \ \text{bw} = 0.1$ & 1.00 ± 0.00 & 1.00 ± 0.00 & 1.00 ± 0.00 & nan ± nan \\
$\ \ \text{bw} = 0.2$ & 1.00 ± 0.00 & 1.00 ± 0.00 & 1.00 ± 0.00 & nan ± nan \\
$\ \ \text{bw} = 0.3$ & 1.00 ± 0.00 & 1.00 ± 0.00 & 1.00 ± 0.00 & nan ± nan \\
$\ \ \text{bw} = 0.4$ & 1.00 ± 0.00 & 1.00 ± 0.00 & 1.00 ± 0.00 & nan ± nan \\
$\ \ \text{bw} = 0.5$ & 1.00 ± 0.00 & 1.00 ± 0.00 & 1.00 ± 0.00 & nan ± nan \\
$\ \ \text{bw} = 0.6$ & 1.00 ± 0.00 & 1.00 ± 0.00 & 1.00 ± 0.00 & nan ± nan \\
$\ \ \text{bw} = 0.7$ & 1.00 ± 0.00 & 0.64 ± 0.00 & 0.50 ± 0.00 & -63.75 ± 6.72 \\
$\ \ \text{bw} = 0.8$ & 1.00 ± 0.00 & 0.64 ± 0.00 & 0.50 ± 0.00 & -63.75 ± 6.72 \\
$\ \ \text{bw} = 0.9$ & 1.00 ± 0.00 & 0.64 ± 0.00 & 0.50 ± 0.00 & -63.75 ± 6.72 \\
$\ \ \text{bw} = 1$ & 1.00 ± 0.00 & 0.64 ± 0.00 & 0.50 ± 0.00 & -63.75 ± 6.72 \\
ConfexTree, $\alpha = 0.05$ &  &  &  &  \\

$\ \ \text{bw} = 0.1$ & 1.00 ± 0.00 & 5.00 ± 0.00 & 5.00 ± 0.00 & nan ± nan \\

$\ \ \text{bw} = 0.2$ & 1.00 ± 0.00 & 5.00 ± 0.00 & 5.00 ± 0.00 & 5.00 ± nan \\

$\ \ \text{bw} = 0.3$ & 1.00 ± 0.00 & 4.64 ± 0.00 & 4.50 ± 0.00 & -31.00 ± 47.38 \\

$\ \ \text{bw} = 0.4$ & 1.00 ± 0.00 & 1.41 ± 0.58 & 2.25 ± 0.35 & -67.75 ± 10.96 \\
$\ \ \text{bw} = 0.5$ & 0.90 ± 0.00 & -1.46 ± 0.00 & 0.50 ± 0.00 & -69.50 ± 7.07 \\
$\ \ \text{bw} = 0.6$ & 0.90 ± 0.00 & -1.41 ± 0.58 & 0.25 ± 0.35 & -62.50 ± 0.71 \\
$\ \ \text{bw} = 0.7$ & 0.95 ± 0.00 & -2.10 ± 0.25 & 0.25 ± 0.35 & -41.50 ± 2.12 \\
$\ \ \text{bw} = 0.8$ & 0.95 ± 0.00 & -2.10 ± 0.25 & 0.25 ± 0.35 & -41.50 ± 2.12 \\
$\ \ \text{bw} = 0.9$ & 0.95 ± 0.00 & -2.10 ± 0.25 & 0.25 ± 0.35 & -36.00 ± 5.66 \\
$\ \ \text{bw} = 1$ & 0.95 ± 0.00 & -2.10 ± 0.25 & 0.25 ± 0.35 & -36.00 ± 5.66 \\
ConfexTree, $\alpha = 0.1$ &  &  &  &  \\
$\ \ \text{bw} = 0.05$ & 1.00 ± 0.00 & 10.00 ± 0.00 & 10.00 ± 0.00 & nan ± nan \\
$\ \ \text{bw} = 0.1$ & 1.00 ± 0.00 & 10.00 ± 0.00 & 10.00 ± 0.00 & 10.00 ± 0.00 \\
$\ \ \text{bw} = 0.2$ & 1.00 ± 0.00 & 9.64 ± 0.00 & 9.50 ± 0.00 & 7.50 ± 0.00 \\
$\ \ \text{bw} = 0.3$ & 0.95 ± 0.00 & 5.23 ± 0.58 & 6.25 ± 0.35 & -51.50 ± 7.78 \\
$\ \ \text{bw} = 0.4$ & 0.95 ± 0.00 & 0.36 ± 1.67 & 3.00 ± 1.41 & -65.00 ± 11.31 \\
$\ \ \text{bw} = 0.45$ & 0.90 ± 0.00 & -0.35 ± 1.67 & 2.00 ± 1.41 & -57.50 ± 2.83 \\
$\ \ \text{bw} = 0.5$ & 0.85 ± 0.00 & -5.63 ± 1.67 & -1.50 ± 1.41 & -42.25 ± 1.77 \\
$\ \ \text{bw} = 0.6$ & 0.90 ± 0.00 & -3.86 ± 0.83 & -0.00 ± 0.71 & -38.00 ± 0.71 \\
$\ \ \text{bw} = 0.7$ & 0.90 ± 0.07 & -5.99 ± 0.51 & -2.00 ± 0.71 & -40.75 ± 23.69 \\
$\ \ \text{bw} = 0.8$ & 0.90 ± 0.07 & -5.99 ± 0.51 & -2.00 ± 0.71 & -40.75 ± 23.69 \\
$\ \ \text{bw} = 0.9$ & 0.90 ± 0.07 & -5.99 ± 0.51 & -2.00 ± 0.71 & -40.75 ± 23.69 \\
$\ \ \text{bw} = 1$ & 0.90 ± 0.07 & -5.99 ± 0.51 & -2.00 ± 0.71 & -40.75 ± 23.69 \\
\bottomrule
\end{tabular}

\caption{Conformal evaluation results, GermanCredit, MLP}
\end{table}

\begin{table}[H]
\begin{tabular}{lllll}
\toprule
Generator & Marginal CovGap & Binning CovGap & Class Cond CovGap & Simulated CovGap \\
\midrule
\textbf{RandomForest} &  &  &  &  \\
ConfexNaive &  &  &  &  \\
$\ \ \alpha = 0.01$ & 1.00 ± 0.00 & 1.00 ± 0.00 & 1.00 ± 0.00 & nan ± nan \\
$\ \ \alpha = 0.05$ & 1.00 ± 0.00 & 5.00 ± 0.00 & 5.00 ± 0.00 & nan ± nan \\
$\ \ \alpha = 0.1$ & 0.92 ± 0.04 & -0.97 ± 1.23 & 2.75 ± 0.35 & -33.75 ± 12.37 \\
ConfexTree, $\alpha = 0.01$ &  &  &  &  \\
$\ \ \text{bw} = 0.1$ & 1.00 ± 0.00 & 1.00 ± 0.00 & 1.00 ± 0.00 & nan ± nan \\
$\ \ \text{bw} = 0.2$ & 1.00 ± 0.00 & 1.00 ± 0.00 & 1.00 ± 0.00 & nan ± nan \\
$\ \ \text{bw} = 0.3$ & 1.00 ± 0.00 & 1.00 ± 0.00 & 1.00 ± 0.00 & nan ± nan \\
$\ \ \text{bw} = 0.4$ & 1.00 ± 0.00 & 1.00 ± 0.00 & 1.00 ± 0.00 & nan ± nan \\
$\ \ \text{bw} = 0.5$ & 1.00 ± 0.00 & 1.00 ± 0.00 & 1.00 ± 0.00 & nan ± nan \\
$\ \ \text{bw} = 0.6$ & 1.00 ± 0.00 & 1.00 ± 0.00 & 1.00 ± 0.00 & nan ± nan \\
$\ \ \text{bw} = 0.7$ & 1.00 ± 0.00 & 1.00 ± 0.00 & 1.00 ± 0.00 & nan ± nan \\
$\ \ \text{bw} = 0.8$ & 1.00 ± 0.00 & 1.00 ± 0.00 & 1.00 ± 0.00 & nan ± nan \\
$\ \ \text{bw} = 0.9$ & 1.00 ± 0.00 & 1.00 ± 0.00 & 1.00 ± 0.00 & nan ± nan \\
$\ \ \text{bw} = 1$ & 1.00 ± 0.00 & 1.00 ± 0.00 & 1.00 ± 0.00 & nan ± nan \\
ConfexTree, $\alpha = 0.05$ &  &  &  &  \\
$\ \ \text{bw} = 0.1$ & 1.00 ± 0.00 & 5.00 ± 0.00 & 5.00 ± 0.00 & nan ± nan \\
$\ \ \text{bw} = 0.15$ & 1.00 ± 0.00 & 5.00 ± 0.00 & 5.00 ± 0.00 & nan ± nan \\
$\ \ \text{bw} = 0.2$ & 1.00 ± 0.00 & 5.00 ± 0.00 & 5.00 ± 0.00 & nan ± nan \\
$\ \ \text{bw} = 0.3$ & 1.00 ± 0.00 & 4.82 ± 0.25 & 4.75 ± 0.35 & 2.50 ± nan \\
$\ \ \text{bw} = 0.4$ & 1.00 ± 0.00 & 4.82 ± 0.25 & 4.75 ± 0.35 & 2.50 ± nan \\
$\ \ \text{bw} = 0.5$ & 0.98 ± 0.04 & 2.13 ± 1.74 & 3.25 ± 1.06 & -16.31 ± 11.59 \\
$\ \ \text{bw} = 0.6$ & 0.98 ± 0.04 & 1.72 ± 1.16 & 3.00 ± 0.71 & -26.15 ± 2.32 \\
$\ \ \text{bw} = 0.7$ & 0.95 ± 0.00 & 0.13 ± 4.57 & 1.75 ± 3.18 & -51.18 ± 17.93 \\
$\ \ \text{bw} = 0.8$ & 0.95 ± 0.00 & 0.13 ± 4.57 & 1.75 ± 3.18 & -51.18 ± 17.93 \\
$\ \ \text{bw} = 0.9$ & 0.95 ± 0.00 & 0.13 ± 4.57 & 1.75 ± 3.18 & -49.43 ± 20.40 \\
$\ \ \text{bw} = 1$ & 0.95 ± 0.00 & 0.13 ± 4.57 & 1.75 ± 3.18 & -49.43 ± 20.40 \\
ConfexTree, $\alpha = 0.1$ &  &  &  &  \\
$\ \ \text{bw} = 0.05$ & 1.00 ± 0.00 & 10.00 ± 0.00 & 10.00 ± 0.00 & nan ± nan \\
$\ \ \text{bw} = 0.1$ & 1.00 ± 0.00 & 10.00 ± 0.00 & 10.00 ± 0.00 & 10.00 ± nan \\
$\ \ \text{bw} = 0.15$ & 1.00 ± 0.00 & 9.82 ± 0.25 & 9.75 ± 0.35 & 7.50 ± nan \\
$\ \ \text{bw} = 0.2$ & 1.00 ± 0.00 & 9.82 ± 0.25 & 9.75 ± 0.35 & 7.50 ± nan \\
$\ \ \text{bw} = 0.25$ & 0.98 ± 0.04 & 5.82 ± 0.40 & 6.75 ± 0.35 & -30.25 ± 43.49 \\
$\ \ \text{bw} = 0.3$ & 0.98 ± 0.04 & 5.59 ± 1.23 & 6.75 ± 0.35 & -36.75 ± 49.85 \\
$\ \ \text{bw} = 0.35$ & 0.98 ± 0.04 & 3.31 ± 0.33 & 5.50 ± 0.00 & -36.75 ± 48.44 \\
$\ \ \text{bw} = 0.4$ & 0.98 ± 0.04 & 3.31 ± 0.33 & 5.50 ± 0.00 & -36.75 ± 48.44 \\
$\ \ \text{bw} = 0.45$ & 0.98 ± 0.04 & 4.13 ± 0.83 & 6.00 ± 0.71 & -22.25 ± 29.34 \\
$\ \ \text{bw} = 0.5$ & 0.95 ± 0.00 & -0.10 ± 2.32 & 3.00 ± 1.41 & -23.75 ± 22.27 \\
$\ \ \text{bw} = 0.6$ & 0.95 ± 0.00 & -0.51 ± 3.41 & 2.75 ± 2.47 & -19.50 ± 12.73 \\
$\ \ \text{bw} = 0.7$ & 0.92 ± 0.04 & -0.79 ± 1.48 & 3.00 ± 0.71 & -34.00 ± 12.02 \\
$\ \ \text{bw} = 0.8$ & 0.92 ± 0.04 & -0.79 ± 1.48 & 3.00 ± 0.71 & -34.00 ± 12.02 \\
$\ \ \text{bw} = 0.9$ & 0.92 ± 0.04 & -0.79 ± 1.48 & 3.00 ± 0.71 & -32.75 ± 13.79 \\
$\ \ \text{bw} = 1$ & 0.92 ± 0.04 & -0.79 ± 1.48 & 3.00 ± 0.71 & -32.75 ± 13.79 \\
\bottomrule
\end{tabular}
\caption{Conformal evaluation results, GermanCredit, RandomForest}
\end{table}

\newpage

\subsection{GiveMeSomeCredit}

This dataset, obtained through Kaggle\footnote{\url{https://www.kaggle.com/competitions/GiveMeSomeCredit}}, contains credit scoring data with 8 numeric features that were scaled to $(0, 1)$ using MinMax scaling.

\subsubsection{Model evaluation results}

\begin{table}[ht!]
\centering
\renewcommand{\arraystretch}{1.2}
\setlength{\tabcolsep}{8pt}
\begin{tabular}{lcccc}
\toprule
Repeat  & Accuracy (\%) & Precision (\%) & F1 Score (\%) \\
\midrule
repeat0,MLP & 93.54 & 91.82 & 91.81 \\
repeat1,MLP & 93.49 & 91.79 & 91.96 \\
repeat0,RF  & 93.40 & 91.57 & 91.78 \\
repeat1,RF  & 93.40 & 91.53 & 91.69 \\
\bottomrule
\end{tabular}
\caption{Model evaluation results, GiveMeSomeCredit.}
\end{table}

\subsubsection{Plots}

\begin{figure}[H]
\centering
\begin{subfigure}{0.315\linewidth}
    \includegraphics[width=\linewidth]{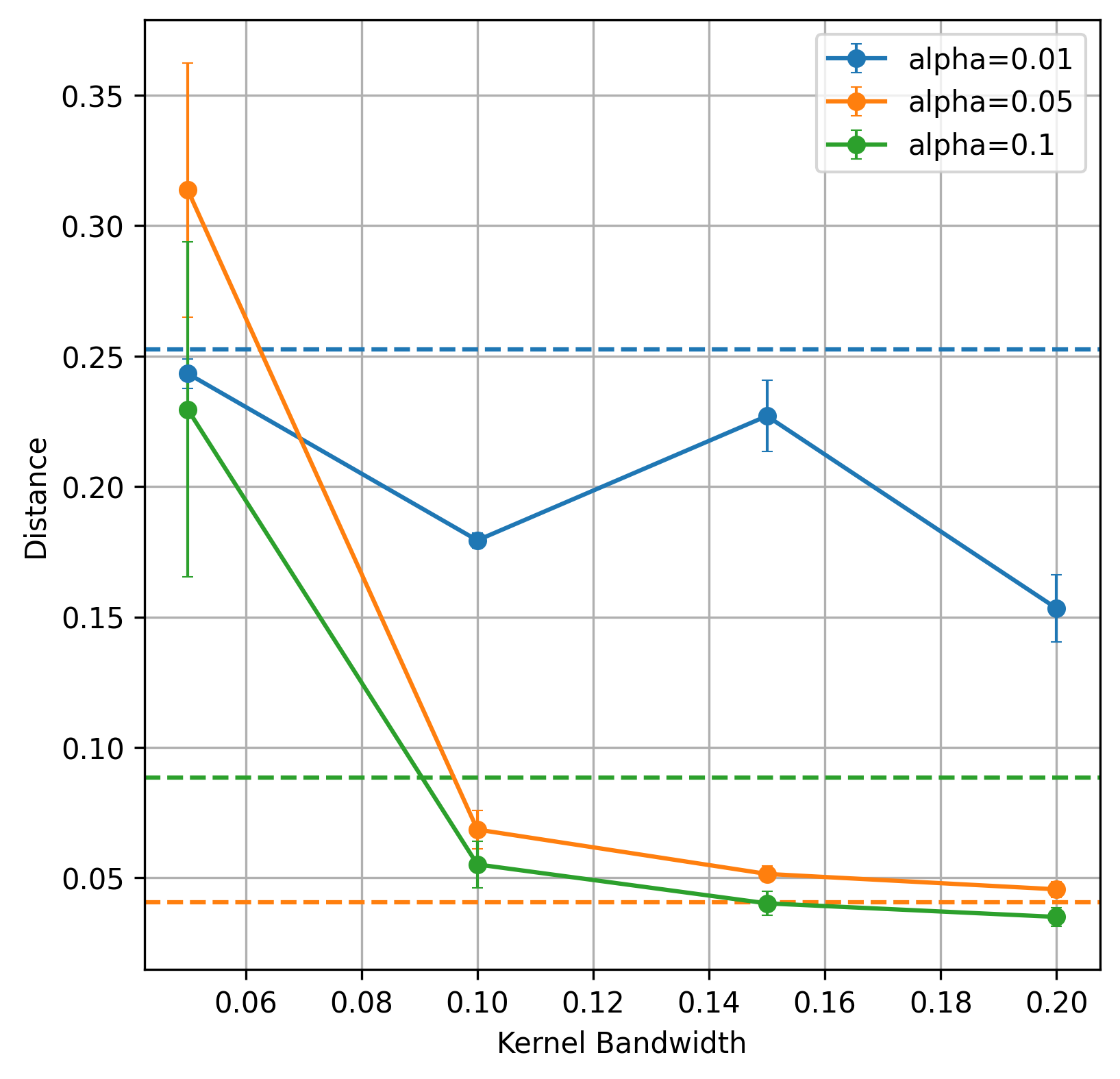}
    \caption{Distance}
\end{subfigure}
\hfill
\begin{subfigure}{0.33\linewidth}
    \includegraphics[width=\linewidth]{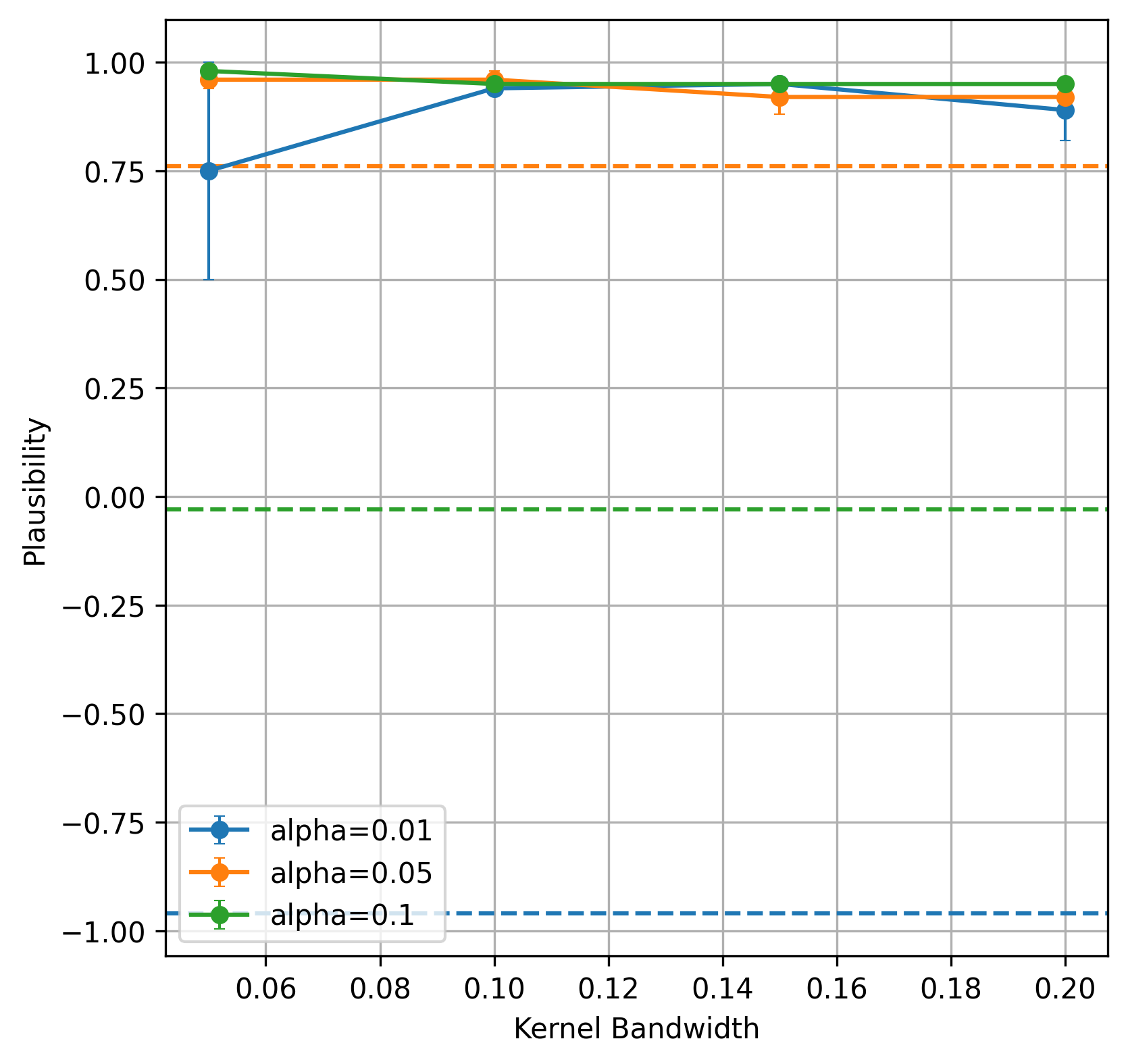}
    \caption{Plausibility}
\end{subfigure}
\hfill
\begin{subfigure}{0.32\linewidth}
    \includegraphics[width=\linewidth]{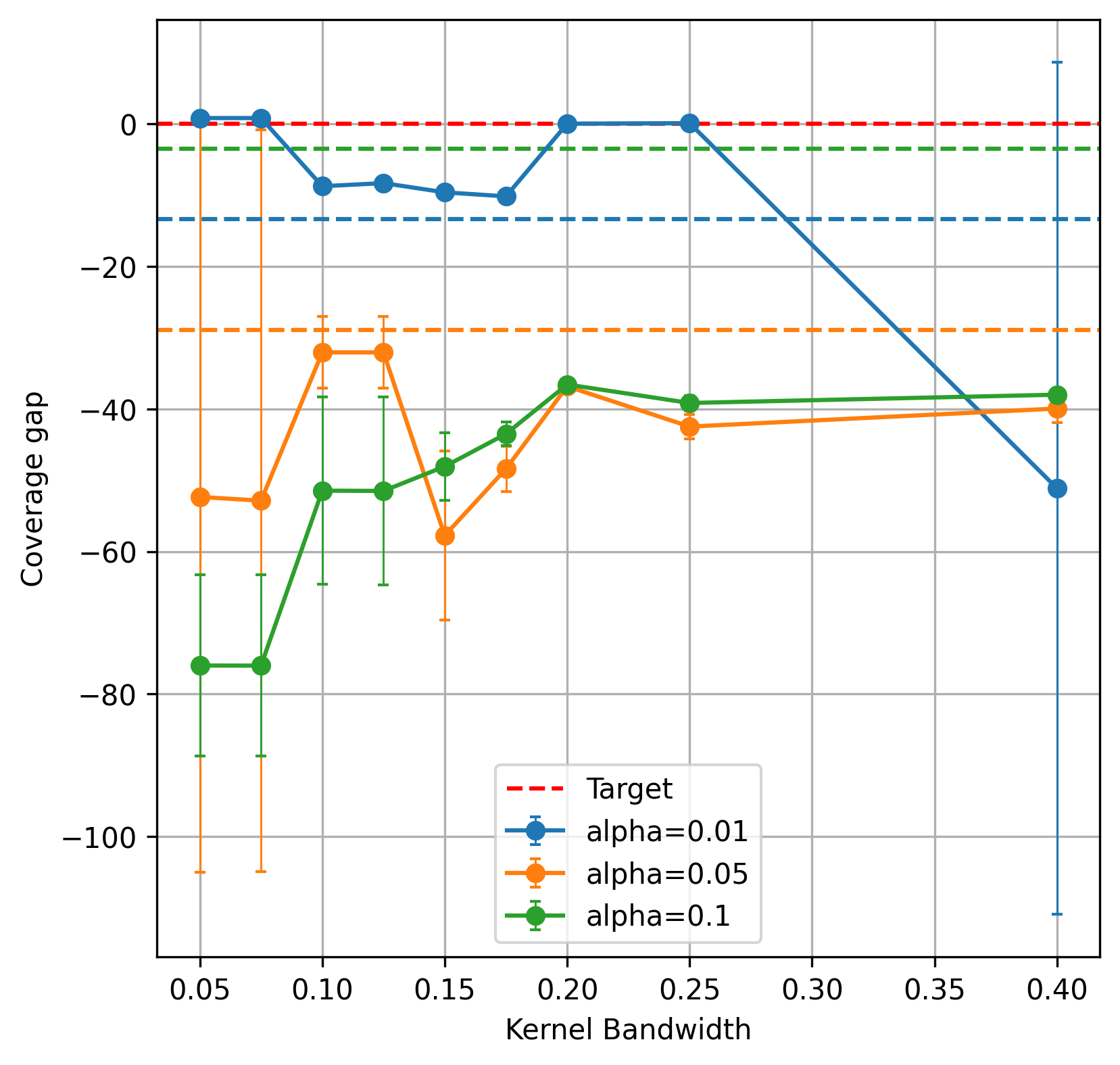}
    \caption{Coverage Gap}
\end{subfigure}
\caption{Effect of coverage rate and kernel bandwidth on metrics for CONFEX-Tree on the GiveMeSomeCredit dataset, MLP. \confexnaive\ is represented by dashed horizontal lines.}
\label{Figure 2}
\end{figure}

\begin{figure}[H]
\centering
\begin{subfigure}{0.315\linewidth}
    \includegraphics[width=\linewidth]{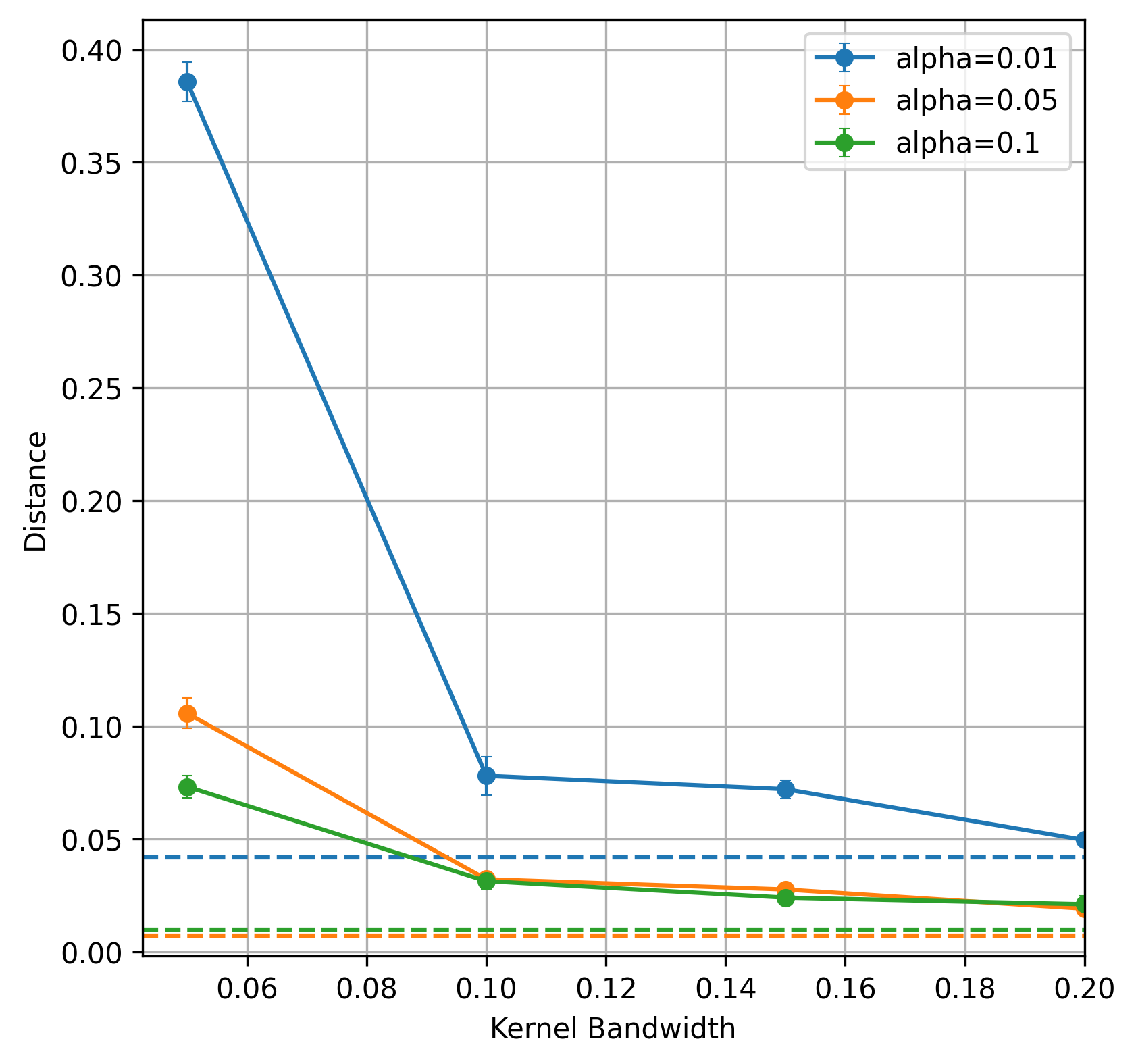}
    \caption{Distance}
\end{subfigure}
\hfill
\begin{subfigure}{0.33\linewidth}
    \includegraphics[width=\linewidth]{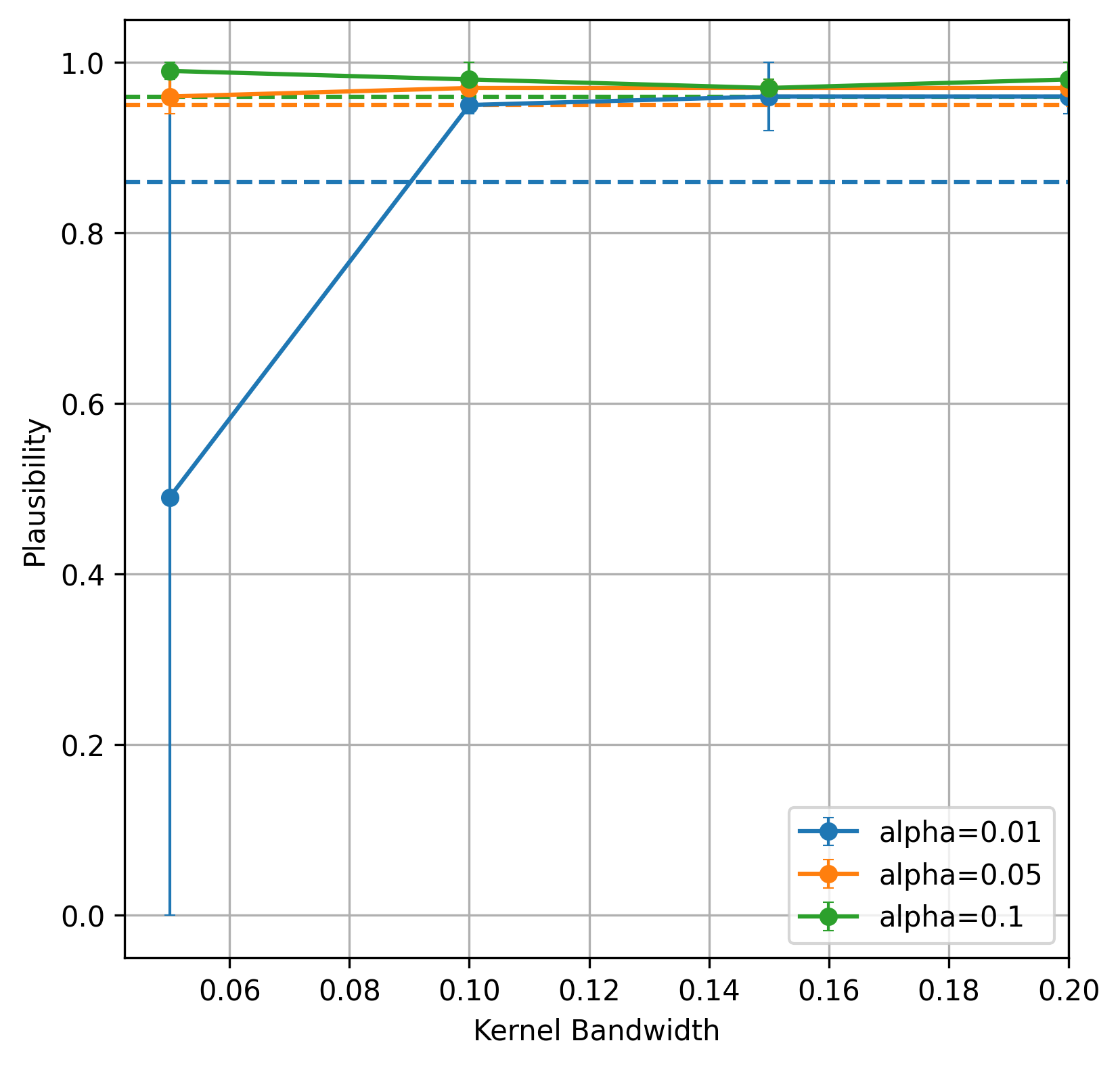}
    \caption{Plausibility}
\end{subfigure}
\hfill
\begin{subfigure}{0.32\linewidth}
    \includegraphics[width=\linewidth]{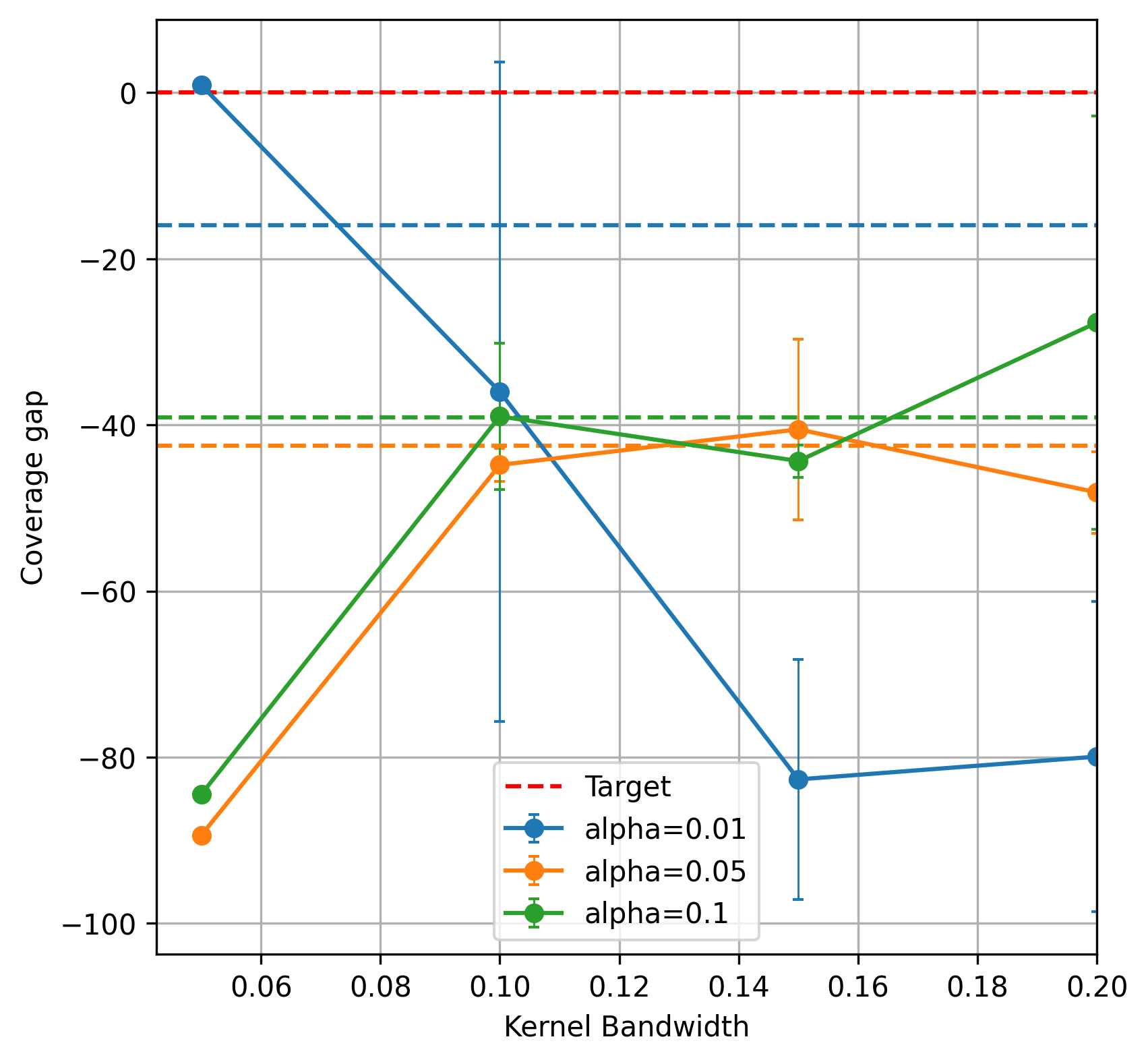}
    \caption{Coverage Gap}
\end{subfigure}
\caption{Effect of coverage rate and kernel bandwidth on metrics for CONFEX-Tree on the GiveMeSomeCredit dataset, RandomForest. \confexnaive\ is represented by dashed horizontal lines.}
\label{Figure 2}
\end{figure}

\subsubsection{CFX generation results}

\begin{table}[H]

\begin{tabular}{llllll}
\toprule
Generator & Distance & Plausibility & Implausibility & Sensitivity $(10^{-1})$ & Stability \\
\midrule
\textbf{MLP} &  &  &  &  &  \\
MinDist & 0.03 ± 0.00 & 0.94 ± 0.00 & 0.09 ± 0.00 & 1.16 ± 0.07 & 0.17 ± 0.03 \\
Wachter & 0.08 ± 0.00 & 0.92 ± 0.00 & 0.09 ± 0.00 & 0.98 ± 0.03 & 0.18 ± 0.03 \\
Greedy & 0.16 ± 0.09 & -0.26 ± 0.44 & 0.16 ± 0.04 & 0.62 ± 0.07 & 0.16 ± 0.06 \\
ConfexNaive &  &  &  &  &  \\
$\ \ \alpha = 0.01$ & 0.25 ± 0.00 & -0.96 ± 0.02 & 0.21 ± 0.01 & 0.20 ± 0.03 & 0.12 ± 0.07 \\
$\ \ \alpha = 0.025$ & 0.18 ± 0.01 & -0.78 ± 0.18 & 0.16 ± 0.02 & 0.26 ± 0.00 & 0.12 ± 0.06 \\
$\ \ \alpha = 0.05$ & 0.04 ± 0.00 & 0.76 ± 0.04 & 0.09 ± 0.00 & 0.76 ± 0.02 & 0.17 ± 0.03 \\
$\ \ \alpha = 0.075$ & 0.04 ± 0.00 & 0.86 ± 0.04 & 0.09 ± 0.00 & 0.85 ± 0.02 & 0.17 ± 0.03 \\
$\ \ \alpha = 0.1$ & 0.09 ± 0.01 & -0.03 ± 0.43 & 0.11 ± 0.00 & 0.46 ± 0.03 & 0.16 ± 0.03 \\
ECCCo &  &  &  &  &  \\
$\ \ \alpha = 0.01$ & 0.62 ± 0.29 & -0.99 ± 0.01 & 0.27 ± 0.14 & 0.23 ± 0.01 & 0.24 ± 0.03 \\
$\ \ \alpha = 0.05$ & 0.58 ± 0.25 & -0.98 ± 0.00 & 0.25 ± 0.12 & 0.24 ± 0.01 & 0.24 ± 0.03 \\
$\ \ \alpha = 0.1$ & 0.58 ± 0.26 & -0.98 ± 0.00 & 0.26 ± 0.13 & 0.24 ± 0.00 & 0.23 ± 0.02 \\
ConfexTree, $\alpha = 0.01$ &  &  &  &  &  \\
$\ \ \text{bw} = 0.05$ & 0.24 ± 0.01 & 0.75 ± 0.25 & 0.07 ± 0.00 & 0.02 ± 0.01 & 0.10 ± 0.02 \\
$\ \ \text{bw} = 0.075$ & 0.24 ± 0.00 & 0.75 ± 0.25 & 0.07 ± 0.00 & 0.02 ± 0.01 & 0.10 ± 0.02 \\
$\ \ \text{bw} = 0.1$ & 0.18 ± 0.00 & 0.94 ± 0.00 & 0.07 ± 0.00 & 0.10 ± 0.02 & 0.19 ± 0.02 \\
$\ \ \text{bw} = 0.125$ & 0.18 ± 0.00 & 0.95 ± 0.01 & 0.07 ± 0.00 & 0.12 ± 0.01 & 0.19 ± 0.02 \\
$\ \ \text{bw} = 0.15$ & 0.23 ± 0.01 & 0.95 ± 0.01 & 0.08 ± 0.01 & 0.12 ± 0.02 & 0.18 ± 0.03 \\
$\ \ \text{bw} = 0.175$ & 0.17 ± 0.01 & 0.96 ± 0.00 & 0.08 ± 0.00 & 0.16 ± 0.01 & 0.18 ± 0.04 \\
$\ \ \text{bw} = 0.2$ & 0.15 ± 0.01 & 0.89 ± 0.07 & 0.08 ± 0.00 & 0.20 ± 0.00 & 0.17 ± 0.04 \\
ConfexTree, $\alpha = 0.05$ &  &  &  &  &  \\
$\ \ \text{bw} = 0.05$ & 0.31 ± 0.05 & 0.96 ± 0.02 & 0.06 ± 0.00 & 0.06 ± 0.01 & 0.21 ± 0.01 \\
$\ \ \text{bw} = 0.075$ & 0.20 ± 0.04 & 0.96 ± 0.02 & 0.07 ± 0.01 & 0.16 ± 0.03 & 0.20 ± 0.01 \\
$\ \ \text{bw} = 0.1$ & 0.07 ± 0.01 & 0.96 ± 0.02 & 0.08 ± 0.00 & 0.54 ± 0.06 & 0.18 ± 0.03 \\
$\ \ \text{bw} = 0.125$ & 0.06 ± 0.01 & 0.96 ± 0.02 & 0.08 ± 0.00 & 0.61 ± 0.06 & 0.17 ± 0.03 \\
$\ \ \text{bw} = 0.15$ & 0.05 ± 0.00 & 0.92 ± 0.04 & 0.08 ± 0.00 & 0.61 ± 0.04 & 0.17 ± 0.03 \\
$\ \ \text{bw} = 0.175$ & 0.05 ± 0.00 & 0.92 ± 0.04 & 0.09 ± 0.00 & 0.65 ± 0.03 & 0.17 ± 0.03 \\
$\ \ \text{bw} = 0.2$ & 0.05 ± 0.00 & 0.92 ± 0.04 & 0.09 ± 0.00 & 0.62 ± 0.00 & 0.17 ± 0.03 \\
ConfexTree, $\alpha = 0.1$ &  &  &  &  &  \\
$\ \ \text{bw} = 0.05$ & 0.23 ± 0.06 & 0.98 ± 0.00 & 0.08 ± 0.01 & 0.13 ± 0.03 & 0.20 ± 0.01 \\
$\ \ \text{bw} = 0.075$ & 0.17 ± 0.05 & 0.98 ± 0.02 & 0.07 ± 0.01 & 0.23 ± 0.04 & 0.19 ± 0.02 \\
$\ \ \text{bw} = 0.1$ & 0.06 ± 0.01 & 0.95 ± 0.01 & 0.08 ± 0.00 & 0.66 ± 0.05 & 0.17 ± 0.03 \\
$\ \ \text{bw} = 0.125$ & 0.05 ± 0.01 & 0.95 ± 0.01 & 0.08 ± 0.00 & 0.69 ± 0.06 & 0.17 ± 0.03 \\
$\ \ \text{bw} = 0.15$ & 0.04 ± 0.00 & 0.95 ± 0.01 & 0.09 ± 0.00 & 0.78 ± 0.01 & 0.17 ± 0.03 \\
$\ \ \text{bw} = 0.175$ & 0.04 ± 0.00 & 0.95 ± 0.01 & 0.09 ± 0.00 & 0.76 ± 0.00 & 0.17 ± 0.03 \\
$\ \ \text{bw} = 0.2$ & 0.03 ± 0.00 & 0.95 ± 0.01 & 0.09 ± 0.00 & 0.76 ± 0.01 & 0.17 ± 0.03 \\
\bottomrule
\end{tabular}

\caption{CFX generation results, GiveMeSomeCredit, MLP. Wachter had 29\% validity, Schut had 80\% validity.}
\end{table}

\begin{table}[H]
\begin{tabular}{llllll}
\toprule
Generator & Distance & Plausibility & Implausibility & Sensitivity $(10^{-1})$ & Stability \\
\midrule
\textbf{RandomForest} &  &  &  &  &  \\
MinDist & 0.01 ± 0.00 & 0.95 ± 0.01 & 0.09 ± 0.00 & 100.94 ± 11.42 & 0.29 ± 0.01 \\
ConfexNaive &  &  &  &  &  \\
$\ \ \alpha = 0.01$ & 0.04 ± 0.00 & 0.86 ± 0.00 & 0.09 ± 0.00 & 1.23 ± 0.05 & 0.30 ± 0.01 \\
$\ \ \alpha = 0.05$ & 0.01 ± 0.00 & 0.95 ± 0.01 & 0.09 ± 0.00 & 42.08 ± 36.87 & 0.29 ± 0.01 \\
$\ \ \alpha = 0.1$ & 0.01 ± 0.00 & 0.96 ± 0.00 & 0.09 ± 0.00 & 40.40 ± 35.97 & 0.29 ± 0.01 \\
ConfexTree, $\alpha = 0.01$ &  &  &  &  &  \\
$\ \ \text{bw} = 0.05$ & 0.39 ± 0.01 & 0.49 ± 0.49 & 0.07 ± 0.01 & 0.01 ± 0.00 & 0.22 ± 0.04 \\
$\ \ \text{bw} = 0.1$ & 0.08 ± 0.01 & 0.95 ± 0.01 & 0.07 ± 0.00 & 0.38 ± 0.06 & 0.29 ± 0.01 \\
$\ \ \text{bw} = 0.15$ & 0.07 ± 0.00 & 0.96 ± 0.04 & 0.07 ± 0.00 & 0.47 ± 0.03 & 0.29 ± 0.01 \\
$\ \ \text{bw} = 0.2$ & 0.05 ± 0.00 & 0.96 ± 0.02 & 0.08 ± 0.00 & 0.71 ± 0.14 & 0.29 ± 0.01 \\
ConfexTree, $\alpha = 0.05$ &  &  &  &  &  \\
$\ \ \text{bw} = 0.05$ & 0.11 ± 0.01 & 0.96 ± 0.02 & 0.07 ± 0.00 & 0.35 ± 0.07 & 0.29 ± 0.02 \\
$\ \ \text{bw} = 0.1$ & 0.03 ± 0.00 & 0.97 ± 0.01 & 0.08 ± 0.00 & 1.29 ± 0.17 & 0.29 ± 0.01 \\
$\ \ \text{bw} = 0.15$ & 0.03 ± 0.00 & 0.97 ± 0.01 & 0.08 ± 0.00 & 1.77 ± 0.49 & 0.29 ± 0.01 \\
$\ \ \text{bw} = 0.2$ & 0.02 ± 0.00 & 0.97 ± 0.01 & 0.08 ± 0.00 & 1.91 ± 0.27 & 0.29 ± 0.01 \\
ConfexTree, $\alpha = 0.1$ &  &  &  &  &  \\
$\ \ \text{bw} = 0.05$ & 0.07 ± 0.00 & 0.99 ± 0.01 & 0.08 ± 0.00 & 0.56 ± 0.12 & 0.29 ± 0.01 \\
$\ \ \text{bw} = 0.1$ & 0.03 ± 0.00 & 0.98 ± 0.02 & 0.08 ± 0.00 & 1.37 ± 0.18 & 0.29 ± 0.01 \\
$\ \ \text{bw} = 0.15$ & 0.02 ± 0.00 & 0.97 ± 0.01 & 0.08 ± 0.00 & 4.83 ± 2.84 & 0.29 ± 0.01 \\
$\ \ \text{bw} = 0.2$ & 0.02 ± 0.00 & 0.98 ± 0.02 & 0.09 ± 0.00 & 5.01 ± 2.90 & 0.29 ± 0.01 \\
FeatureTweak & 0.26 ± 0.01 & -1.00 ± 0.00 & 0.18 ± 0.00 & 0.18 ± 0.01 & 0.31 ± 0.02 \\
FeatureTweak & 0.03 ± 0.00 & 0.96 ± 0.00 & 0.09 ± 0.00 & 1.37 ± 0.07 & 0.26 ± 0.01 \\
FOCUS & 0.05 ± 0.00 & 0.92 ± 0.02 & 0.09 ± 0.00 & 2.14 ± 0.84 & 0.30 ± 0.01 \\
\bottomrule
\end{tabular}

\caption{CFX generation results, GiveMeSomeCredit, MLP.  ConfexTree with alpha=0.01,bw=0.05 had 49\% failures (i.e. one class had no singleton regions). Valdiity 46\% for FeatureTweak.}
\end{table}

\subsubsection{Conformal evaluation results}
\begin{tabular}{lllll}
\toprule
Generator & Marginal CovGap & Binning CovGap & Class Cond CovGap & Simulated CovGap \\
\midrule
\textbf{MLP} &  &  &  &  \\
ConfexNaive &  &  &  &  \\
$\ \ \alpha = 0.01$ & 0.99 ± 0.00 & -5.27 ± 0.02 & 0.14 ± 0.00 & -13.38 ± 0.90 \\
$\ \ \alpha = 0.05$ & 0.96 ± 0.00 & -30.29 ± 1.42 & -0.05 ± 0.07 & -28.92 ± 5.15 \\
$\ \ \alpha = 0.1$ & 0.90 ± 0.00 & -41.37 ± 0.35 & -0.04 ± 0.05 & -3.52 ± 10.55 \\
ConfexTree, $\alpha = 0.01$ &  &  &  &  \\
$\ \ \text{bw} = 0.05$ & 1.00 ± 0.00 & 0.98 ± 0.00 & 1.00 ± 0.00 & 0.80 ± 0.00 \\
$\ \ \text{bw} = 0.075$ & 1.00 ± 0.00 & 0.98 ± 0.00 & 1.00 ± 0.00 & 0.80 ± 0.00 \\
$\ \ \text{bw} = 0.1$ & 1.00 ± 0.00 & -1.85 ± 0.02 & 0.61 ± 0.00 & -8.76 ± 0.17 \\
$\ \ \text{bw} = 0.125$ & 1.00 ± 0.00 & -1.73 ± 0.02 & 0.63 ± 0.00 & -8.32 ± 0.17 \\
$\ \ \text{bw} = 0.15$ & 1.00 ± 0.00 & -2.17 ± 0.05 & 0.57 ± 0.01 & -9.62 ± 0.00 \\
$\ \ \text{bw} = 0.175$ & 1.00 ± 0.00 & -2.29 ± 0.05 & 0.55 ± 0.01 & -10.18 ± 0.03 \\
$\ \ \text{bw} = 0.2$ & 1.00 ± 0.00 & -4.10 ± 0.26 & 0.30 ± 0.04 & 0.01 ± 0.03 \\
ConfexTree, $\alpha = 0.05$ &  &  &  &  \\
$\ \ \text{bw} = 0.05$ & 1.00 ± 0.00 & -1.04 ± 0.04 & 4.18 ± 0.01 & -52.35 ± 52.66 \\
$\ \ \text{bw} = 0.075$ & 1.00 ± 0.00 & -1.29 ± 0.04 & 4.14 ± 0.01 & -52.87 ± 52.02 \\
$\ \ \text{bw} = 0.1$ & 0.98 ± 0.00 & -12.34 ± 0.05 & 2.63 ± 0.00 & -32.06 ± 5.04 \\
$\ \ \text{bw} = 0.125$ & 0.98 ± 0.00 & -12.36 ± 0.05 & 2.62 ± 0.00 & -32.07 ± 5.03 \\
$\ \ \text{bw} = 0.15$ & 0.98 ± 0.00 & -14.45 ± 0.13 & 2.31 ± 0.00 & -57.76 ± 11.84 \\
$\ \ \text{bw} = 0.175$ & 0.98 ± 0.00 & -15.12 ± 0.18 & 2.21 ± 0.01 & -48.40 ± 3.21 \\
$\ \ \text{bw} = 0.2$ & 0.97 ± 0.00 & -19.02 ± 0.56 & 1.63 ± 0.06 & -36.83 ± 0.40 \\
ConfexTree, $\alpha = 0.1$ &  &  &  &  \\
$\ \ \text{bw} = 0.05$ & 0.99 ± 0.00 & -0.30 ± 0.02 & 8.58 ± 0.00 & -76.00 ± 12.72 \\
$\ \ \text{bw} = 0.075$ & 0.99 ± 0.00 & -0.49 ± 0.02 & 8.55 ± 0.00 & -76.01 ± 12.72 \\
$\ \ \text{bw} = 0.1$ & 0.97 ± 0.00 & -13.62 ± 0.18 & 6.73 ± 0.00 & -51.47 ± 13.17 \\
$\ \ \text{bw} = 0.125$ & 0.97 ± 0.00 & -13.82 ± 0.18 & 6.70 ± 0.00 & -51.50 ± 13.17 \\
$\ \ \text{bw} = 0.15$ & 0.97 ± 0.00 & -17.27 ± 0.31 & 6.15 ± 0.01 & -48.07 ± 4.72 \\
$\ \ \text{bw} = 0.175$ & 0.96 ± 0.00 & -17.89 ± 0.33 & 6.06 ± 0.01 & -43.51 ± 1.64 \\
$\ \ \text{bw} = 0.2$ & 0.96 ± 0.00 & -21.56 ± 0.53 & 5.47 ± 0.03 & -36.58 ± 0.58 \\
\bottomrule
\end{tabular}

\begin{table}[H]

\caption{Conformal evaluation results, GiveMeSomeCredit, MLP}
\end{table}

\begin{table}[H]
\begin{tabular}{lllll}
\toprule
Generator & Marginal CovGap & Binning CovGap & Class Cond CovGap & Simulated CovGap \\
\midrule
\textbf{RandomForest} &  &  &  &  \\
ConfexNaive &  &  &  &  \\
$\ \ \alpha = 0.01$ & 0.99 ± 0.00 & -6.39 ± 0.00 & -0.01 ± 0.00 & -15.95 ± 0.10 \\
$\ \ \alpha = 0.05$ & 0.95 ± 0.00 & -30.52 ± 0.16 & -0.16 ± 0.03 & -42.48 ± 3.63 \\
$\ \ \alpha = 0.1$ & 0.90 ± 0.00 & -39.82 ± 0.29 & -0.08 ± 0.15 & -39.07 ± 1.77 \\
ConfexTree, $\alpha = 0.01$ &  &  &  &  \\
$\ \ \text{bw} = 0.05$ & 1.00 ± 0.00 & 0.99 ± 0.02 & 1.00 ± 0.00 & 0.90 ± 0.14 \\
$\ \ \text{bw} = 0.1$ & 1.00 ± 0.00 & -1.85 ± 0.10 & 0.58 ± 0.01 & -36.01 ± 39.68 \\
$\ \ \text{bw} = 0.15$ & 1.00 ± 0.00 & -2.01 ± 0.17 & 0.55 ± 0.01 & -82.67 ± 14.45 \\
$\ \ \text{bw} = 0.2$ & 0.99 ± 0.00 & -3.68 ± 0.09 & 0.32 ± 0.02 & -79.92 ± 18.68 \\
ConfexTree, $\alpha = 0.05$ &  &  &  &  \\
$\ \ \text{bw} = 0.05$ & 0.99 ± 0.00 & -1.43 ± 0.02 & 3.38 ± 0.07 & -89.38 ± 0.00 \\
$\ \ \text{bw} = 0.1$ & 0.97 ± 0.00 & -12.86 ± 0.31 & 1.32 ± 0.01 & -44.81 ± 2.00 \\
$\ \ \text{bw} = 0.15$ & 0.96 ± 0.00 & -14.93 ± 0.01 & 1.03 ± 0.02 & -40.53 ± 10.86 \\
$\ \ \text{bw} = 0.2$ & 0.96 ± 0.00 & -20.17 ± 0.01 & 0.38 ± 0.02 & -48.12 ± 4.92 \\
ConfexTree, $\alpha = 0.1$ &  &  &  &  \\
$\ \ \text{bw} = 0.05$ & 0.96 ± 0.00 & -1.35 ± 0.08 & 6.19 ± 0.12 & -84.46 ± 0.04 \\
$\ \ \text{bw} = 0.1$ & 0.93 ± 0.00 & -15.88 ± 0.18 & 2.50 ± 0.04 & -38.97 ± 8.82 \\
$\ \ \text{bw} = 0.15$ & 0.92 ± 0.00 & -19.85 ± 0.08 & 1.83 ± 0.09 & -44.34 ± 1.95 \\
$\ \ \text{bw} = 0.2$ & 0.91 ± 0.00 & -25.26 ± 0.22 & 1.00 ± 0.01 & -27.67 ± 24.88 \\
\bottomrule
\end{tabular}
\caption{Conformal evaluation results, GiveMeSomeCredit, RandomForest}
\end{table}

\newpage
\subsection{AdultIncome}

This dataset~\cite{kohavi1996adult}, obtained through Kaggle\footnote{\url{https://www.kaggle.com/datasets/wenruliu/adult-income-dataset}}, predicts whether an individual's income exceeds \$50,000. We processed the following features: numeric features (Age, Capital Gain, Capital Loss, Hours per week) scaled to $(0, 1)$, ordinal features (education), and categorical features (Workclass, Occupation, Race, Relationship, Gender, Marital status) using one-hot encoding. In CONFEXTree, to avoid splitting over many categorical feature combinations, we consider the first (Workclass) as a feature to split by and do not split the rest.

\subsubsection{Model evaluation results}

\begin{table}[ht!]
\centering
\renewcommand{\arraystretch}{1.2}
\setlength{\tabcolsep}{8pt}
\begin{tabular}{lcccc}
\toprule
Repeat  & Accuracy (\%) & Precision (\%) & F1 Score (\%) \\
\midrule
repeat0,RF  & 85.73 & 85.20 & 85.14 \\
repeat1,RF  & 85.32 & 84.76 & 84.72 \\
repeat0,MLP & 85.41 & 85.05 & 85.17 \\
repeat1,MLP & 85.04 & 84.70 & 84.83 \\
\bottomrule
\end{tabular}
\caption{Model evaluation results, AdultIncome.}
\end{table}

\subsubsection{Plots}

\begin{figure}[H]
\centering
\begin{subfigure}{0.315\linewidth}
    \includegraphics[width=\linewidth]{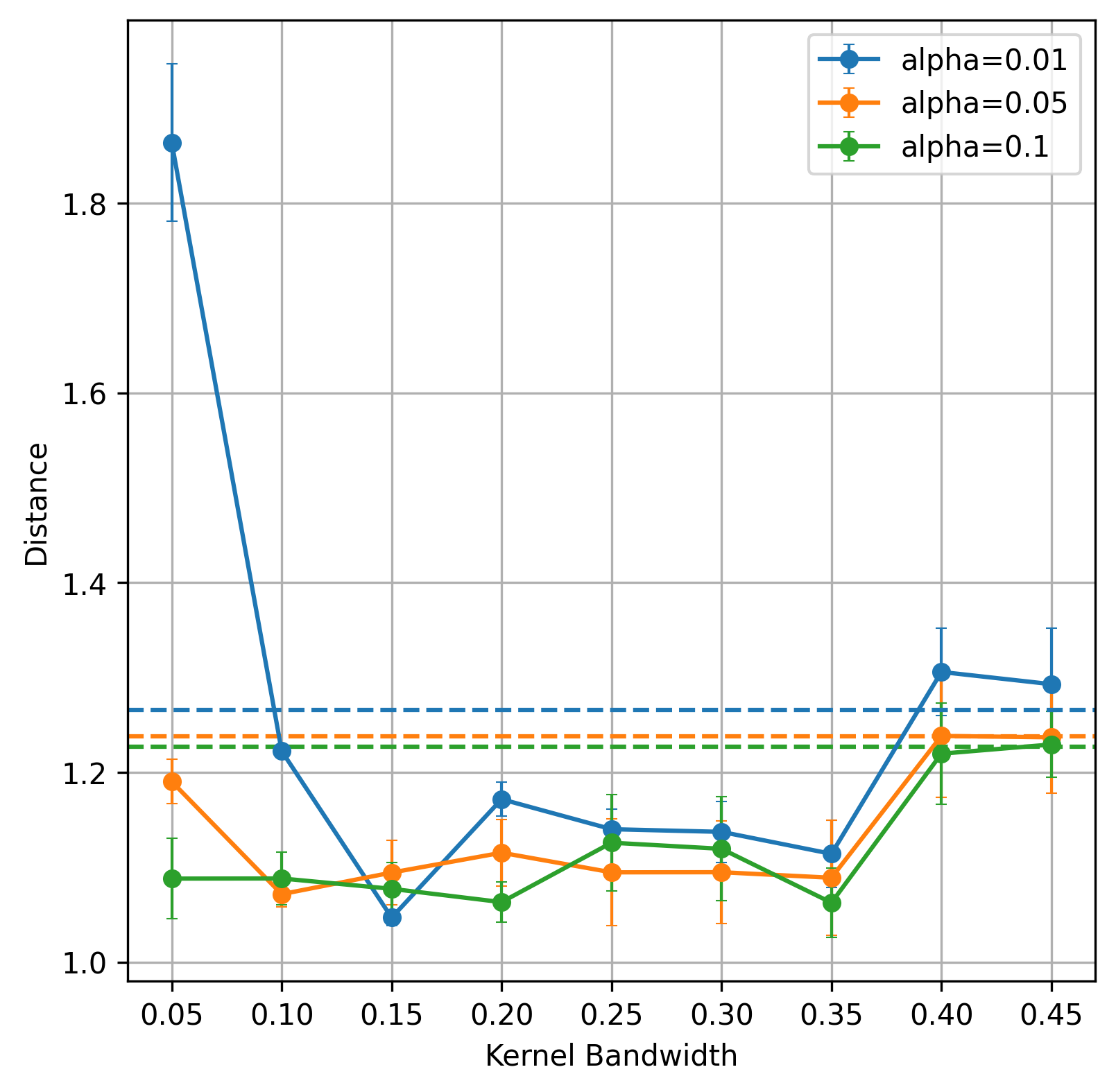}
    \caption{Distance}
\end{subfigure}
\hfill
\begin{subfigure}{0.33\linewidth}
    \includegraphics[width=\linewidth]{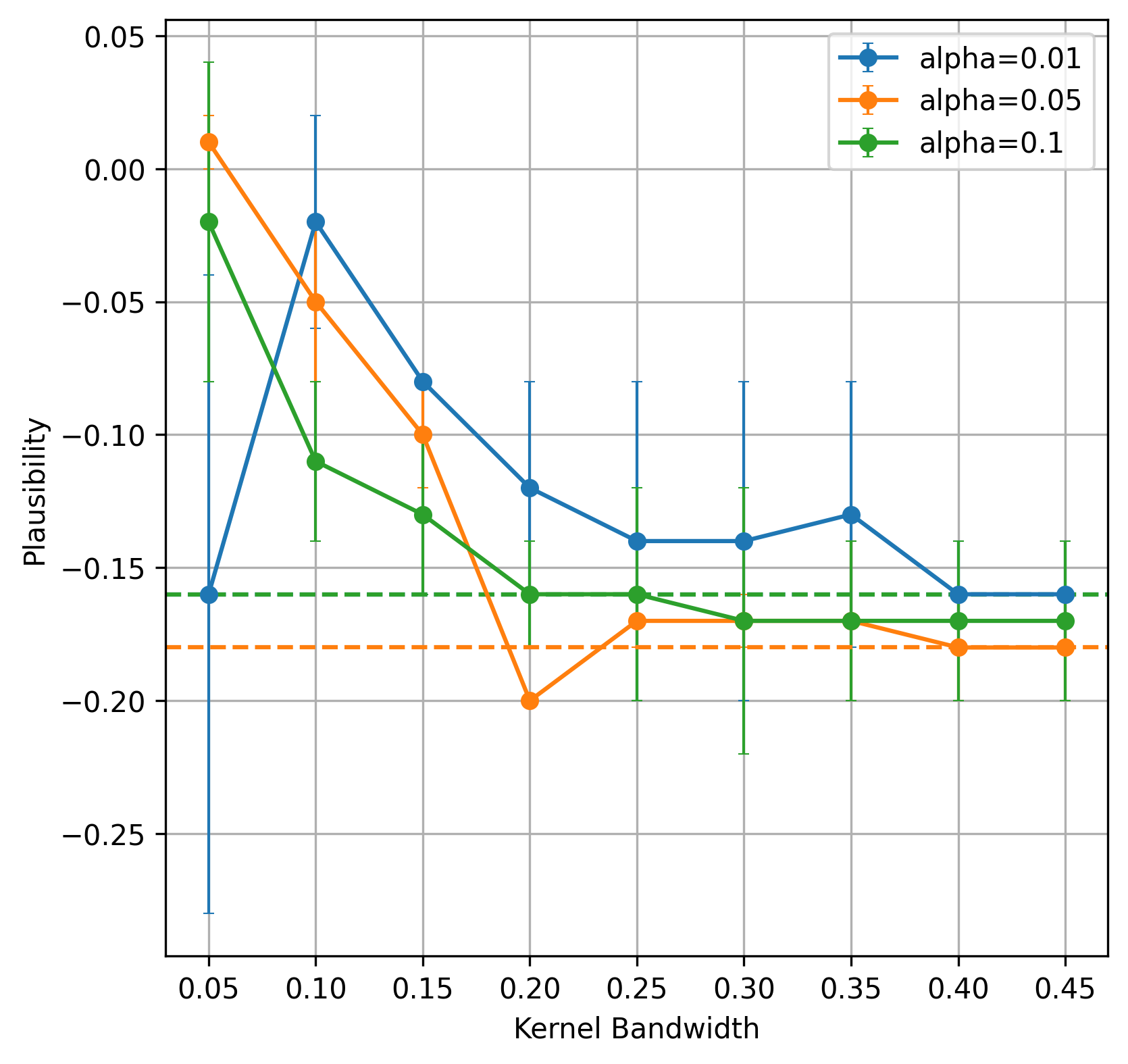}
    \caption{Plausibility}
\end{subfigure}
\hfill
\begin{subfigure}{0.32\linewidth}
    \includegraphics[width=\linewidth]{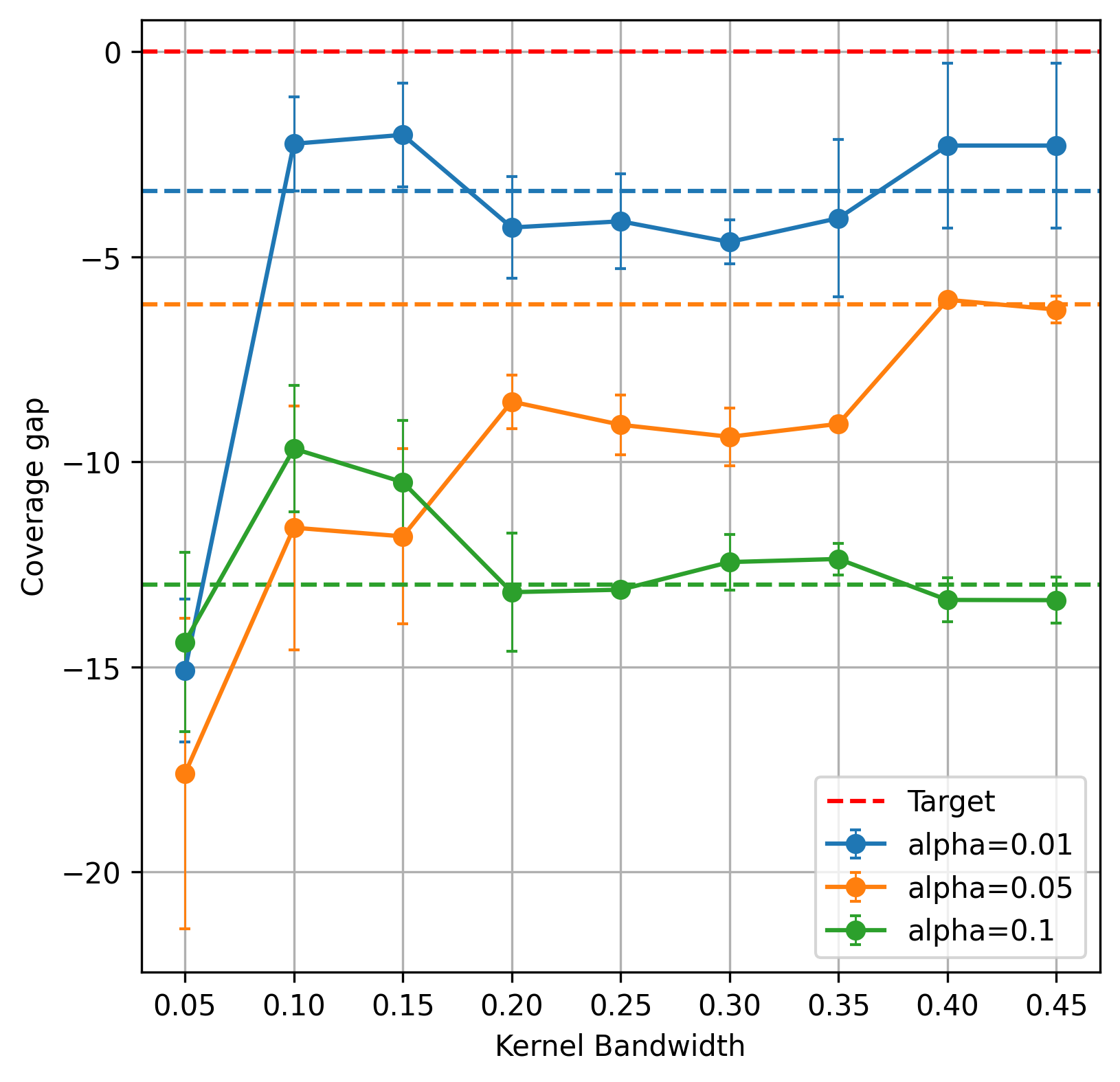}
    \caption{Coverage Gap}
\end{subfigure}
\caption{Effect of coverage rate and kernel bandwidth on metrics for CONFEX-Tree on the AdultIncome dataset, MLP. \confexnaive\ is represented by dashed horizontal lines.}
\label{Figure 2}
\end{figure}

\begin{figure}[H]
\centering
\begin{subfigure}{0.315\linewidth}
    \includegraphics[width=\linewidth]{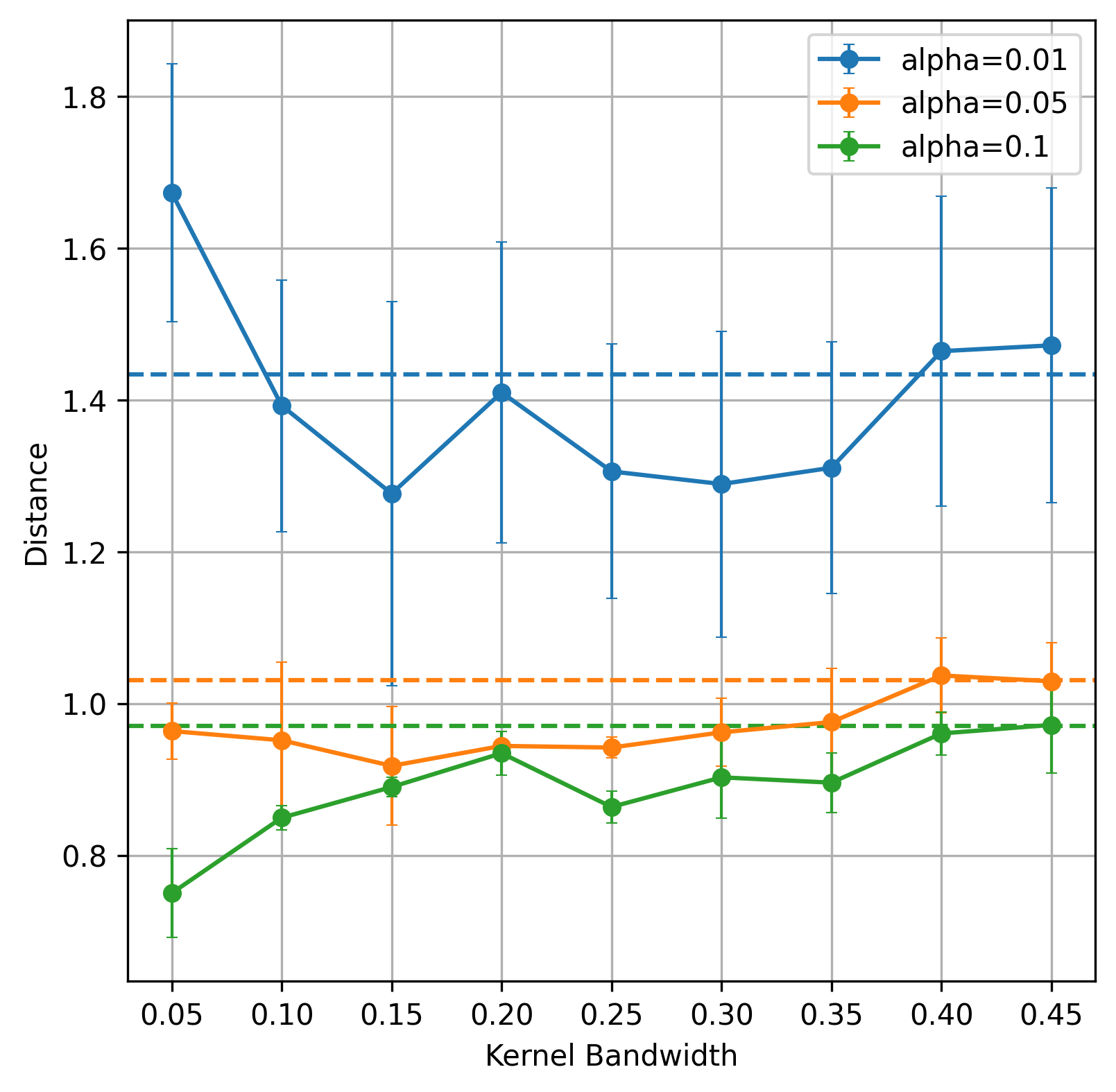}
    \caption{Distance}
\end{subfigure}
\hfill
\begin{subfigure}{0.33\linewidth}
    \includegraphics[width=\linewidth]{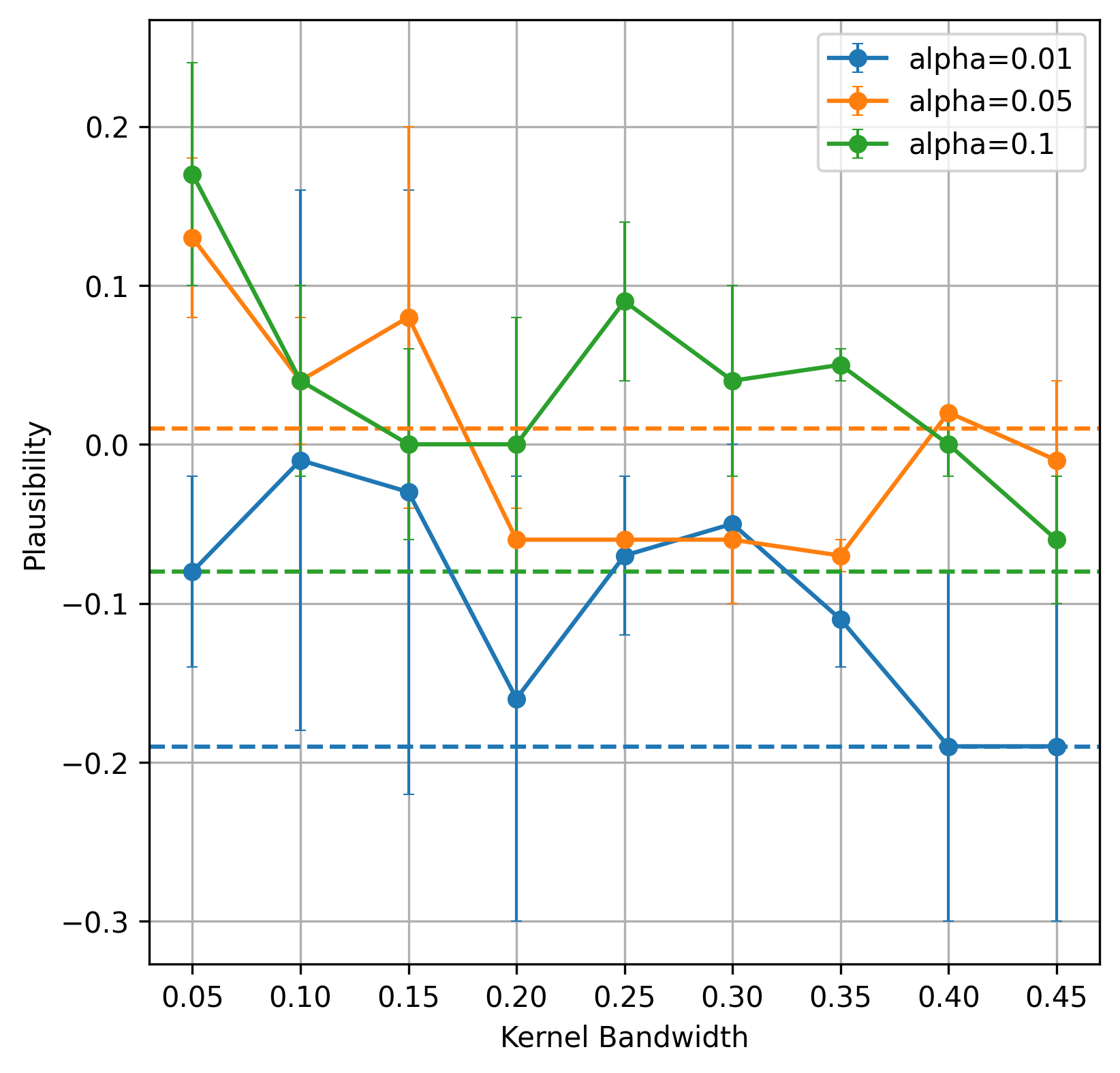}
    \caption{Plausibility}
\end{subfigure}
\hfill
\begin{subfigure}{0.32\linewidth}
    \includegraphics[width=\linewidth]{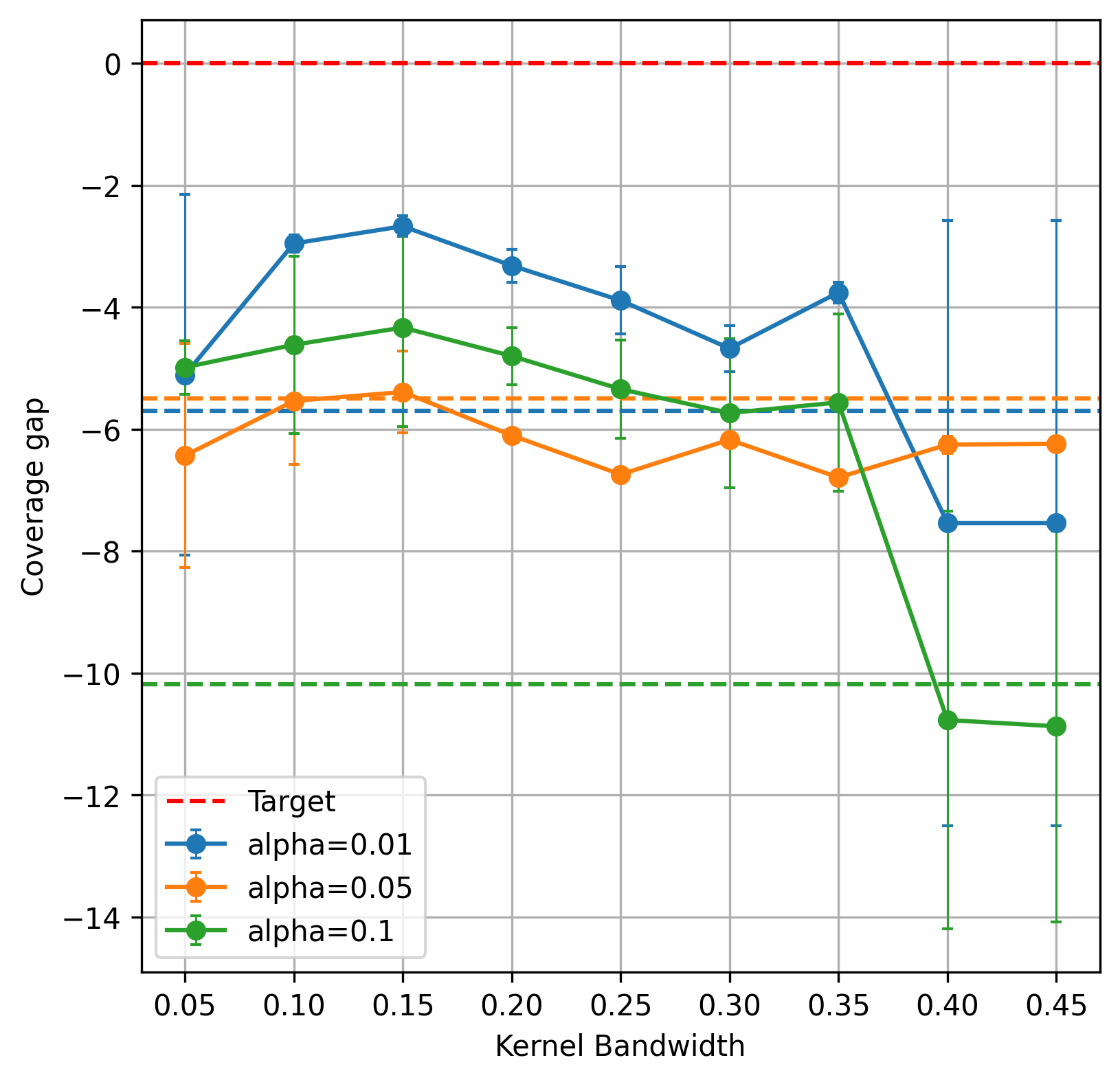}
    \caption{Coverage Gap}
\end{subfigure}
\caption{Effect of coverage rate and kernel bandwidth on metrics for CONFEX-Tree on the AdultIncome dataset, RandomForest. \confexnaive\ is represented by dashed horizontal lines.}
\label{Figure 2}
\end{figure}

\subsubsection{CFX generation results}

\begin{table}[H]
\begin{tabular}{llllll}
\toprule
Generator & Distance & Plausibility & Implausibility & Sensitivity $(10^{-1})$ & Stability \\
\midrule
\textbf{MLP} &  &  &  &  &  \\
MinDist & 1.23 ± 0.02 & -0.17 ± 0.05 & 2.00 ± 0.00 & 0.06 ± 0.01 & 0.31 ± 0.00 \\
Wachter & 0.51 ± 0.06 & 0.23 ± 0.01 & 1.90 ± 0.02 & 0.15 ± 0.00 & 0.18 ± 0.00 \\
Greedy & 0.85 ± 0.02 & 0.11 ± 0.07 & 2.07 ± 0.00 & 0.01 ± 0.00 & 0.84 ± 0.01 \\
ConfexNaive &  &  &  &  &  \\
$\ \ \alpha = 0.01$ & 1.27 ± 0.07 & -0.16 ± 0.02 & 2.00 ± 0.00 & 0.05 ± 0.00 & 0.41 ± 0.01 \\
$\ \ \alpha = 0.05$ & 1.24 ± 0.05 & -0.18 ± 0.02 & 2.00 ± 0.00 & 0.05 ± 0.01 & 0.35 ± 0.00 \\
$\ \ \alpha = 0.1$ & 1.23 ± 0.02 & -0.16 ± 0.04 & 2.00 ± 0.01 & 0.06 ± 0.01 & 0.33 ± 0.00 \\
ECCCo &  &  &  &  &  \\
$\ \ \alpha = 0.01$ & 0.57 ± 0.01 & 0.13 ± 0.01 & 1.88 ± 0.01 & 0.05 ± 0.00 & 0.37 ± 0.01 \\
$\ \ \alpha = 0.01$ & 0.71 ± 0.02 & -0.05 ± 0.03 & 1.91 ± 0.01 & 0.06 ± 0.00 & 0.53 ± 0.04 \\
$\ \ \alpha = 0.05$ & 0.57 ± 0.00 & 0.12 ± 0.02 & 1.88 ± 0.01 & 0.05 ± 0.00 & 0.37 ± 0.02 \\
$\ \ \alpha = 0.05$ & 0.70 ± 0.02 & -0.04 ± 0.02 & 1.91 ± 0.01 & 0.06 ± 0.00 & 0.51 ± 0.04 \\
$\ \ \alpha = 0.1$ & 0.56 ± 0.01 & 0.12 ± 0.02 & 1.88 ± 0.01 & 0.05 ± 0.00 & 0.37 ± 0.02 \\
$\ \ \alpha = 0.1$ & 0.70 ± 0.02 & -0.04 ± 0.02 & 1.91 ± 0.01 & 0.06 ± 0.00 & 0.51 ± 0.05 \\
ConfexTree, $\alpha = 0.01$ &  &  &  &  &  \\
$\ \ \text{bw} = 0.05$ & 1.86 ± 0.08 & -0.16 ± 0.12 & 1.84 ± 0.00 & 0.04 ± 0.01 & 0.27 ± 0.00 \\
$\ \ \text{bw} = 0.1$ & 1.22 ± 0.00 & -0.02 ± 0.04 & 1.87 ± 0.01 & 0.06 ± 0.00 & 0.32 ± 0.01 \\
$\ \ \text{bw} = 0.15$ & 1.05 ± 0.01 & -0.08 ± 0.00 & 1.90 ± 0.00 & 0.05 ± 0.01 & 0.33 ± 0.02 \\
$\ \ \text{bw} = 0.2$ & 1.17 ± 0.02 & -0.12 ± 0.04 & 1.95 ± 0.01 & 0.05 ± 0.01 & 0.34 ± 0.01 \\
$\ \ \text{bw} = 0.25$ & 1.14 ± 0.02 & -0.14 ± 0.06 & 1.96 ± 0.01 & 0.04 ± 0.01 & 0.34 ± 0.01 \\
$\ \ \text{bw} = 0.3$ & 1.14 ± 0.03 & -0.14 ± 0.06 & 1.96 ± 0.01 & 0.05 ± 0.02 & 0.33 ± 0.01 \\
$\ \ \text{bw} = 0.35$ & 1.11 ± 0.04 & -0.13 ± 0.05 & 1.95 ± 0.01 & 0.05 ± 0.02 & 0.35 ± 0.01 \\
$\ \ \text{bw} = 0.4$ & 1.31 ± 0.05 & -0.16 ± 0.02 & 2.00 ± 0.00 & 0.05 ± 0.00 & 0.42 ± 0.01 \\
$\ \ \text{bw} = 0.45$ & 1.29 ± 0.06 & -0.16 ± 0.02 & 2.00 ± 0.00 & 0.05 ± 0.00 & 0.42 ± 0.01 \\
ConfexTree, $\alpha = 0.05$ &  &  &  &  &  \\
$\ \ \text{bw} = 0.05$ & 1.19 ± 0.02 & 0.01 ± 0.01 & 1.89 ± 0.01 & 0.07 ± 0.00 & 0.27 ± 0.00 \\
$\ \ \text{bw} = 0.1$ & 1.07 ± 0.01 & -0.05 ± 0.03 & 1.94 ± 0.01 & 0.06 ± 0.01 & 0.28 ± 0.00 \\
$\ \ \text{bw} = 0.15$ & 1.09 ± 0.03 & -0.10 ± 0.02 & 1.96 ± 0.01 & 0.06 ± 0.00 & 0.30 ± 0.01 \\
$\ \ \text{bw} = 0.2$ & 1.12 ± 0.04 & -0.20 ± 0.00 & 1.97 ± 0.01 & 0.05 ± 0.00 & 0.33 ± 0.01 \\
$\ \ \text{bw} = 0.25$ & 1.09 ± 0.06 & -0.17 ± 0.01 & 1.97 ± 0.00 & 0.05 ± 0.00 & 0.32 ± 0.00 \\
$\ \ \text{bw} = 0.3$ & 1.09 ± 0.05 & -0.17 ± 0.01 & 1.97 ± 0.00 & 0.05 ± 0.00 & 0.33 ± 0.01 \\
$\ \ \text{bw} = 0.35$ & 1.09 ± 0.06 & -0.17 ± 0.03 & 1.97 ± 0.00 & 0.05 ± 0.00 & 0.33 ± 0.00 \\
$\ \ \text{bw} = 0.4$ & 1.24 ± 0.07 & -0.18 ± 0.02 & 2.00 ± 0.01 & 0.05 ± 0.01 & 0.34 ± 0.00 \\
$\ \ \text{bw} = 0.45$ & 1.24 ± 0.06 & -0.18 ± 0.02 & 2.00 ± 0.00 & 0.05 ± 0.00 & 0.35 ± 0.00 \\
ConfexTree, $\alpha = 0.1$ &  &  &  &  &  \\
$\ \ \text{bw} = 0.05$ & 1.09 ± 0.04 & -0.02 ± 0.06 & 1.94 ± 0.01 & 0.07 ± 0.00 & 0.26 ± 0.01 \\
$\ \ \text{bw} = 0.1$ & 1.09 ± 0.03 & -0.11 ± 0.03 & 1.95 ± 0.01 & 0.06 ± 0.01 & 0.29 ± 0.01 \\
$\ \ \text{bw} = 0.15$ & 1.08 ± 0.03 & -0.13 ± 0.03 & 1.96 ± 0.01 & 0.06 ± 0.01 & 0.30 ± 0.00 \\
$\ \ \text{bw} = 0.2$ & 1.06 ± 0.02 & -0.16 ± 0.02 & 1.94 ± 0.01 & 0.06 ± 0.01 & 0.32 ± 0.00 \\
$\ \ \text{bw} = 0.25$ & 1.13 ± 0.05 & -0.16 ± 0.04 & 1.98 ± 0.01 & 0.05 ± 0.01 & 0.32 ± 0.00 \\
$\ \ \text{bw} = 0.3$ & 1.12 ± 0.05 & -0.17 ± 0.05 & 1.97 ± 0.01 & 0.06 ± 0.00 & 0.32 ± 0.01 \\
$\ \ \text{bw} = 0.35$ & 1.06 ± 0.04 & -0.17 ± 0.03 & 1.95 ± 0.01 & 0.06 ± 0.00 & 0.32 ± 0.00 \\
$\ \ \text{bw} = 0.4$ & 1.22 ± 0.05 & -0.17 ± 0.03 & 2.00 ± 0.00 & 0.05 ± 0.00 & 0.33 ± 0.01 \\
$\ \ \text{bw} = 0.45$ & 1.23 ± 0.03 & -0.17 ± 0.03 & 2.00 ± 0.01 & 0.05 ± 0.00 & 0.33 ± 0.00 \\
\bottomrule
\end{tabular}

\caption{CFX generation results, AdultIncome, MLP.  Validity 70\% for Wachter, 84-85\% for all ECCCo methods, 81.5\% for Greedy.}
\end{table}

\begin{table}[H]

\begin{tabular}{llllll}
\toprule
Generator & Distance & Plausibility & Implausibility & Sensitivity $(10^{-1})$ & Stability \\
\midrule
\textbf{RandomForest} &  &  &  &  &  \\
MinDist & 0.93 ± 0.00 & -0.02 ± 0.08 & 1.94 ± 0.00 & 0.14 ± 0.02 & 0.26 ± 0.01 \\
ConfexNaive &  &  &  &  &  \\
$\ \ \alpha = 0.01$ & 1.43 ± 0.20 & -0.19 ± 0.11 & 1.91 ± 0.04 & 0.09 ± 0.02 & 0.31 ± 0.02 \\
$\ \ \alpha = 0.05$ & 1.03 ± 0.07 & 0.01 ± 0.01 & 1.95 ± 0.01 & 0.08 ± 0.01 & 0.26 ± 0.01 \\
$\ \ \alpha = 0.1$ & 0.97 ± 0.08 & -0.08 ± 0.06 & 1.95 ± 0.02 & 0.13 ± 0.02 & 0.26 ± 0.01 \\
ConfexTree, $\alpha = 0.01$ &  &  &  &  &  \\
$\ \ \text{bw} = 0.05$ & 1.67 ± 0.17 & -0.08 ± 0.06 & 1.84 ± 0.02 & 0.05 ± 0.01 & 0.19 ± 0.03 \\
$\ \ \text{bw} = 0.1$ & 1.39 ± 0.17 & -0.01 ± 0.17 & 1.80 ± 0.01 & 0.11 ± 0.01 & 0.25 ± 0.02 \\
$\ \ \text{bw} = 0.15$ & 1.28 ± 0.25 & -0.03 ± 0.19 & 1.83 ± 0.01 & 0.09 ± 0.01 & 0.26 ± 0.02 \\
$\ \ \text{bw} = 0.2$ & 1.41 ± 0.20 & -0.16 ± 0.14 & 1.92 ± 0.02 & 0.11 ± 0.00 & 0.26 ± 0.02 \\
$\ \ \text{bw} = 0.25$ & 1.31 ± 0.17 & -0.07 ± 0.05 & 1.95 ± 0.01 & 0.09 ± 0.02 & 0.26 ± 0.01 \\
$\ \ \text{bw} = 0.3$ & 1.29 ± 0.20 & -0.05 ± 0.05 & 1.96 ± 0.02 & 0.07 ± 0.01 & 0.25 ± 0.01 \\
$\ \ \text{bw} = 0.35$ & 1.31 ± 0.17 & -0.11 ± 0.03 & 1.97 ± 0.00 & 0.09 ± 0.03 & 0.26 ± 0.01 \\
$\ \ \text{bw} = 0.4$ & 1.46 ± 0.20 & -0.19 ± 0.11 & 1.93 ± 0.04 & 0.07 ± 0.01 & 0.31 ± 0.02 \\
$\ \ \text{bw} = 0.45$ & 1.47 ± 0.21 & -0.19 ± 0.11 & 1.93 ± 0.03 & 0.10 ± 0.03 & 0.31 ± 0.02 \\
ConfexTree, $\alpha = 0.05$ &  &  &  &  &  \\
$\ \ \text{bw} = 0.05$ & 0.96 ± 0.04 & 0.13 ± 0.05 & 1.90 ± 0.00 & 0.11 ± 0.03 & 0.18 ± 0.00 \\
$\ \ \text{bw} = 0.1$ & 0.95 ± 0.10 & 0.04 ± 0.04 & 1.92 ± 0.01 & 0.08 ± 0.02 & 0.23 ± 0.01 \\
$\ \ \text{bw} = 0.15$ & 0.92 ± 0.08 & 0.08 ± 0.12 & 1.93 ± 0.01 & 0.14 ± 0.04 & 0.22 ± 0.01 \\
$\ \ \text{bw} = 0.2$ & 0.94 ± 0.01 & -0.06 ± 0.02 & 1.96 ± 0.01 & 0.10 ± 0.01 & 0.24 ± 0.02 \\
$\ \ \text{bw} = 0.25$ & 0.94 ± 0.01 & -0.06 ± 0.00 & 1.97 ± 0.01 & 0.09 ± 0.00 & 0.24 ± 0.01 \\
$\ \ \text{bw} = 0.3$ & 0.96 ± 0.04 & -0.06 ± 0.04 & 1.97 ± 0.00 & 0.07 ± 0.02 & 0.25 ± 0.01 \\
$\ \ \text{bw} = 0.35$ & 0.98 ± 0.07 & -0.07 ± 0.01 & 1.96 ± 0.00 & 0.10 ± 0.01 & 0.25 ± 0.01 \\
$\ \ \text{bw} = 0.4$ & 1.04 ± 0.05 & 0.02 ± 0.00 & 1.95 ± 0.01 & 0.10 ± 0.03 & 0.26 ± 0.01 \\
$\ \ \text{bw} = 0.45$ & 1.03 ± 0.05 & -0.01 ± 0.05 & 1.95 ± 0.01 & 0.10 ± 0.01 & 0.26 ± 0.01 \\
ConfexTree, $\alpha = 0.1$ &  &  &  &  &  \\
$\ \ \text{bw} = 0.05$ & 0.75 ± 0.06 & 0.17 ± 0.07 & 1.90 ± 0.00 & 0.22 ± 0.08 & 0.18 ± 0.01 \\
$\ \ \text{bw} = 0.1$ & 0.85 ± 0.02 & 0.04 ± 0.06 & 1.93 ± 0.01 & 0.11 ± 0.00 & 0.22 ± 0.01 \\
$\ \ \text{bw} = 0.15$ & 0.89 ± 0.01 & 0.00 ± 0.06 & 1.94 ± 0.00 & 0.11 ± 0.01 & 0.23 ± 0.01 \\
$\ \ \text{bw} = 0.2$ & 0.93 ± 0.03 & 0.00 ± 0.08 & 1.91 ± 0.01 & 0.09 ± 0.01 & 0.25 ± 0.01 \\
$\ \ \text{bw} = 0.25$ & 0.86 ± 0.02 & 0.09 ± 0.05 & 1.90 ± 0.00 & 0.12 ± 0.01 & 0.25 ± 0.01 \\
$\ \ \text{bw} = 0.3$ & 0.90 ± 0.05 & 0.04 ± 0.06 & 1.91 ± 0.01 & 0.14 ± 0.02 & 0.26 ± 0.00 \\
$\ \ \text{bw} = 0.35$ & 0.90 ± 0.04 & 0.05 ± 0.01 & 1.91 ± 0.01 & 0.12 ± 0.01 & 0.26 ± 0.00 \\
$\ \ \text{bw} = 0.4$ & 0.96 ± 0.03 & 0.00 ± 0.02 & 1.95 ± 0.02 & 0.14 ± 0.01 & 0.26 ± 0.00 \\
$\ \ \text{bw} = 0.45$ & 0.97 ± 0.06 & -0.06 ± 0.04 & 1.96 ± 0.02 & 0.13 ± 0.01 & 0.26 ± 0.00 \\
FeatureTweak & 0.34 ± 0.07 & 0.23 ± 0.09 & 1.87 ± 0.01 & 0.05 ± 0.01 & 0.14 ± 0.00 \\
FOCUS & 0.55 ± 0.12 & 0.36 ± 0.06 & 1.86 ± 0.00 & 0.21 ± 0.09 & 0.17 ± 0.00 \\
\bottomrule
\end{tabular}

\caption{CFX generation results, AdultIncome, RandomForest. Methods with nan values had 100\% failures. Validity 73.5\% for FeatureTweak.}
\end{table}

\subsubsection{Conformal evaluation results}
\begin{tabular}{lllll}
\toprule
Generator & Marginal CovGap & Binning CovGap & Class Cond CovGap & Simulated CovGap \\
\midrule
\textbf{MLP} &  &  &  &  \\
ConfexNaive &  &  &  &  \\
$\ \ \alpha = 0.01$ & 0.99 ± 0.00 & -0.90 ± 0.05 & -0.03 ± 0.01 & -3.39 ± 0.72 \\
$\ \ \alpha = 0.05$ & 0.95 ± 0.00 & -2.52 ± 0.11 & 0.29 ± 0.09 & -6.16 ± 0.50 \\
$\ \ \alpha = 0.1$ & 0.90 ± 0.00 & -4.05 ± 0.23 & 0.22 ± 0.12 & -13.00 ± 0.41 \\
ConfexTree, $\alpha = 0.01$ &  &  &  &  \\
$\ \ \text{bw} = 0.05$ & 1.00 ± 0.00 & 0.17 ± 0.02 & 0.56 ± 0.01 & -15.09 ± 1.74 \\
$\ \ \text{bw} = 0.1$ & 0.99 ± 0.00 & -0.38 ± 0.20 & 0.25 ± 0.11 & -2.24 ± 1.14 \\
$\ \ \text{bw} = 0.15$ & 1.00 ± 0.00 & -0.38 ± 0.20 & 0.25 ± 0.11 & -2.03 ± 1.27 \\
$\ \ \text{bw} = 0.2$ & 0.99 ± 0.00 & -1.07 ± 0.10 & -0.12 ± 0.06 & -4.28 ± 1.24 \\
$\ \ \text{bw} = 0.25$ & 0.99 ± 0.00 & -1.01 ± 0.04 & -0.09 ± 0.04 & -4.13 ± 1.16 \\
$\ \ \text{bw} = 0.3$ & 0.99 ± 0.00 & -1.09 ± 0.04 & -0.15 ± 0.05 & -4.64 ± 0.54 \\
$\ \ \text{bw} = 0.35$ & 0.99 ± 0.00 & -0.94 ± 0.05 & -0.05 ± 0.05 & -4.06 ± 1.91 \\
$\ \ \text{bw} = 0.4$ & 0.99 ± 0.00 & -0.88 ± 0.12 & 0.01 ± 0.09 & -2.29 ± 2.00 \\
$\ \ \text{bw} = 0.45$ & 0.99 ± 0.00 & -0.88 ± 0.12 & 0.01 ± 0.09 & -2.29 ± 2.00 \\
ConfexTree, $\alpha = 0.05$ &  &  &  &  \\
$\ \ \text{bw} = 0.05$ & 0.97 ± 0.00 & 0.60 ± 0.08 & 2.35 ± 0.06 & -17.60 ± 3.79 \\
$\ \ \text{bw} = 0.1$ & 0.96 ± 0.00 & -1.45 ± 0.05 & 1.04 ± 0.12 & -11.60 ± 2.97 \\
$\ \ \text{bw} = 0.15$ & 0.95 ± 0.00 & -1.85 ± 0.08 & 0.81 ± 0.10 & -11.82 ± 2.14 \\
$\ \ \text{bw} = 0.2$ & 0.95 ± 0.01 & -2.40 ± 0.41 & 0.43 ± 0.25 & -8.53 ± 0.65 \\
$\ \ \text{bw} = 0.25$ & 0.95 ± 0.01 & -2.50 ± 0.26 & 0.31 ± 0.18 & -9.09 ± 0.73 \\
$\ \ \text{bw} = 0.3$ & 0.95 ± 0.00 & -2.58 ± 0.27 & 0.24 ± 0.14 & -9.39 ± 0.70 \\
$\ \ \text{bw} = 0.35$ & 0.95 ± 0.00 & -2.60 ± 0.33 & 0.26 ± 0.21 & -9.07 ± 0.16 \\
$\ \ \text{bw} = 0.4$ & 0.95 ± 0.01 & -2.88 ± 0.23 & -0.05 ± 0.16 & -6.05 ± 0.00 \\
$\ \ \text{bw} = 0.45$ & 0.95 ± 0.01 & -2.88 ± 0.23 & -0.05 ± 0.16 & -6.29 ± 0.32 \\
ConfexTree, $\alpha = 0.1$ &  &  &  &  \\
$\ \ \text{bw} = 0.05$ & 0.95 ± 0.00 & 1.84 ± 0.03 & 4.68 ± 0.01 & -14.39 ± 2.19 \\
$\ \ \text{bw} = 0.1$ & 0.93 ± 0.00 & 0.29 ± 0.15 & 3.65 ± 0.04 & -9.68 ± 1.54 \\
$\ \ \text{bw} = 0.15$ & 0.93 ± 0.00 & -0.37 ± 0.17 & 3.19 ± 0.05 & -10.49 ± 1.50 \\
$\ \ \text{bw} = 0.2$ & 0.92 ± 0.00 & -2.21 ± 0.04 & 1.79 ± 0.06 & -13.18 ± 1.44 \\
$\ \ \text{bw} = 0.25$ & 0.92 ± 0.00 & -2.16 ± 0.17 & 1.74 ± 0.07 & -13.12 ± 0.12 \\
$\ \ \text{bw} = 0.3$ & 0.92 ± 0.00 & -2.17 ± 0.17 & 1.72 ± 0.05 & -12.44 ± 0.68 \\
$\ \ \text{bw} = 0.35$ & 0.92 ± 0.00 & -2.32 ± 0.15 & 1.62 ± 0.05 & -12.37 ± 0.38 \\
$\ \ \text{bw} = 0.4$ & 0.90 ± 0.00 & -4.18 ± 0.24 & 0.27 ± 0.18 & -13.37 ± 0.54 \\
$\ \ \text{bw} = 0.45$ & 0.90 ± 0.00 & -4.18 ± 0.24 & 0.27 ± 0.18 & -13.37 ± 0.56 \\
\bottomrule
\end{tabular}
\begin{table}[H]

\caption{Conformal evaluation results, AdultIncome, MLP}
\end{table}

\begin{table}[H]

\begin{tabular}{lllll}
\toprule
Generator & Marginal CovGap & Binning CovGap & Class Cond CovGap & Simulated CovGap \\
\midrule
\textbf{RandomForest} &  &  &  &  \\
ConfexNaive &  &  &  &  \\
$\ \ \alpha = 0.01$ & 0.99 ± 0.00 & -0.90 ± 0.22 & 0.01 ± 0.09 & -5.70 ± 5.11 \\
$\ \ \alpha = 0.05$ & 0.96 ± 0.00 & -2.81 ± 0.08 & 0.35 ± 0.12 & -5.49 ± 2.52 \\
$\ \ \alpha = 0.1$ & 0.91 ± 0.00 & -5.78 ± 0.58 & 0.13 ± 0.14 & -10.19 ± 0.48 \\
ConfexTree, $\alpha = 0.01$ &  &  &  &  \\
$\ \ \text{bw} = 0.05$ & 1.00 ± 0.00 & 0.20 ± 0.02 & 0.56 ± 0.01 & -5.11 ± 2.96 \\
$\ \ \text{bw} = 0.1$ & 0.99 ± 0.00 & -0.43 ± 0.08 & 0.23 ± 0.03 & -2.95 ± 0.14 \\
$\ \ \text{bw} = 0.15$ & 0.99 ± 0.00 & -0.45 ± 0.12 & 0.23 ± 0.05 & -2.67 ± 0.17 \\
$\ \ \text{bw} = 0.2$ & 0.99 ± 0.00 & -0.80 ± 0.12 & 0.05 ± 0.04 & -3.32 ± 0.27 \\
$\ \ \text{bw} = 0.25$ & 0.99 ± 0.00 & -0.78 ± 0.13 & 0.04 ± 0.03 & -3.89 ± 0.55 \\
$\ \ \text{bw} = 0.3$ & 0.99 ± 0.00 & -0.87 ± 0.19 & -0.00 ± 0.07 & -4.68 ± 0.38 \\
$\ \ \text{bw} = 0.35$ & 0.99 ± 0.00 & -0.79 ± 0.20 & 0.04 ± 0.08 & -3.76 ± 0.17 \\
$\ \ \text{bw} = 0.4$ & 0.99 ± 0.00 & -0.91 ± 0.24 & -0.00 ± 0.10 & -7.54 ± 4.96 \\
$\ \ \text{bw} = 0.45$ & 0.99 ± 0.00 & -0.91 ± 0.24 & -0.00 ± 0.10 & -7.54 ± 4.96 \\
ConfexTree, $\alpha = 0.05$ &  &  &  &  \\
$\ \ \text{bw} = 0.05$ & 0.97 ± 0.01 & 0.03 ± 0.06 & 1.84 ± 0.08 & -6.43 ± 1.84 \\
$\ \ \text{bw} = 0.1$ & 0.96 ± 0.00 & -1.60 ± 0.23 & 0.88 ± 0.11 & -5.54 ± 1.04 \\
$\ \ \text{bw} = 0.15$ & 0.95 ± 0.00 & -2.03 ± 0.04 & 0.58 ± 0.05 & -5.39 ± 0.67 \\
$\ \ \text{bw} = 0.2$ & 0.95 ± 0.00 & -2.59 ± 0.01 & 0.44 ± 0.02 & -6.10 ± 0.05 \\
$\ \ \text{bw} = 0.25$ & 0.96 ± 0.00 & -2.66 ± 0.19 & 0.38 ± 0.10 & -6.74 ± 0.05 \\
$\ \ \text{bw} = 0.3$ & 0.96 ± 0.00 & -2.59 ± 0.12 & 0.43 ± 0.05 & -6.17 ± 0.05 \\
$\ \ \text{bw} = 0.35$ & 0.96 ± 0.00 & -2.52 ± 0.08 & 0.47 ± 0.02 & -6.79 ± 0.05 \\
$\ \ \text{bw} = 0.4$ & 0.95 ± 0.00 & -2.82 ± 0.10 & 0.31 ± 0.18 & -6.25 ± 0.14 \\
$\ \ \text{bw} = 0.45$ & 0.95 ± 0.00 & -2.82 ± 0.10 & 0.31 ± 0.18 & -6.24 ± 0.12 \\
ConfexTree, $\alpha = 0.1$ &  &  &  &  \\
$\ \ \text{bw} = 0.05$ & 0.93 ± 0.00 & -0.11 ± 0.80 & 2.91 ± 0.35 & -4.98 ± 0.44 \\
$\ \ \text{bw} = 0.1$ & 0.91 ± 0.00 & -2.48 ± 0.49 & 1.38 ± 0.18 & -4.62 ± 1.45 \\
$\ \ \text{bw} = 0.15$ & 0.91 ± 0.01 & -3.14 ± 0.83 & 1.00 ± 0.34 & -4.33 ± 1.62 \\
$\ \ \text{bw} = 0.2$ & 0.91 ± 0.00 & -4.33 ± 0.10 & 0.66 ± 0.05 & -4.80 ± 0.47 \\
$\ \ \text{bw} = 0.25$ & 0.91 ± 0.00 & -4.50 ± 0.07 & 0.54 ± 0.16 & -5.34 ± 0.81 \\
$\ \ \text{bw} = 0.3$ & 0.91 ± 0.00 & -4.67 ± 0.15 & 0.38 ± 0.20 & -5.73 ± 1.22 \\
$\ \ \text{bw} = 0.35$ & 0.91 ± 0.00 & -4.88 ± 0.36 & 0.25 ± 0.33 & -5.56 ± 1.45 \\
$\ \ \text{bw} = 0.4$ & 0.91 ± 0.00 & -6.14 ± 0.35 & 0.04 ± 0.06 & -10.77 ± 3.43 \\
$\ \ \text{bw} = 0.45$ & 0.91 ± 0.00 & -6.14 ± 0.34 & 0.03 ± 0.05 & -10.87 ± 3.20 \\
\bottomrule
\end{tabular}

\caption{Conformal evaluation results, AdultIncome, RandomForest}
\end{table}

\end{document}